\documentclass[10pt,journal,compsoc]{IEEEtran}
\usepackage{hyperref}
\usepackage{amssymb}
\usepackage{mathtools}
\usepackage{amsmath}                   
\usepackage{color}
\usepackage{mathptmx}
\usepackage{amssymb}
\usepackage{enumitem}
\usepackage{multicol}
\usepackage{anyfontsize}
\usepackage{amsmath}
\usepackage{algorithmic}
\usepackage{algorithm}
\usepackage{letltxmacro}
\usepackage{esvect}
\usepackage{wrapfig}
\usepackage{multicol}
\usepackage[dvipsnames]{xcolor}
\DeclareMathAlphabet{\mathcal}{OMS}{cmsy}{m}{n}

\DeclareMathOperator*{\argmin}{arg\,min}

\ifCLASSOPTIONcompsoc
  \usepackage[nocompress]{cite}
\else
  \usepackage{cite}
\fi
\ifCLASSINFOpdf
\else
\fi
\begin{document}
%
\title{Wasserstein Auto-Encoders of Merge Trees \\(and Persistence Diagrams)}
%
%
%
%

\author{Mathieu Pont and Julien Tierny
\IEEEcompsocitemizethanks{\IEEEcompsocthanksitem Mathieu Pont and Julien Tierny are with the CNRS/Sorbonne Universit\'e.\protect\\
E-mails: \{firstname.lastname\}@sorbonne-universite.fr}
\thanks{}
}

\newcommand{\domain}{\mathcal{M}}
\newcommand{\range}{\mathbb{R}}
\newcommand{\sublevelset}[1]{#1^{-1}_{-\infty}}
\newcommand{\superlevelset}[1]{#1^{-1}_{+\infty}}
\newcommand{\Star}{St}
\newcommand{\Link}{Lk}
\newcommand{\simplex}{\sigma}
\newcommand{\face}{\tau}
\newcommand{\lowerlink}{\Link^{-}}
\newcommand{\upperlink}{\Link^{+}}
\newcommand{\Index}{\mathcal{I}}
\newcommand{\offset}{o}
\newcommand{\Natural}{\mathbb{N}}
\newcommand{\criticalSet}{\mathcal{C}}
\newcommand{\diagram}{\mathcal{D}}
\newcommand{\wasserstein}[1]{W^{\diagram}_#1}
\newcommand{\projection}{\Delta}
\newcommand{\hierarchy}{\mathcal{H}}
\newcommand{\decimation}{D}
\newcommand{\xDimD}{L_x^\decimation}
\newcommand{\yDimD}{L_y^\decimation}
\newcommand{\zDimD}{L_z^\decimation}
\newcommand{\xDim}{L_x}
\newcommand{\yDim}{L_y}
\newcommand{\zDim}{L_z}
\newcommand{\Grid}{\mathcal{G}}
\newcommand{\GridD}{\mathcal{G}^\decimation}
\newcommand{\x}{\phantom{x}}
\newcommand{\Mod}{\;\mathrm{mod}\;}
\newcommand{\NN}{\mathbb{N}}
\newcommand{\forwardIntegralLine}{\mathcal{L}^+}
\newcommand{\backwardIntegralLine}{\mathcal{L}^-}
\newcommand{\triangulationOp}{\phi}
\newcommand{\decimationOp}{\Pi}
\newcommand{\isovalue}{w}
\newcommand{\persistence}{\mathcal{P}}
\newcommand{\pointMetric}{d}
\newcommand{\diagramSet}{\mathcal{S}_\mathcal{D}}
\newcommand{\diagramSpace}{\mathbb{D}}
\newcommand{\jointree}{\mathcal{T}^-}
\newcommand{\splittree}{\mathcal{T}^+}
\newcommand{\mergetree}{\mathcal{T}}
\newcommand{\mergetreeSet}{\mathcal{S}_\mathcal{T}}
\newcommand{\branchset}{\mathcal{S}_\mathcal{B}}
\newcommand{\branchspace}{\mathbb{B}}
\newcommand{\mergetreeSpace}{\mathbb{T}}
\newcommand{\editdistance}{D_E}
\newcommand{\wassersteinTree}{W^{\mergetree}_2}
\newcommand{\distanceSequence}{d_S}
\newcommand{\branchtree}{\mathcal{B}}
\newcommand{\branchtreeSet}{\mathcal{S}_\mathcal{B}}
\newcommand{\branchtreeSpace}{\mathbb{B}}
\newcommand{\forest}{\mathcal{F}}
\newcommand{\sequenceSpace}{\mathbb{S}}
\newcommand{\forestMatrix}{\mathbb{F}}
\newcommand{\treeMatrix}{\mathbb{T}}
\newcommand{\normalizedLocation}{\mathcal{N}}
\newcommand{\normalizedWasserstein}{W^{\normalizedLocation}_2}
\newcommand{\geodesictree}{\mathcal{G}}
\newcommand{\dummyVector}{\mathcal{V}}
\newcommand{\geodesictreeVec}{g}
\newcommand{\geodesicAxis}{\mathcal{A}}
\newcommand{\directionVector}{\mathcal{V}}
\newcommand{\geodesicdiagram}{\mathcal{G}^{\diagram}}
\newcommand{\reconstructionError}{E_{L_2}}
\newcommand{\pcaBasis}{B_{\mathbb{R}^d}}
\renewcommand{\pcaBasis}{B}
\newcommand{\origin}{o_b}
\newcommand{\sizeEncoding}{n_e}
\newcommand{\sizeDecoding}{n_d}
\newcommand{\linearTransformation}{\psi}
\newcommand{\unitTransformation}{\Psi}
\renewcommand{\origin}{o}
\newcommand{\bdtOrigin}{\mathcal{O}}
\newcommand{\activation}{\sigma}
\newcommand{\validBDT}{\gamma}
\newcommand{\mtPgaBasis}{B_{\branchtreeSpace}}
\newcommand{\mtPgaError}{E_{\wassersteinTree}}
\newcommand{\frechetEnergy}{E_F}
\newcommand{\geodesicExtremity}{\mathcal{E}}
\newcommand{\vectorNotation}[1]{\protect\vv{#1}}
\renewcommand{\vectorNotation}[1]{#1}
\newcommand{\axisNotation}[1]{\protect\overleftrightarrow{#1}}
\newcommand{\individualEnergy}{E}
\newcommand{\ensembleSize}{N}
\newcommand{\numberBranchinBarycenter}{N_1}
\newcommand{\numberGeodesicSamples}{N_2}
\newcommand{\planarGridX}{N_x}
\newcommand{\planarGridY}{N_y}
\newcommand{\regularGrid}{G}
\newcommand{\distanceMatrix}{\mathbb{D}}
\newcommand{\maxDimensions}{{d_{max}}}
\newcommand{\projectionOperator}{\mathcal{P}}
\newcommand{\reconstructed}[1]{\widehat{#1}}
\newcommand{\gt}{>}
\newcommand{\lt}{<}
\newcommand{\branch}{b}
\newcommand{\nonLinearFunction}{\sigma}
\newcommand{\batchSequence}{S}

\newcommand{\julien}[1]{\textcolor{red}{#1}}
\renewcommand{\julien}[1]{\textcolor{black}{#1}}
\newcommand{\mathieu}[1]{\textcolor{green}{#1}}
\renewcommand{\mathieu}[1]{\textcolor{green}{#1}}
\newcommand{\jules}[1]{\textcolor{orange}{#1}}
\renewcommand{\jules}[1]{\textcolor{black}{#1}}
\newcommand{\note}[1]{\textcolor{magenta}{#1}}
\newcommand{\cutout}[1]{\textcolor{blue}{#1}}
\renewcommand{\cutout}[1]{}

\newcommand{\revision}[1]{\textcolor{black}{#1}}
\newcommand{\minor}[1]{\textcolor{black}{#1}}

\renewcommand{\figureautorefname}{Fig.}
\renewcommand{\sectionautorefname}{Sec.}
\renewcommand{\subsectionautorefname}{Sec.}
\renewcommand{\equationautorefname}{Eq.}
\renewcommand{\tableautorefname}{Tab.}
\newcommand{\algorithmautorefname}{Alg.}
\newcommand{\lineautorefname}{Alg.}

\newcommand{\eqSpace}{-1.75ex}

\newcommand{\mycaption}[1]{
\caption{#1}
}

\newcommand{\journal}[1]{\textcolor{blue}{#1}}
\renewcommand{\journal}[1]{\textcolor{black}{#1}}

\newlength{\commentindent}
\setlength{\commentindent}{0.6\linewidth}
\makeatletter
\renewcommand{\algorithmiccomment}[1]{\unskip\hfill\makebox[\commentindent][l]{/
/~#1}\par}
\LetLtxMacro{\oldalgorithmic}{\algorithmic}
\renewcommand{\algorithmic}[1][0]{%
  \oldalgorithmic[#1]%
  \renewcommand{\ALC@com}[1]{%
    \ifnum\pdfstrcmp{##1}{default}=0\else\algorithmiccomment{##1}\fi}%
}
\makeatother

\IEEEtitleabstractindextext{%
\begin{abstract}
This paper presents a computational framework for the
Wasserstein
auto-encoding of merge
trees (MT-WAE), a novel extension of the classical auto-encoder neural
network architecture to the Wasserstein metric space of
merge trees. 
In contrast to traditional auto-encoders which operate on vectorized data, our
formulation explicitly manipulates merge trees on their associated metric
space at each layer of the network, resulting in superior
accuracy and
interpretability.
Our novel
neural network approach
can be interpreted as
a non-linear generalization of previous
linear attempts \journal{\cite{pont_tvcg23}}
at merge tree encoding.
It also trivially extends to persistence diagrams.
Extensive experiments on public ensembles demonstrate the efficiency of our
algorithms, with MT-WAE computations in the orders of minutes on average.
We show the utility of our contributions in two
applications
\journal{adapted from previous work on merge tree encoding \cite{pont_tvcg23}.}
First, we apply MT-WAE to
\minor{\emph{merge tree compression},}
by concisely representing them with their coordinates in
the final layer of our auto-encoder.
Second, we
\journal{document an application to \emph{dimensionality reduction}, by }
exploiting the
latent space of our auto-encoder, for the visual analysis
of
ensemble data.
We illustrate the
versatility of our framework by
introducing two penalty terms,
to help preserve
in the latent space
 both the Wasserstein distances between merge trees, as well as their
clusters. 
In both applications, quantitative experiments assess the relevance of
our framework.
Finally, we provide a C++ implementation that can be used
for reproducibility.
\end{abstract}

}

\maketitle

\IEEEdisplaynontitleabstractindextext

%
\IEEEpeerreviewmaketitle


\IEEEraisesectionheading{\section{Introduction}\label{sec:introduction}}

\begin{figure*}
  \centering
  \includegraphics[width=\linewidth]{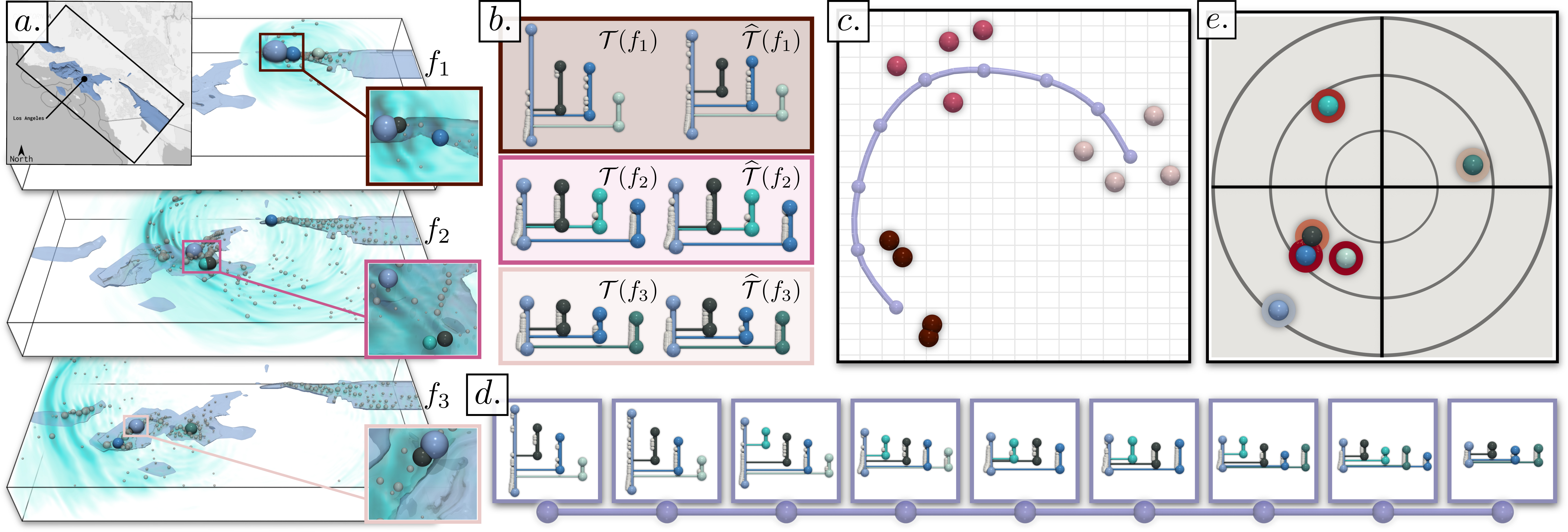}
  \caption{
  Visual analysis
  of the \emph{Earthquake} ensemble $\big((a)$
  \journal{each ground-truth class is represented by one of its members}$\big)$,
  with our Wasserstein Auto-Encoder of Merge Trees (MT-WAE).
  We apply our contributions
to
\minor{merge tree compression}
$\big((b)$, right$\big)$ by simply
storing their coordinates in the last
decoding layer of our network. We exploit the latent space of our network
to generate 2D layouts of the ensemble $(c)$.
In contrast to classical
auto-encoders, MT-WAE
\minor{explicitly manipulates}
merge trees at each layer of the network,
which results in improved accuracy
and interpretability.
Specifically, the reconstruction of user-defined locations $\big((c)$,
purple$)$ enables
an interactive exploration of the latent space:
the reconstructed
curve
$(d)$
enables
\journal{a continuous navigation between the clusters $\big($from dark red to
pink and light pink, $(c)\big)$.}
%
%
MT-WAE also supports persistence correlation views
$(e)$  \journal{(adapted from \cite{pont_tvcg23})}, which
reveal
the barycenter's \journal{persistent} features
\journal{which exhibit the most}
variability in the ensemble (far from the center). Finally,
by tracking the persistence evolution of individual features as they traverse
the network down to its latent space, we introduce a \emph{Feature Latent
Importance} measure, which
\journal{identifies}
the most \emph{informative} features within the
ensemble $\big((e)$, red circles$)$.
}
	\label{fig:teaser}
	\label{fig_teaser}
\end{figure*}

\IEEEPARstart{W}{ith} the recent advances in the development of computation hardware and
acquisition devices, datasets are constantly increasing in size. This size
increase induces an increase in the geometrical complexity of the
features present in the datasets, which challenges interactive data analysis
and interpretation. To address this issue, Topological Data Analysis (TDA)
\cite{edelsbrunner09} has shown over the years its ability to
\journal{reveal, in a generic, robust and efficient manner,}
%
%
the main structural patterns hidden in complex
datasets, in particular for visual data analysis tasks \cite{heine16}.
\journal{Successful applications have been documented in multiple fields}
(turbulent combustion \cite{bremer_tvcg11, gyulassy_ev14},
material sciences \cite{gyulassy_vis15, soler_ldav19},
nuclear energy \cite{beiNuclear16},
fluid dynamics \cite{kasten_tvcg11, nauleau_ldav22},
bioimaging \cite{topoAngler, beiBrain18},
quantum chemistry \cite{Malgorzata19,
olejniczak_pccp23} or
astrophysics \cite{sousbie11, shivashankar2016felix}).  Among the  
representations studied in TDA, the merge tree \cite{carr00}
(\autoref{fig_mergeTree}) has been 
prominent in 
\journal{data}
visualization
\cite{carr04, bremer_tvcg11,
topoAngler}.

\journal{In addition to}
the increase in geometrical complexity discussed
\journal{above,}
a new
challenge has recently emerged in many applications,
\journal{with the notion of}
\emph{ensemble
datasets}.
Such datasets
\journal{encode}
a given phenomenon not only with a single
dataset, but with a collection of datasets, called \emph{ensemble members}.
In
that context, the topological analysis of an ensemble dataset consequently
yields an ensemble of corresponding topological representations (e.g. one merge
tree per ensemble member).

\journal{Then, developing statistical analysis tools to support the interactive
analysis and interpretation of ensemble data becomes an important challenge.}
\journal{Recently, several}
works explored this direction, in particular with the notion of \emph{average
topological representation} \cite{Turner2014,
lacombe2018, vidal_vis19, YanWMGW20, pont_vis21}.
These approaches can produce a topological representation which nicely
summarizes the ensemble. Moreover, their application to clustering
\cite{pont_vis21} reveal its main trends.
However, they do not provide any hints regarding
the variability of the features in the ensemble. For this, Pont et al.
\cite{pont_tvcg23}
recently
extended the notion of principal geodesic
analysis to
ensembles of
merge trees.
However, this approach implicitly assumes a linear relation between the merge
trees of the ensemble. Specifically, it assumes that
\journal{merge tree branches evolve linearly (in the birth/death space,
\autoref{sec_preliminaries}) within the ensemble}.

This paper addresses this issue
with a novel formulation based on neural networks and introduces
the first
framework for the
non-linear encoding of
merge trees, hence resulting in superior
accuracy.
Specifically, we formulate
merge tree
non-linear encoding
as an auto-encoding problem (\autoref{sec_formulation}).
We contribute a novel neural
network
called \emph{Wasserstein
Auto-Encoder of Merge Trees}. This network is based on a novel layer model,
capable of
processing merge trees natively, without pre-vectorization.
We believe this contribution to be of independent interest, as it enables an
accurate
and interpretable
processing of merge
trees
by neural
networks (without restrictions to
auto-encoders).
We contribute an algorithm for the optimization of such a network
(\autoref{sec_algorithm}).
We illustrate the relevance of our contributions for visual analysis with two
applications, data reduction (\autoref{sec_dataReduction}) and
dimensionality reduction (\autoref{sec_dimensionalityReduction}).
\journal{Similarly to previous linear attempts \cite{pont_tvcg23}, since}
our
 approach is based on the Wasserstein distance between merge trees
\cite{pont_vis21}, which
generalizes the Wasserstein distance between
persistence diagrams \cite{edelsbrunner09}, our framework trivially extends to
persistence diagrams by simply adjusting a single parameter.

\subsection{Related work}

\journal{We classify the literature related to our approach into two
categories:}
ensemble
analysis and
topological methods for ensembles.

\noindent
\textbf{\emph{(1)} Ensemble \journal{analysis}:}
\journal{Typical approaches to ensemble visualization first characterize each
member of the ensemble by extracting a geometrical object representing its
features of interest (level sets, streamlines, etc). Next, a
second step considers the ensemble of geometrical objects computed in the
first step, and estimates a single \emph{representative} object,
representing an aggregate of the features of interest found in the ensemble.
For example, level-set variability has been studied with spaghetti plots
\cite{spaghettiPlot}, with specific applications to weather data
\cite{Potter2009,Sanyal2010}. More generally, the variability in curves and
contours have been studied with the notion of box-plots \cite{whitaker2013} and
its variants \cite{Mirzargar2014}. Hummel et al. \cite{Hummel2013} analyzed the
variability in ensembles of flows with a Lagrangian approach. The main trends
present in an ensemble have been studied for ensembles of streamlines
\cite{Ferstl2016} and level sets \cite{Ferstl2016b} via clustering techniques.
Other approaches
focused on
visualizing the geometrical variability in the domain of the position of
critical points \cite{gunther, favelier_vis18} or gradient separatrices
\cite{athawale_tvcg19}.
For ensemble of contour trees,
consistent planar layouts have also been studied \cite{LohfinkWLWG20},
to support the direct visual comparison
of the
trees.
While the latter approaches have a topological aspect, they
focused on the direct visualization of the variability, and not on the
statistical analysis of an ensemble of topological descriptors.}

General purpose methods have been documented for non-linear
encoding
(e.g. \emph{topological auto-encoders}
\cite{MoorHRB20},
or
\emph{Wasserstein
auto-encoders} \cite{TolstikhinBGS18}). Our work drastically
differs from these methods, in terms of design and purpose.
These methods \cite{MoorHRB20, TolstikhinBGS18, HenselMR21}  employ a classical
auto-encoder
(\autoref{sec_euclideanAutoEncoders}) to which they add specialized penalty
terms.
Then, their input
is restricted to point sets (or
vectorized data).
In contrast, our work
focuses
on sets of merge trees (or persistence diagrams). This different
kind
of
input requires a novel neural network model, capable of processing these
topological objects natively
(\autoref{sec_formulation}).

\begin{figure}
\centering
\includegraphics[width=\linewidth]{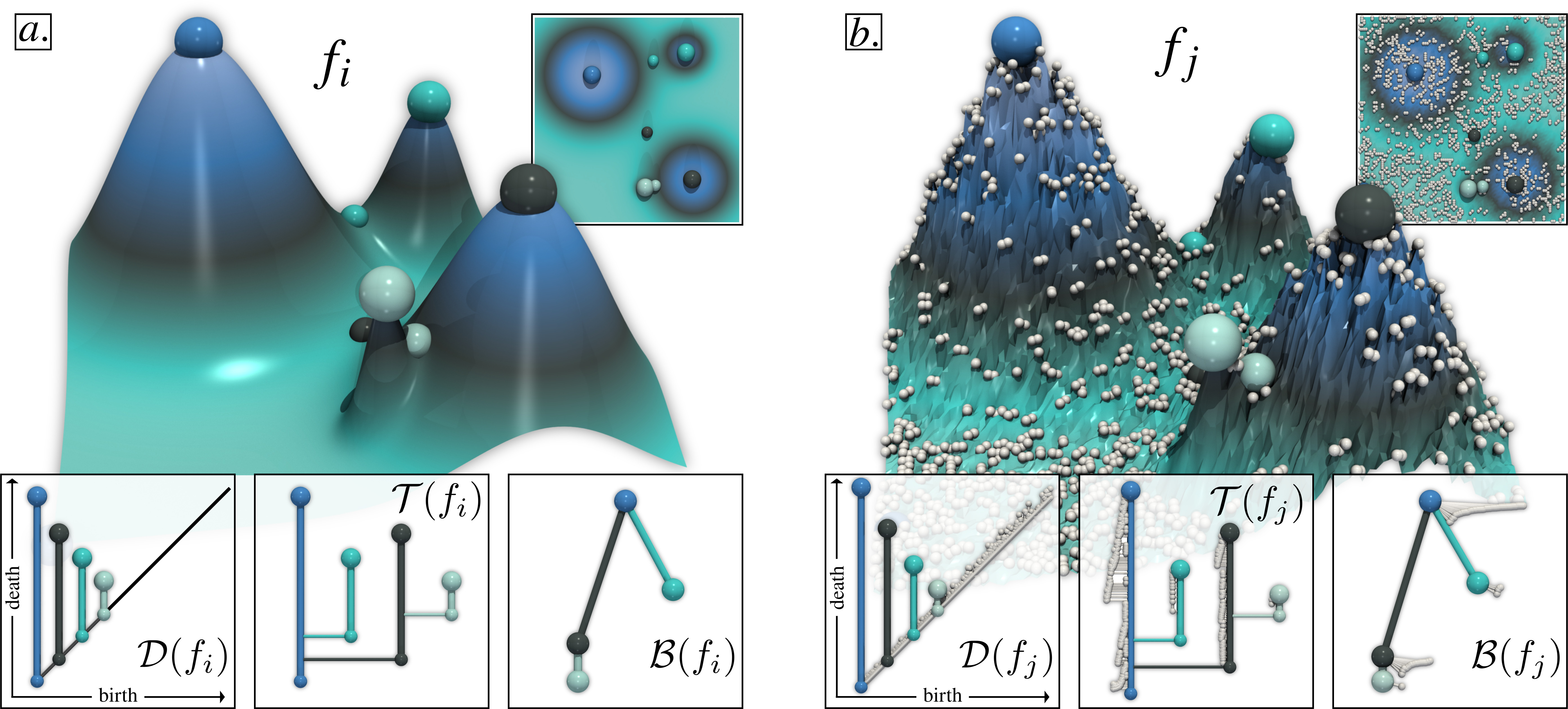}
\mycaption{
\journal{Illustration of the topological descriptors considered in this work,
on a clean $(a)$ and a noisy $(b)$ variant of a 2D toy dataset.
For all descriptors, the color code indicates the persistence of the
corresponding saddle-maximum pair.
Critical points are represented with spheres (larger ones for maxima).
Persistence diagrams, merge trees and branch decomposition trees (BDTs) are
respectively represented in the left, center and right insets.
For both datasets, the four main features (the larger hills) are represented
with salient pairs in the diagram and the merge tree.
To avoid clutter in the visualization, the branches with low persistence (less
than $10\%$ of the
function
range) are rendered with small white arcs while larger,
and colored arcs
represent persistent branches (more than $10\%$ of the
function
range). Figure adapted from \cite{pont_vis21, pont_tvcg23}}
}
\label{fig_mergeTree}
\end{figure}

\noindent
\textbf{\emph{(2)} Topological methods:}
\journal{Over the last two decades, the visualization community has
investigated, adapted and extended \cite{heine16, surveyComparison2021} several
tools and notions from computational topology \cite{edelsbrunner09}.
The persistence diagram \cite{edelsbrunner02, edelsbrunner09, dipha,
guillou_tvcg23}, the Reeb graph \cite{biasotti08, 
parsa12, gueunet_egpgv19}, the
merge (\autoref{fig_mergeTree})
and contour trees \cite{carr00,
MaadasamyDN12, AcharyaN15, CarrWSA16, gueunet_tpds19}, and the
Morse-Smale complex \cite{BremerEHP03, Defl15, robins_pami11, 
ShivashankarN12, gyulassy_vis18} are
popular examples of topological representations in visualization.
}

\journal{In order to design a statistical framework for the analysis of
ensembles of topological descriptors, one first needs to define a metric to
measure distances between these objects.}
\journal{The Wasserstein distance
\cite{edelsbrunner09}
(\autoref{sec_metricSpace}) is now an established and well-documented metric
for persistence diagrams. It is inspired by optimal transport
\cite{Kantorovich, monge81} and it is defined (\autoref{sec_metricSpace}) via a
bipartite assignment problem (open-source software implementing exact
computations \cite{Munkres1957} or fast approximations  \cite{Bertsekas81,
Kerber2016} is available \cite{ttk17}).
However, as discussed in previous work \cite{morozov14, BeketayevYMWH14,
SridharamurthyM20, pont_vis21, pont_tvcg23}, the
persistence diagram can lack specificity in its encoding of the features of
interest, motivating more advanced descriptors, like the merge trees
(\autoref{sec_background_mergeTrees}), which better distinguishes datasets.
The comparison of Reeb graphs and their variants has been addressed with
several similarity measures \cite{HilagaSKK01,
SaikiaSW14_branch_decomposition_comparison}. Several works investigated the
theoretical aspects of distances between topological descriptors, in particular
with a focus towards stable distances \cite{bauer14, morozov14, bollen21,
BollenTL23}. However, the computation of such distances rely on algorithms with
exponential time complexity, which is not practicable for real-life datasets.
In contrast, a distinct line of research focused on a balance between
practical stability and computability, by focusing on polynomial time
computation algorithms. Beketayev et al. \cite{BeketayevYMWH14} introduced a
distance for the \emph{branch decomposition tree} (BDT, Appendix A).}
\journal{Efficient algorithms for constrained edit distances \cite{zhang96}
have been specialized to the specific case of merge trees, hence providing an
edit distance for merge trees \cite{SridharamurthyM20} which is both computable
in practice and with acceptable practical stability. Pont et al.
\cite{pont_vis21} extended this work to generalize the $L_2$-Wasserstein
distance
between persistence diagrams 
\journal{\cite{edelsbrunner09}}  to merge trees, hence 
enabling
the efficient computation of distances, geodesics and barycenters of
merge trees.}
%
Wetzel et al. \cite{WetzelsLG22, WetzelsTopoInVis22} introduce
metrics independent of a particular branch decomposition, but this comes at
the cost of a significantly larger computational effort \journal{(with quartic
time complexity instead of quadratic), which prevents their
practical computation on full-sized merge trees}.


Once a metric is available, statistical notions can
be
developed
for topological
descriptors.
Several
methods
\cite{Turner2014,
lacombe2018, vidal_vis19} have been introduced for the
estimation of barycenters
of persistence diagrams (or
vectorized variants
\cite{Adams2015, Bubenik15}). Similar approaches have been specifically derived
for the merge trees \cite{YanWMGW20, pont_vis21}.
\journal{Another set of approaches \cite{robins16, AnirudhVRT16, LI2021}  first
considers \emph{vectorizations} of
topological descriptors (i.e. by converting them into high-dimensional
Euclidean vectors) and then leverages traditional linear-algebra tools on
these vectors (e.g. the classical PCA \cite{pearson01} or its variants from
matrix sketching \cite{woodRuff2014}).
\minor{Several approaches in machine learning are constructed on top of vectorizations of topological descriptors \cite{Adams2015, Bubenik15, kim20} or kernel-based representations \cite{ReininghausHBK15, CarriereCO17}}.
However, \minor{such vectorizations} have
several limitations in practice. First, they are prone to approximation errors
(resulting from quantization and linearization). Also, they can be difficult to
revert (especially for barycenters), which makes them impractical for
visualization tasks. Moreover, their stability is not always guaranteed.}
In contrast, Pont et al. \cite{pont_tvcg23} extended the generic notion of
principal geodesic analysis to the Wasserstein metric space of merge trees,
resulting in improved accuracy and interpretability with regard to the
straightforward application of PCA on merge tree vectorizations.
\journal{Similarly, Sisouk et al. \cite{sisouk_techrep23} introduced a simpler approach for the linear encoding of persistence diagrams, with a less constrained framework based on dictionaries.}
However,
\journal{these approaches implicitly assume}
a linear relation between the
\journal{topological descriptors}
of the
ensemble. For instance, it assumes that a given feature (i.e. a given branch of
the merge tree) evolves linearly in the birth/death space
(\autoref{sec_preliminaries}) within the ensemble.
However, this
hypothesis
is
easily challenged in practice
(Figs. \ref{fig_pca_vs_eae} and \ref{fig_overview}), potentially leading to
inaccuracies. Our work
overcomes
this limitation with a drastically different formulation (based on
auto-encoding
neural
networks)
and introduces
the first
framework for
the
non-linear encoding of merge trees,
resulting in superior accuracy.
\minor{Several approaches \cite{khrulkov18, Gabrielsson19, zhou2021} investigated the use of topological methods for the analysis of neural networks. In contrast, our work
targets a different research problem, specifically the encoding of topological descriptors with neural networks.}

\subsection{Contributions}
This paper makes the following new contributions:
\begin{enumerate}
 \item \emph{An approach 
 \journal{to Merge tree non-linear encoding:}
 }
 We formulate
 the non-linear parametrization of the Wasserstein
metric space of merge trees (and persistence diagrams)
as an auto-encoding
problem.
Our formulation (\autoref{sec_formulation})
generalizes and improves 
\journal{previous}
linear attempts
\cite{pont_tvcg23}.
 \item \emph{A vectorization-free neural network architecture for Merge Trees:}
We contribute
a novel neural network architecture called
\emph{Wasserstein Auto-Encoder of Merge Trees},
inspired by the
classical auto-encoder, which
can
\emph{natively}
process
merge
trees (and
persistence diagrams) \emph{without} prior vectorization.
For this, we contribute a novel layer model, which
takes a set of merge trees on its input and produces another set of merge trees
on its output, along with their coordinates in the layer's
parametrization.
This results in superior accuracy
(\autoref{sec_quality})
and
interpretability (\autoref{sec_dimensionalityReduction}).
We contribute an algorithm for the optimization of
this
network
(\autoref{sec_algorithm}).
We believe this contribution to be of independent interest.
 \item \emph{An application to
 \minor{merge tree compression:}}
 \journal{We describe how to adapt previous work
 \cite{pont_tvcg23} to our
novel non-linear framework, in
\minor{merge tree compression}
applications
(\autoref{sec_dataReduction}). Specifically,}
the merge trees of the
input ensemble are significantly compressed, by solely storing the final
decoding
layer of the network, as well as the coordinates of the input trees in this
layer.
\journal{We illustrate the interest of our approach with comparisons to linear
encoding \cite{pont_tvcg23} in the context of feature tracking and ensemble
clustering applications.}
 \item \emph{An application to dimensionality reduction:}
 \journal{We describe how to adapt previous work
 \cite{pont_tvcg23} to our
novel non-linear framework, in the context of dimensionality reduction
applications
(\autoref{sec_dimensionalityReduction}). Specifically,}
each tree of
\journal{the}
ensemble \journal{is embedded} as a point in a
planar
view, based on its
coordinates in our
auto-encoder's latent space. To illustrate the versatility of our framework, we
introduce two penalty terms, to improve the preservation of
\journal{clusters and}
distances between merge trees.
 \item \emph{Implementation:} We provide a C++ implementation of our algorithms
that can be used for reproducibility purposes.
\end{enumerate}

\section{Preliminaries}
\label{sec_preliminaries}


This section presents the
background of our work.
First, we describe the input data and its
topological
representation.
Second, 
we recap
the Wasserstein 
metric space of merge trees \cite{pont_vis21}, used by \journal{our} approach.
We refer
to textbooks \cite{edelsbrunner09} for
an
introduction to
computational topology.


\subsection{Input data}
\label{sec_inputData}
\label{sec_background_mergeTrees}
\label{sec_mergeTrees}
The input data is an ensemble of $\ensembleSize$ piecewise linear (PL) scalar
fields
$f_i : \domain \rightarrow \range$, with $i \in \{1, \dots,  \ensembleSize\}$,
defined on a
PL $d$-manifold $\domain$, with
$d\leq3$ in our applications. 
Each ensemble member $f_i$ is represented by a \emph{topological descriptor}. 
In this work, we focus on the \emph{Persistence Diagram} (PD), noted
$\diagram(f_i)$, as well as
a variant of the \emph{Merge Tree} (MT), called
the \emph{Branch Decomposition Tree} (BDT),
noted $\branchtree(f_i)$.
A formal description of these descriptors is given in \julien{Appendix A}.

In short, $\diagram(f_i)$ is a 2D point cloud (\autoref{fig_mergeTree}),
where each off-diagonal point $p = (x, y)$ denotes a \emph{topological
feature} in the data filtration (e.g. a connected component, a
cycle, a void, etc.). $x$ and $y$ denote the \emph{birth} and the
\emph{death} of $p$ (i.e. the scalar values for the creation and destruction of
the corresponding topological feature in the data). The
\emph{persistence} of $p$ is given by its height to the diagonal 
(\autoref{fig_mergeTree}, vertical
cylinders, left insets). Important 
features are typically
associated with a large persistence,
while low amplitude noise is in 
the
vicinity of the diagonal.

The Merge Tree (MT, \autoref{fig_mergeTree}, center insets) is a slightly more
informative descriptor, as it additionally encodes
the merge
history
between the topological features.
To mitigate a phenomenon called \emph{saddle swap} \cite{SridharamurthyM20,
pont_vis21},
it is often pre-processed
to merge adjacent forking nodes whose
relative scalar value difference is smaller than a threshold $\epsilon_1 \in
[0, 1]$. The merge tree
can be represented in
a dual form called the \emph{Branch Decomposition Tree} (BDT) $\branchtree(f_i)$
(\autoref{fig_mergeTree}, right insets), where each persistent branch of the
merge tree (vertical cylinder, center insets) is transformed into a node in the
BDT (sphere, right insets) and where each horizontal segment of the merge tree
is transformed into an arc.
Given an arbitrary BDT $\branchtree$, since each node $b \in \branchtree$
embeds its own
\journal{birth/death}
values (numbers in \autoref{fig_mergeTree}), it
is possible
to reconstruct the corresponding merge tree, as long as
$\branchtree$
respects the \emph{Elder rule} \cite{edelsbrunner09}.

\begin{figure}
\includegraphics[width=\linewidth]{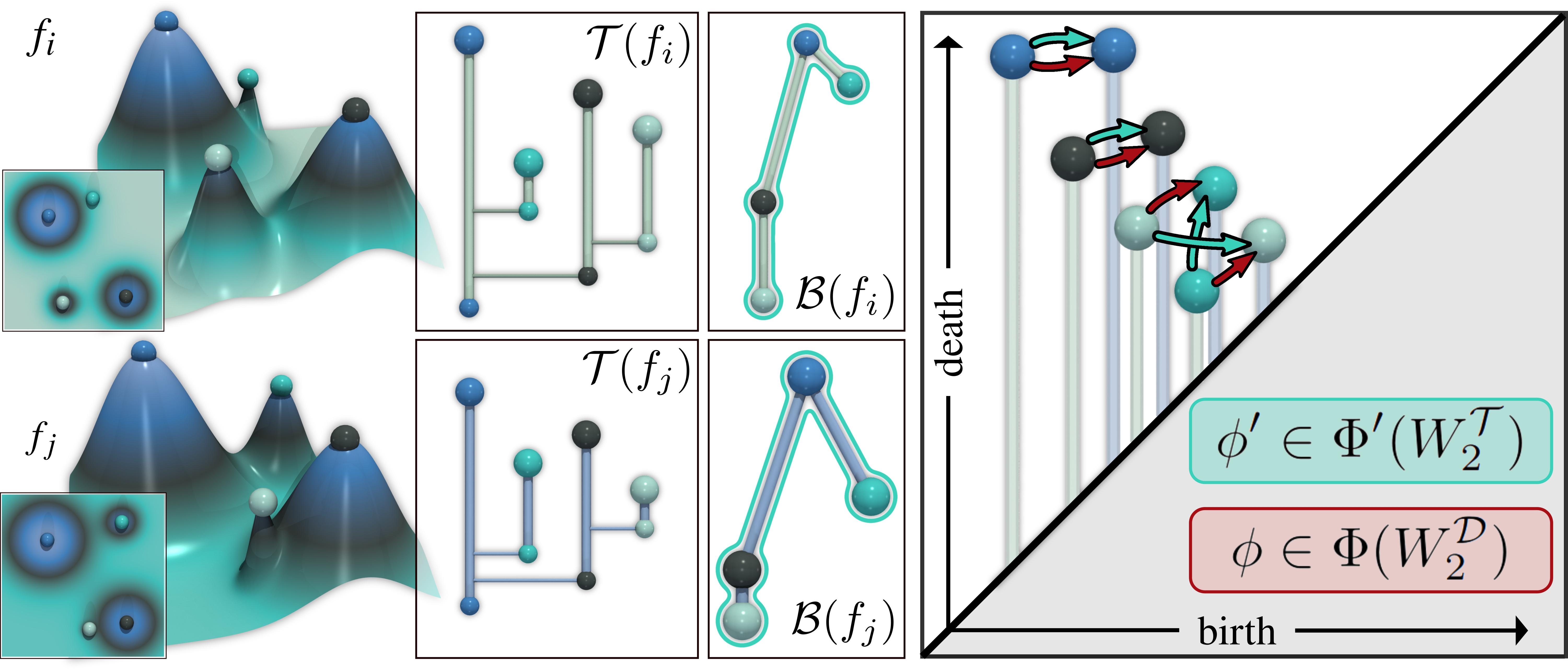}
\mycaption{
\journal{Illustration of the computation of the Wasserstein distance
$\wassersteinTree$ \cite{pont_vis21}
between the BDTs
$\branchtree(f_i)$ (white, top) and $\branchtree(f_j)$ (blue, bottom).
$\wassersteinTree$
is computed by solving an optimal assignment problem in the
birth/death plane (right), whose search space
is constrained to partial rooted isomorphims (the optimal solution is
represented with a cyan halo on the BDTs and with cyan arrows in the
birth/death plane).
When $\epsilon_1 = 1$, all the saddles are collapsed and the structure of the
BDT is completely ignored. In that case, $\wassersteinTree$ is equal to the
Wasserstein distance between persistence diagrams $\wasserstein{2}$ (the
optimal assignment is shown with red arrows, right). In this example,
$\wasserstein{2}$ reports a small distance whereas two hills have been swapped
in the datasets. Figure adapted from \cite{pont_tvcg23}.
}
%
}
\label{fig_wassersteinMetric}
\end{figure}

\subsection{Wasserstein metric space}
\label{sec_metricSpace}
\label{sec_wassersteinMetricSpace}
\journal{In this section, we first formalize the Wasserstein distance
between persistence diagrams \cite{edelsbrunner09}. Next, we recall its
generalization to merge trees \cite{pont_vis21}.}
\journal{This generalization enables our approach to support both
topological descriptors (persistence diagrams and merge trees).}
\journal{This
section includes elements adapted from \cite{pont_tvcg23}, which have
been considered  to make this manuscript self-contained.}

\journal{The evaluation of the distance between two diagrams $\diagram(f_i)$ and
$\diagram(f_j)$ is typically preceded by a pre-processing step aiming at
transforming the diagrams, such that they admit the same number of points,
which will facilitate the evaluation of their distance. This
procedure augments}
each
diagram with the diagonal projection of the off-diagonal points of the other
diagram:
\begin{eqnarray}
  \nonumber
 \diagram'(f_i) = \diagram(f_i) \cup \{
\projection(p_j) ~ | ~ p_j \in \diagram(f_j)
\}\\
\nonumber
  \diagram'(f_j) = \diagram(f_j) \cup \{
\projection(p_i) ~ | ~ p_i \in \diagram(f_i)
\},
\end{eqnarray}

\noindent
where $\projection(p_i) =
(\frac{x_i+y_i}{2},\frac{x_i+y_i}{2})$ is  the diagonal projection of
the off-diagonal point $p_i = (x_i, y_i)$.
\journal{Overall, this
augmentation procedure inserts dummy features in the diagram (with zero
persistence, on the diagonal), hence preserving the topological information
of the diagrams, while guaranteeing that the two diagrams now have the same number of points ($| \diagram'(f_i)| = |
\diagram'(f_j)|$).}


\journal{In order to compare two points $p_i = (x_i, y_i) \in
\diagram'(f_i)$ and $p_j = (x_j, y_j) \in \diagram'(f_j)$, a \emph{ground
distance} needs to be introduced in the birth/death plane. Specifically, we
consider the distance $\pointMetric_q$ ($q > 0$):}
%
\begin{eqnarray}
  \nonumber
\label{eq_pointMetric}
\pointMetric_{q}(p_i, p_j) = (|x_j - x_i|^q + |y_j - y_i|^q)^{1/q}
= \| p_i - p_j
\|_q.
\end{eqnarray}

\noindent
\journal{In the special case where both $p_i$ and $p_j$ are dummy features
located on the diagonal (i.e. $x_i = y_i$ and $x_j = y_j$), $\pointMetric_q(p_i,
p_j)$ is set to zero (such that these dummy features do not intervene in the
distance evaluation between the diagrams).}
%
 Then, the
$L^q$-Wasserstein
distance
$\wasserstein{q}$
\journal{can be
introduced as}:
\begin{eqnarray}
\label{eq_wasserstein}
\wasserstein{q}\big(\diagram(f_i), \diagram(f_j)\big)
=
\label{eq_wasserstein_mapping}
&
\hspace{-.15cm}
\underset{\phi \in \Phi}\min
 \hspace{-.15cm}
&
\Big(
\sum_{p_i \in \diagram'(f_i)}
\pointMetric_q\big(p_i,
\phi(p_i)\big)^q\Big)^{1/q},
\end{eqnarray}

\noindent
where $\Phi$ is the set of all possible assignments $\phi$ mapping a point $p_i
\in \diagram'(f_i)$ to a point $p_j \in \diagram'(f_j)$.
\journal{Note that it is possible that $\phi$ maps a point $p_i \in
\diagram'(f_i)$ to its diagonal projection (i.e. $\phi(p_i) =
\projection(p_i) = p_j \in \diagram'(f_j)$), which indicates the destruction of
the corresponding feature (or symmetrically, its appearance).}

\journal{Pont et al. recently generalized this metric to BDTs
\cite{pont_vis21}.
The expression of this distance, noted
$\wassersteinTree\big(\branchtree(f_i), \branchtree(f_j)\big)$, is the same as
\autoref{eq_wasserstein} (for $q = 2$), with the important difference of the
search space of possible assignments, noted $\Phi'
\subseteq \Phi$. Specifically, $\Phi'$ is constrained to the set of (rooted)
partial isomorphisms \cite{pont_vis21} between $\branchtree(f_i)$ and
$\branchtree(f_j)$
(cyan halo on the BDTs of \autoref{fig_wassersteinMetric}).}


\journal{This novel metric comes with a clear interpretation. The
control parameter $\epsilon_1$ (\autoref{sec_background_mergeTrees})
balances the importance of the BDT
structure in the distance. Specifically, when $\epsilon_1 = 1$, all saddles are
collapsed and we have $\wassersteinTree\big(\branchtree(f_i),
\branchtree(f_j)\big) =
\wasserstein{2}\big(\diagram(f_i), \diagram(f_j)\big)$. Generally speaking, as
illustrated experimentally by Pont et al. \cite{pont_vis21}, 
$\epsilon_1$ acts as a control knob balancing the practical stability of the metric with its
discriminative power.}
\journal{This generalized metric enables our framework to support both
topological descriptors. For persistence diagrams, we set $\epsilon_1 = 1$,
while for merge trees, we set it to the default recommended value (i.e.
$\epsilon_1 = 0.05$ \cite{pont_vis21}). In the following,
the metric space induced by this metric is noted $\branchtreeSpace$.}

For interpolation purposes, Pont et al. introduce a 
local normalization \cite{pont_vis21} in a pre-processing step (to 
guarantee the invertibility of any interpolated BDT into a valid MT). We will 
use the same procedure in this work to guarantee that the projected BDTs 
(\autoref{sec_lowLevelGeometricalTools}) indeed describe valid MTs.
Specifically, 
%
\journal{we normalize}
the persistence of each branch $b_i
\in \branchtree(f_i)$
with regard to that of its parent $b'_i \in
\branchtree(f_i)$, by \journal{moving}
$b_i$ in the birth/death
\journal{plane}
from $(x_i, y_i)$ to
$\normalizedLocation(b_i) = \big(\normalizedLocation_x(b_i),
\normalizedLocation_y(b_i)\big)$:
\begin{eqnarray}
\begin{array}{c}
\normalizedLocation_x(b_i) = (x_i - x'_i) / (y'_i - x'_i)\\
\vspace{-1.5ex}\\
\normalizedLocation_y(b_i) = (y_i - x'_i) / (y'_i - x'_i).
\end{array}
\label{eq_localNormalization}
\end{eqnarray}

\noindent
\journal{This pre-normalization procedure guarantees that any interpolated BDT
can indeed be reverted into a valid MT, by recursively applying
\autoref{eq_localNormalization}, which  explicitly enforces the
\emph{Elder rule} \cite{edelsbrunner09} on BDTs ($[x_i, y_i] \subseteq
[x'_i, y'_i]$, \autoref{sec_background_mergeTrees}), hence the validity of the
reconstructed MT. Two additional parameters were introduced by Pont et al.
\cite{pont_vis21} in order to control the effect of this pre-normalization
procedure  ($\epsilon_2$ balances the normalized persistence of small branches,
selected
via the threshold $\epsilon_3$). These parameters are set to their default
recommended values ($\epsilon_2 = 0.95$, $\epsilon_3 = 0.9$). In the following,
we consider that all the input BDTs are pre-normalized with this procedure.}

\section{Formulation}
\label{sec_formulation}

This section describes our novel extension of the classical auto-encoder neural
network architecture to the Wasserstein metric space of merge trees,
with the novel notion of \emph{Merge Tree Wasserstein Auto-Encoder}
(MT-WAE).
First, we describe a geometric
interpretation of the classical auto-encoders
(\autoref{sec_euclideanAutoEncoders}), which we
call
in the following
\emph{Euclidean Auto-Encoders} (EAE).
Next, we describe how to generalize each
geometrical tool used in EAE
(\autoref{sec_euclideanAutoEncoders}) to the Wasserstein metric space of
merge trees
(\autoref{sec_lowLevelGeometricalTools}).
Finally, once these tools are available, we formalize our notion of MT-WAE with
a novel neural network architecture (\autoref{sec_mtWAE}), for which we document
an optimization algorithm in \autoref{sec_algorithm}.


\subsection{\journal{An} interpretation of Euclidean Auto-Encoders}
\label{sec_euclideanAutoEncoders}

\begin{figure}
  \centering
\includegraphics[width=\linewidth]{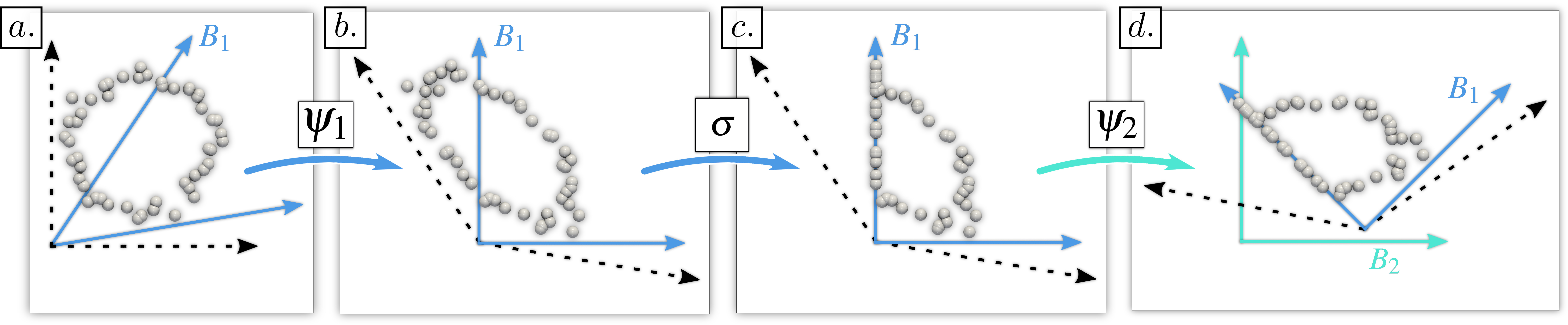}
  \caption{Geometric interpretation of Euclidean Auto-Encoders
(EAE, \autoref{sec_euclideanAutoEncoders}). In its simplest form (one
encoding and one decoding layer\journal{, $d'=2$}), an EAE can be
\journal{viewed}
as the composition
of a linear transformation $\linearTransformation_1$ $\big((a)$ to $(b)\big)$
defined with respect to a first basis $\pcaBasis_1$, followed by a
non-linearity $\activation$ $\big($here \emph{ReLU}, $(b)$ to $(c)\big)$,
followed by a second linear transformation $\linearTransformation_2$ $\big((c)$
to $(d)\big)$ defined with respect to a second basis $\pcaBasis_2$.
Specifically, both $\linearTransformation_1$ and $\linearTransformation_2$ are
optimized (via the optimization of $\pcaBasis_1$ and $\pcaBasis_2$) to minimize
the reconstruction error between the input $(a)$ and the output $(d)$. In the
case where $\activation$ is the identity, this reconstruction
optimization is equivalent to Principal Component Analysis \cite{bourlard88}.}
  \label{fig_sec3_geometricAutoEncoder}
\end{figure}

Let $P = \{p_1, p_2, \dots, p_\ensembleSize\}$ be a 
point set 
in a Euclidean
space $\mathbb{R}^d$ (\autoref{fig_sec3_geometricAutoEncoder}a).
The goal of Euclidean Auto-Encoders
(EAE) is to define a $d'$-dimensional
parameterization of $P$ (with $d' \leq d$) which describes well the data
(which 
enables its accurate \emph{reconstruction}).
Let $\pcaBasis_1 = \{\vectorNotation{b_1}, \vectorNotation{b_2},
\dots, \vectorNotation{b_{d'}}\}$ be a basis of linearly independent vectors in
$\mathbb{R}^{d}$ (\julien{\autoref{fig_sec3_geometricAutoEncoder}a}).
$\pcaBasis_1$ can
be
written in the form of a $d\times d'$ matrix, for which each
of the $d'$ columns is a vector of the basis.
Then, one can express the coordinates
$\linearTransformation_1(p_i) \in \mathbb{R}^{d'}$ of each point $p_i \in P$
with
the 
basis 
$\pcaBasis_1$:
\begin{eqnarray}
\label{eq_projectionMinimization}
\linearTransformation_1(p_i) = \argmin_{\alpha^i} ||p_i - \pcaBasis_1
\alpha^i||_2^2 + \vectorNotation{\origin_1},
\end{eqnarray}
where $\vectorNotation{\origin_1}$ is an \emph{offset} vector of
$\mathbb{R}^{d'}$, and where $\alpha^i \in \mathbb{R}^{d'}$ can be seen
as a set of
$d'$ coefficients, to apply on the $d'$ vectors of the basis
$\pcaBasis_1$ to best estimate $p_i$.
Note that \autoref{eq_projectionMinimization} can be re-written as a linear
transformation:
\begin{eqnarray}
  \label{eq_pseudoInverse}
  \linearTransformation_1(p_i) = \pcaBasis_1^{+}p_i +
\vectorNotation{\origin_1},
\end{eqnarray}
where $\pcaBasis_1^{+}$ is the pseudoinverse of the matrix $\pcaBasis_1$ 
(Figs. \ref{fig_sec3_geometricAutoEncoder}a-b).


Given this new parameterization
$\linearTransformation_1$, one can estimate a \emph{reconstruction} of the
point $p_i$ in $\mathbb{R}^{d}$, noted $\widehat{p_i}$.
For this, let us consider another, similar, linear transformation
$\linearTransformation_2$ (Figs.
\ref{fig_sec3_geometricAutoEncoder}c-d),
defined respectively to
a second basis
$\pcaBasis_2$ (given as a $d'\times d$ matrix) and a second offset vector
$\vectorNotation{\origin_2}$ (in $\mathbb{R}^d$). Then, the reconstruction
$\widehat{p_i}$ of each point $p_i$ is given by:
\begin{eqnarray}
  \nonumber
  \widehat{p_i} = \linearTransformation_2 \circ \linearTransformation_1 (p_i)
= \pcaBasis_2^{+}(\pcaBasis_1^{+}p_i + \vectorNotation{\origin_1}) +
\vectorNotation{\origin_2}.
\end{eqnarray}
To 
get
an accurate reconstruction $\widehat{p_i}$ 
(\autoref{fig_sec3_geometricAutoEncoder}d) for all the points
$p_i \in P$,
one needs to optimize both $\linearTransformation_1$ and
$\linearTransformation_2$, to minimize the 
\journal{following
data fitting
energy:}
\begin{eqnarray}
\label{eq_dataFitting}
\reconstructionError = \sum_{i = 1}^N || p_i - \widehat{p_i}||_2^2 =
\sum_{i = 1}^N || p_i - \linearTransformation_2 \circ \linearTransformation_1
(p_i)||_2^2.
\end{eqnarray}

\begin{figure}
  \centering
  \includegraphics[width=\linewidth]{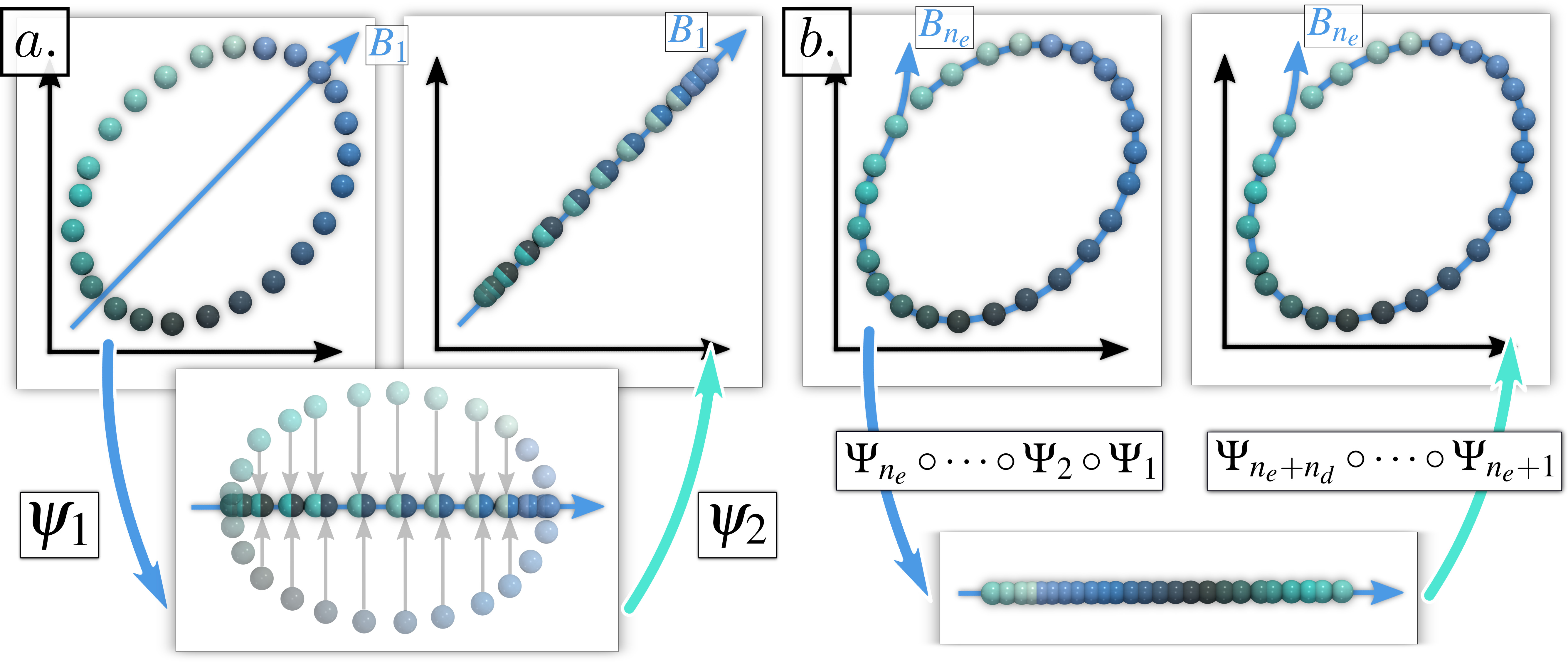}
  \caption{Comparison between PCA $(a)$ and EAE $(b)$ for the $1$-dimensional
encoding
of a 2D point set sampling a $1$-manifold (color: rotation
angle). In its latent space $\big((a)$, bottom$\big)$, PCA linearly projects
the input points to a line,
hence interleaving points from
the upper and lower parts of the circle. This results in a poor reconstruction
$\big((a)$, right$\big)$, where points are interleaved along the axis $B_1$.
In contrast, EAE optimizes a composition of
non-linear transformations, which consistently \emph{unwraps} the circle onto a
line in its latent space $\big((b)$, bottom$\big)$, while nicely preserving the
intrinsic parameterization of the circle (rotation angle). This results in
an accurate reconstruction $\big((b)$, right$\big)$\journal{:}
the
embedding of the
axis $B_{\sizeEncoding}$ in the
data defines a faithful $1$-dimensional
parameterization of the circle.
}
  \label{fig_pca_vs_eae}
\end{figure}

As discussed by Bourlard and Kamp \cite{bourlard88},
this
formulation is a generalization of
Principal Component Analysis (PCA) \cite{pearson01},
a seminal statistical tool for variability analysis.
However,
PCA
assumes that the input point cloud can be faithfully
approximated via linear projections.
As shown in
\autoref{fig_pca_vs_eae}, this hypothesis can be easily challenged in practice.
This motivates a non-linear generalization of PCA, capable of faithfully
approximating point clouds exhibiting non-linear structures
(\autoref{fig_pca_vs_eae}).

Specifically, to introduce non-linearity, the above linear transformations
$\linearTransformation$
are
typically composed with a non-linear function $\activation$,
called \emph{activation function},
such that
the transformation of each point $p_i$, noted $\unitTransformation(p_i)$,
is now given by:
$\unitTransformation(p_i) = \activation\big(\linearTransformation(p_i)\big)$.
For example, the
rectifier activation function (\emph{``ReLU''})
will take
the $j^{th}$ coordinate of each data point (i.e.
$\big(\linearTransformation(p_i)\big)_j$) and \emph{snap} it to zero if it is
negative (\autoref{fig_sec3_geometricAutoEncoder}b-c).
%
We call the above non-linear transformation $\unitTransformation$ a
\emph{transformation layer}. It is characterized by its own vector basis
$\pcaBasis$
and its own offset vector $\origin$.

Next, to faithfully approximate complicated non-linear
input distributions, the above transformation layer is typically composed with a
number ($\sizeEncoding + \sizeDecoding$) of other transformation layers, defined
similarly. Then, the
initial data fitting energy (\autoref{eq_dataFitting}) can now be
generalized into:
\begin{eqnarray}
\label{eq_eAE}
\reconstructionError = \sum_{i = 1}^\ensembleSize ||p_i -
\unitTransformation_{\sizeEncoding + \sizeDecoding}
\circ
\dots
\circ
\unitTransformation_{\sizeEncoding+1}
\circ
\unitTransformation_{\sizeEncoding}
\circ \dots \circ \unitTransformation_2 \circ \unitTransformation_1 (p_i)
||_2^2,
\end{eqnarray}
where each
transformation layer $\unitTransformation_{k}$ is associated
with its own $d_k$-dimensional vector basis
$\pcaBasis_k$ and its  offset vector $\origin_k$.
%
Then, the notion of
\emph{Auto-Encoder} is a specific instance of the
above formulation, with:
\begin{enumerate}
 \item $d_1 > d_2 > \dots > d_{\sizeEncoding}$, and
 \item $d_{\sizeEncoding} < d_{\sizeEncoding + 1} < \dots < d_{\sizeEncoding
 + \sizeDecoding} = d$, and
 \item $\activation_{\sizeEncoding+\sizeDecoding}$ is the identity.
\end{enumerate}
Specifically, $\sizeEncoding$ and $\sizeDecoding$ respectively denote
the number of \emph{Encoding} and \emph{Decoding} transformation layers, while
$d_{\sizeEncoding}$ is the dimension of the so-called \emph{latent space}.
In practice, $d_{\sizeEncoding}$ is typically chosen to be much smaller
than the input dimensionality
($d$), for non-linear dimensionality
reduction purposes (each input point $p_i$ is then represented in
$d_{\sizeEncoding}$ dimensions, according to its
coordinates in
$\pcaBasis_{\sizeEncoding}$, noted  $\alpha_{\sizeEncoding}^i \in
\mathbb{R}^{d_{\sizeEncoding}}$).

\autoref{eq_eAE} defines an optimization problem whose
variables (the $\sizeEncoding + \sizeDecoding$ bases and offset vectors) can be
efficiently optimized (e.g. with gradient descent \cite{KingmaB14})
by composing the transformation layers within a neural network.
Then, the
gradient $\nabla \reconstructionError$ of $\reconstructionError$ can be
estimated by exploiting the
automatic differentiation capabilities of modern neural network implementations
\cite{pytorch},
themselves based on the application of the chain-rule derivation on the above
composition of transformation layers.

\begin{figure}
  \centering
  \includegraphics[width=\linewidth]{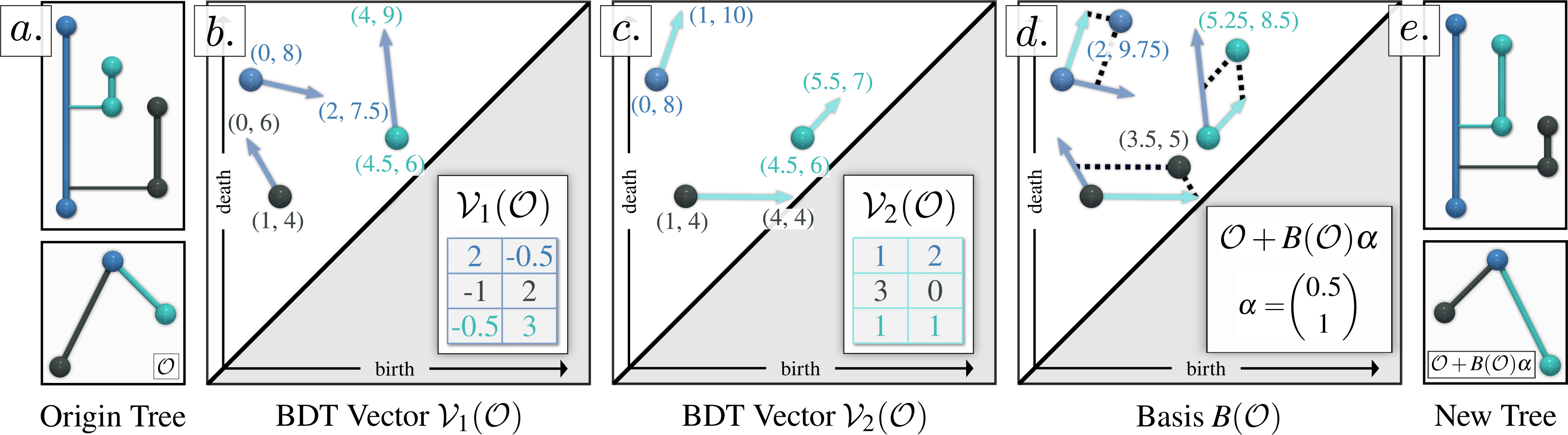}
  \caption{\minor{Low-level geometrical tools in the Wasserstein metric space of merge
trees (\autoref{sec_lowLevelGeometricalTools}). Given an origin MT
and its BDT $\bdtOrigin$ $\big($counting $|\bdtOrigin|$ nodes $(a)\big)$, a 
BDT vector $\directionVector_1(\bdtOrigin)$ is
defined in the birth/death space as a concatenation of $|\bdtOrigin|$ 2D
vectors $\big($blue arrows $(b)\big)$. Given a second BDT vector
$\directionVector_2(\bdtOrigin)$ $\big($cyan arrows $(c)\big)$, a
 basis $\pcaBasis(\bdtOrigin)$ can be defined $\big((d)$, here with $d' =
2\big)$. For a given set of coefficients $\alpha \in \mathbb{R}^{d'}$, a new
merge tree and its BDT $(e)$ can be reconstructed by applying a sum of 2D
displacements $\sum_{j = 1}^{j = d'} \alpha_j
\big(\directionVector_j(\bdtOrigin)\big)_i$ to each branch $b_i$ of $\bdtOrigin$
$\big((d)$, black dashes$\big)$.
  }}
  \label{fig_sec3_geometryWassersteinSpace}
\end{figure}

\subsection{From EAE to MT-WAE}
\label{sec_lowLevelGeometricalTools}
\begin{figure*}
  \centering
  \includegraphics[width=\linewidth]{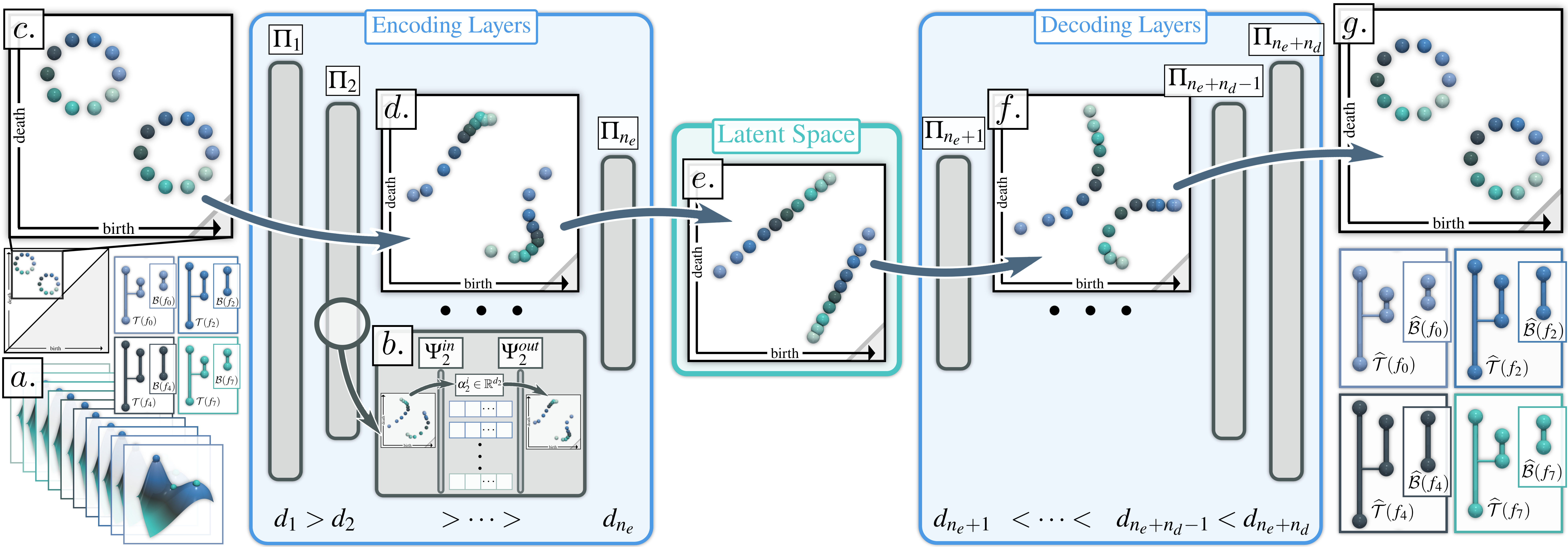}
  \caption{\textbf{Overview:}
given an input ensemble, with its
merge trees and BDTs $(a)$,
our Wasserstein Auto-Encoder of Merge Trees (MT-WAE)
optimizes an auto-encoder 
(for this example, $\sizeEncoding = \sizeDecoding = 2$)
where each layer $\Pi_k$
natively processes BDTs (without pre-vectorization).
Specifically, each layer $\Pi_k$ can be interpreted as a local auto-encoder
$(b)$, where an \emph{input} sub-layer $\unitTransformation_k^{in}$ transforms
the input BDT $\branchtree_{k-1}(f_i)$ into a set of coefficients $\alpha_k^i
\in \mathbb{R}^{d_k}$ and where an \emph{output} sub-layer
$\unitTransformation_k^{out}$ transforms these coefficients back into a valid
BDT $\branchtree_{k}(f_i)$.
The aggregated views
$\big((c), (d), (e), (f), (g)\big)$, which overlap all the BDTs in the
birth/death
space (one color per BDT), show the ability of MT-WAE to progressively
\emph{unwrap} non-linear structures (circles) as the BDTs progress down
the network,
%
resulting in faithful local parameterizations in
latent space
$\big((e)$, the individual angular parameterizations of the circles are well
preserved$\big)$, as well as accurate reconstructions $(g)$.
This native support of BDTs results in a superior accuracy
(\autoref{sec_quality})
and
an improved interpretability: individual
features can now be tracked as they traverse the network,
\journal{enabling}
new
visual analysis capabilities
(\autoref{sec_dimensionalityReduction}).
}
  \label{fig_overview}
\end{figure*}

When the input data is not given as a point cloud in a Euclidean space
(\autoref{sec_euclideanAutoEncoders}) but as an abstract set equipped with a
metric, the
above EAE formulation
needs to be
extended.
%
For this, we
redefine in this section
the low-level
geometrical tools used in EAE (\autoref{sec_euclideanAutoEncoders}), but within
the context of the Wasserstein metric space
$\branchspace$ \cite{pont_vis21}.
In particular,
we formalize the following
notions:
\begin{enumerate}
 \item BDT vector \journal{(\autoref{fig_sec3_geometryWassersteinSpace}b)};
 \item BDT basis \journal{(\autoref{fig_sec3_geometryWassersteinSpace}d)};
 \item BDT basis projection
\journal{(\autoref{fig_sec3_geometryWassersteinSpace}d)};
 \item BDT transformation layer (\autoref{fig_overview}b).
\end{enumerate}

\noindent
\textbf{\emph{(1)} BDT vector:} Given a BDT $\branchtree$ with $|\branchtree|$
branches, a \emph{BDT vector} $\directionVector(\branchtree) \in \branchspace$
is a vector in
$\mathbb{R}^{2 |\branchtree|}$, which maps each branch $b \in
\branchtree$ to a new location in the 2D birth/death space.
$\branchtree$ is
the 
origin
of $\directionVector(\branchtree)$.
\journal{This is illustrated for example in
\autoref{fig_sec3_geometryWassersteinSpace}b), where the branches of a given
merge tree $\big($\autoref{fig_sec3_geometryWassersteinSpace}a$\big)$ are
displaced in the birth-death plane (light blue vectors from the spheres of
matching color).}

\noindent
\textbf{\emph{(2)} BDT basis:}
Given an origin BDT $\bdtOrigin$,
a $d'$-dimensional basis of BDT vectors, noted
$\pcaBasis(\bdtOrigin)$, is a set $\{\directionVector_1(\bdtOrigin),
\directionVector_2(\bdtOrigin), \dots, \directionVector_{d'}(\bdtOrigin)\}$ of
$d'$ linearly independent BDT vectors, having for common origin $\bdtOrigin$.
\journal{This is shown in \autoref{fig_sec3_geometryWassersteinSpace}d),
where two BDT vectors (from \autoref{fig_sec3_geometryWassersteinSpace}b),
and \autoref{fig_sec3_geometryWassersteinSpace}c), blue and green
arrows) are combined into a basis. \autoref{sec_initialization} clarifies the
basis initialization by our approach.}

\noindent
\textbf{\emph{(3)} BDT basis projection:} 
\journal{Given an arbitrary BDT $\branchtree$, its \emph{projection error}
$e(\branchtree)$ in a $d'$-dimensional BDT basis $\pcaBasis(\bdtOrigin)$ is:}
%
\begin{eqnarray}
  \label{eq_bdtProjectionError}
  e(\branchtree) = 
\min_{\alpha}\Big(\wassersteinTree\big(\branchtree,\bdtOrigin +
\pcaBasis(\bdtOrigin)\alpha    \big)\Big)^2,
\end{eqnarray}
where $\alpha \in \mathbb{R}^{d'}$ can be
interpreted as a set of coefficients 
 to apply on the $d'$ BDT vectors 
of
$\pcaBasis(\bdtOrigin)$
to best
estimate $\branchtree$.
Then, the projection of $\branchtree$ in $\pcaBasis(\bdtOrigin)$, noted 
$\linearTransformation(\branchtree)$, is given by the optimal coefficients 
$\alpha$ associated to the projection 
error of $\branchtree$
(\autoref{eq_bdtProjectionError}):
\begin{eqnarray}
  \label{eq_bdtProjection}
  \linearTransformation(\branchtree) = 
  \argmin_{\alpha}
\Big(\wassersteinTree\big(\branchtree,\bdtOrigin +
\pcaBasis(\bdtOrigin)\alpha    \big)\Big)^2.
\end{eqnarray}
%
%
%
%
The above equation is a re-interpretation of
\autoref{eq_projectionMinimization} (\autoref{sec_euclideanAutoEncoders}),
where the $L_2$ norm (used for EAE) is replaced by the Wasserstein
distance
$\wassersteinTree$. 
\journal{This projection procedure is illustrated in
\autoref{fig_sec3_geometryWassersteinSpace}d).
Given a BDT basis $\pcaBasis(\bdtOrigin)$ (blue and green arrows), the
projection of an arbitrary BDT $\branchtree$ is the linear combination
$\linearTransformation(\branchtree)$ of the BDT vectors of the basis which
minimizes its Wasserstein distance $\wassersteinTree$ to $\branchtree$.
In the birth-death space, the branches of the basis origin $\bdtOrigin$ are
represented by the colored spheres at the intersection between the blue and green
arrows, while the branches of $\linearTransformation(\branchtree)$ are
represented by the other spheres of matching color. In this example, the linear
combination $\linearTransformation(\branchtree)$ which
minimizes its
distance to $\branchtree$ is obtained for the coefficients $\alpha = (0.5, 1)$.
Then, to go from $\bdtOrigin$ to $\linearTransformation(\branchtree)$, each
branch $b_i$ of $\bdtOrigin$ is displaced by $0.5 ~
\directionVector_1(\bdtOrigin)$ (intersection between the dashed lines and the
blue arrows) and then by $1~ \directionVector_2(\bdtOrigin)$ (intersection
between the dashed lines and the green arrows). The resulting merge tree is
shown in \autoref{fig_sec3_geometryWassersteinSpace}e). In contrast to the
merge tree of the basis origin
$\big($\autoref{fig_sec3_geometryWassersteinSpace}a)$\big)$, the persistence of
the cyan branch has increased while that of the black branch has decreased.}


\noindent
\textbf{\emph{(4)} BDT transformation layer:} Once the above tools have been
formalized, we can introduce the novel notion of 
\emph{BDT transformation layer}.
Similarly to the Euclidean case (\autoref{sec_euclideanAutoEncoders}),
a non-linear activation function $\activation$ (in our case,
\emph{Leaky ReLU}) can
be
composed with the above projection, yielding the new function
$\unitTransformation(\branchtree)
= \activation \big(\psi(\branchtree)\big)$. Note that, at this stage,
$\unitTransformation(\branchtree)$ can be interpreted as a set of coefficients
to apply on the
BDT basis $\pcaBasis(\bdtOrigin)$ to best estimate $\branchtree$. In other
words, $\unitTransformation(\branchtree)$ is not a BDT yet, but simply a set of
coefficients,
which can be used later to reconstruct a BDT. Thus, 
a second transformation
needs to be considered, to transform the set of coefficients
$\unitTransformation(\branchtree)$ back into a BDT. Then, we define the notion 
of
\emph{BDT
transformation layer}, noted $\Pi(\branchtree)$, as the composition
$\Pi(\branchtree) = \unitTransformation^{out} \circ \unitTransformation^{in}
(\branchtree)$ (\autoref{fig_overview}b):
\begin{eqnarray}
\nonumber
\begin{array}{l}
  \unitTransformation^{in}(\branchtree) =
  \activation \bigg(\argmin_{\alpha}
\Big(\wassersteinTree\big(\branchtree,
\bdtOrigin^{in} +
\pcaBasis^{in}(\bdtOrigin^{in})\alpha    \big)\Big)^2\bigg) \\
  \unitTransformation^{out}(\alpha) = \validBDT\big(\bdtOrigin^{out} +
\pcaBasis^{out}(\bdtOrigin^{out})\alpha \big),
\end{array}
\end{eqnarray}
\begin{wrapfigure}{r}{3cm}
  \begin{center}
\vspace{-1cm}
\includegraphics[width=\linewidth]{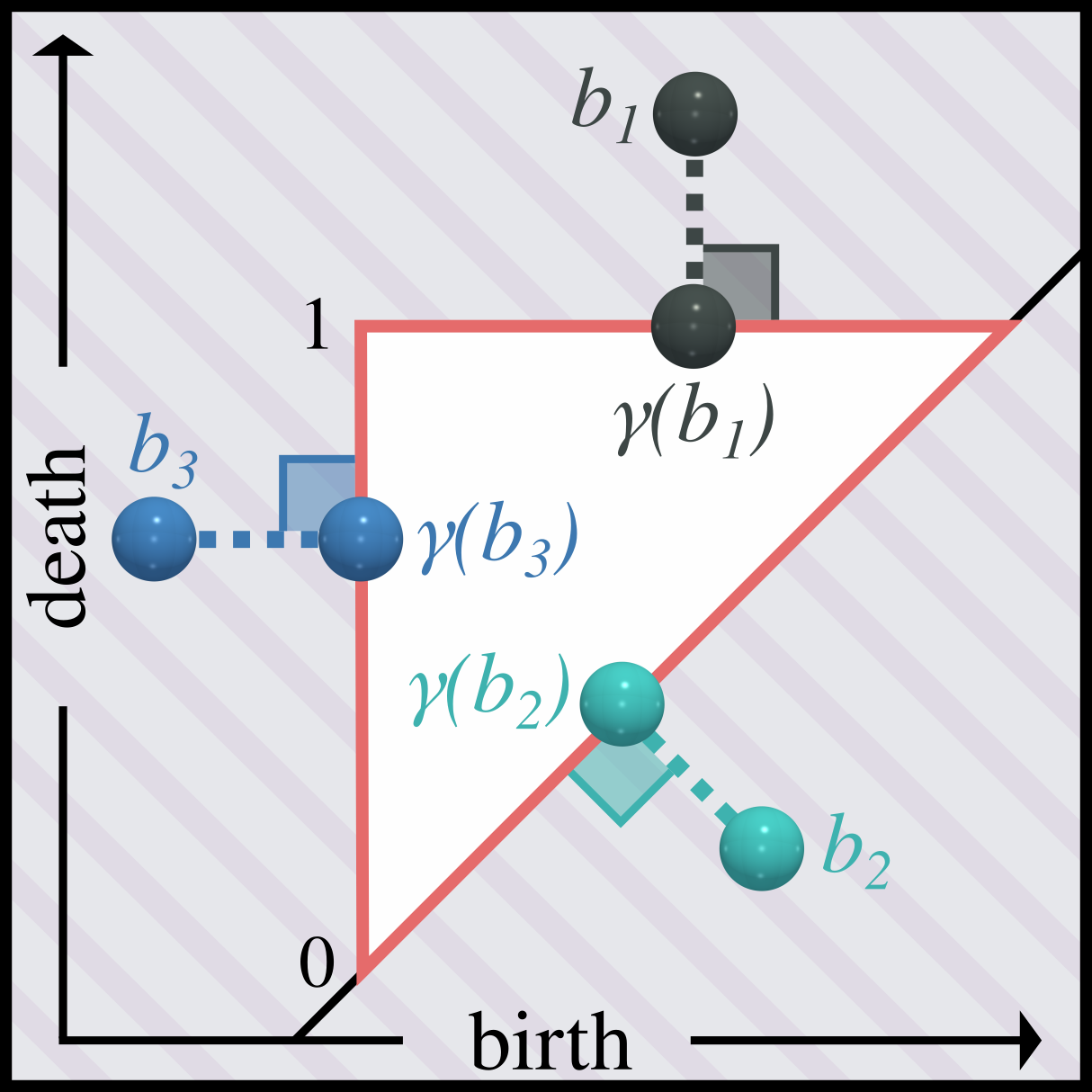}
\vspace{-0.7cm}
\caption{\minor{Projection $\validBDT$ ensuring the Elder rule.}}
\label{fig_projection}
\vspace{-0.5cm}
\end{center}
\end{wrapfigure}
where $\validBDT(\branchtree)$ is a  projection which transforms
$\branchtree$
into a \emph{valid} BDT, i.e.
%
which respects the
\emph{Elder rule} (\autoref{sec_mergeTrees}).
Given a branch
$b \in \branchtree$,
$\validBDT$
enforces that: $\gamma(b)_x < \gamma(b)_y$ and $[\gamma(b)_x, \gamma(b)_y]
\subseteq [0, 1]$ (\minor{\autoref{fig_projection}}).

\emph{BDT transformation layers} can be seen as
\emph{local} auto-encoders (\autoref{fig_overview}b): the first step
$\unitTransformation^{in}$ converts
an
input BDT into a
set of coefficients with a basis projection
and
a non-linearity,
while the second step $\unitTransformation^{out}$ converts these
coefficients into a  BDT.
Note that each
BDT
transformation layer is associated with its own
 input \emph{and} output  $d'$-dimensional bases
$\pcaBasis^{in}(\bdtOrigin^{in})$ and $\pcaBasis^{out}(\bdtOrigin^{out})$.

\journal{The processing of an ensemble of BDTs by a BDT transformation
layer is illustrated in \autoref{fig_overview}. Specifically,
\autoref{fig_overview}c) shows a zoom of the birth-death space, where all the
BDTs have been aggregated (one color per BDT, each BDT has two branches, hence
two patterns appear, one circle per branch).
The left inset of \autoref{fig_overview}b) shows the non-linearly
transformed set of BDTs after the first BDT transformation layer,
$\Pi_1$. Next, the first step $\unitTransformation^{in}_2$ of the next BDT
layer $\Pi_2$ converts each BDT
$\Pi_1\big(\branchtree(f_i)\big)$ into a set of coefficients $\alpha_2^i$.
Finally, the second step $\unitTransformation^{out}_2$ of the layer $\Pi_2$
converts each of these set of coefficients into a new, non-linearly
transformed BDT $\Pi_2 \circ \Pi_1\big(\branchtree(f_i)\big)$
$\big($\autoref{fig_overview}b), right inset$\big)$}.

%
%


\subsection{MT-WAE formulation}
\label{sec_mtWAE}
Now that the above
geometrical tools
have been introduced for the
Wasserstein metric space of merge trees,
we can now formulate
MT-WAE
by direct analogy to the
Euclidean case (\autoref{sec_euclideanAutoEncoders}).
Given a set $\branchtreeSet = \{\branchtree(f_1), \dots, \branchtree(f_N)\}$ of 
input BDTs, a
MT-WAE
is
a composition of BDT
transformation layers $\Pi_k(\branchtree)$ (\autoref{fig_overview}), 
minimizing
\journal{the following}
energy:
\begin{equation}
\label{eq_mtWaeEnergy}
\mtPgaError = \sum_{i = 1}^N \bigg(\wassersteinTree\Big(\branchtree(f_i), 
\Pi_{\sizeEncoding+\sizeDecoding} \circ \dots 
\circ
\Pi_{\sizeEncoding} \circ 
\dots \circ \Pi_1\big(\branchtree(f_i)\big)
\Big)\bigg)^2,
\end{equation}
where each BDT transformation layer $\Pi_k$ is associated with its own 
$d_k$-dimensional input \emph{and} output vector bases 
$\pcaBasis^{in}_k(\bdtOrigin_k^{in})$ and 
$\pcaBasis^{out}_k(\bdtOrigin_k^{out})$. Moreover, the dimensions of the 
successive bases are chosen such that:
\begin{enumerate}
 \item $d_1 > d_2 > \dots > d_{\sizeEncoding}$, and
 \item $d_{\sizeEncoding} < d_{\sizeEncoding + 1} < \dots < 
d_{\sizeEncoding+\sizeDecoding}$,
\end{enumerate}
where $\sizeEncoding$ and $\sizeDecoding$  denote the number of 
\emph{Encoding} and \emph{Decoding}  
layers and where 
$d_{\sizeEncoding}$ is the dimension of the MT-WAE latent space.
\autoref{eq_mtWaeEnergy} is a direct analog to the classical EAE
(\autoref{eq_eAE}): the standard transformation layers $\unitTransformation_k$
have been
replaced by BDT transformation layers $\Pi_k$ and the $L_2$ norm by the 
distance $\wassersteinTree$.

\journal{\autoref{fig_overview} illustrates a network of BDT transformation
layers optimized on a
synthetic ensemble (our optimization
algorithm is described in \autoref{sec_algorithm}). As mentioned in the
previous section, each input BDT (spheres in the aggregated birth-death views,
one color per BDT) is non-linearly transformed by the BDT transformation layers
$\big($see for instance \autoref{fig_overview}d) and
\autoref{fig_overview}f)$\big)$. As a result, in this example, the BDT
transformation layers
progressively \emph{unwrap} the non-linear
structures in the birth-death plane $\big($circles in
\autoref{fig_overview}c)$\big)$ as the BDTs traverse the network down to the
latent space $\big($\autoref{fig_overview}e)$\big)$, where the resulting layout
in the birth-death plane (line segments) manages to faithfully encode the
purposely designed
parameterization of the ensemble: the order of rotation angles
(colors) is well preserved along the segments in latent space.}

\section{Algorithm}
\label{sec_algorithm}
This section presents our
algorithm for the minimization of
\autoref{eq_mtWaeEnergy}.



\begin{algorithm}[t]
 \fontsize{5.625}{9}\selectfont
 \algsetup{linenosize=\tiny}
    \caption{Wasserstein Auto-Encoder (algorithm overview).}
    \label{alg_overview}

    \hspace*{\algorithmicindent} \textbf{Input}: Set of BDTs $\branchtreeSet =
\{\branchtree(f_1),
\dots, \branchtree(f_\ensembleSize)\}$.

    \hspace*{\algorithmicindent} \textbf{Output1}: Set of $(\sizeEncoding +
\sizeDecoding)$ input origins $\theta_{\bdtOrigin^{in}} =
\{\bdtOrigin^{in}_1, \bdtOrigin^{in}_2, \dots, \bdtOrigin^{in}_{\sizeEncoding +
\sizeDecoding}\}$;

\hspace*{\algorithmicindent} \textbf{Output2}: Set of $(\sizeEncoding +
\sizeDecoding)$ input bases $\theta_{\pcaBasis^{in}} =
\{\pcaBasis^{in}_1(\bdtOrigin^{in}_1),
\pcaBasis^{in}_2(\bdtOrigin^{in}_2), \dots, \pcaBasis^{in}_{\sizeEncoding +
\sizeDecoding}(\bdtOrigin^{in}_{\sizeEncoding + \sizeDecoding})\}$;

\hspace*{\algorithmicindent} \textbf{Output3}: For each of the input BDTs
($i \in \{1, 2, \dots, \ensembleSize\}$),
set of $(\sizeEncoding +
\sizeDecoding)$ input coefficients:\\
\hspace*{\algorithmicindent}
\hspace*{\algorithmicindent}
$\theta_{\alpha}^i = \{\alpha_1^i \in
\mathbb{R}^{d_1}, \alpha_2^i \in
\mathbb{R}^{d_2}, \dots, \alpha_{\sizeEncoding + \sizeDecoding}^i \in
\mathbb{R}^{d_{\sizeEncoding + \sizeDecoding}}\}$;

\hspace*{\algorithmicindent} \textbf{Output4}: Set of $(\sizeEncoding +
\sizeDecoding)$ output origins $\theta_{\bdtOrigin^{out}} =
\{\bdtOrigin^{out}_1,
\bdtOrigin^{out}_2, \dots, \bdtOrigin^{out}_{\sizeEncoding +
\sizeDecoding}\}$;

\hspace*{\algorithmicindent} \textbf{Output5}: Set of $(\sizeEncoding +
\sizeDecoding)$ output bases $\theta_{\pcaBasis^{out}} =
\{\pcaBasis^{out}_1(\bdtOrigin^{out}_1),
\pcaBasis^{out}_2(\bdtOrigin^{out}_2), \dots, \pcaBasis^{out}_{\sizeEncoding +
\sizeDecoding}(\bdtOrigin^{out}_{\sizeEncoding + \sizeDecoding})\}$;

\hrulefill

\begin{algorithmic}[1]

\STATE $\theta \leftarrow \{\theta_{\bdtOrigin^{in}},
\theta_{\pcaBasis^{in}}, \theta_{\bdtOrigin^{out}}, \theta_{\pcaBasis^{out}}\};$
\COMMENT{Overall set of optimization variables.}
\label{line_overallVariables}
\STATE Initialize$(\theta);$
\label{line_initialization}
\COMMENT{Initialization of the optimization variables
(\autoref{sec_initialization}).}
\WHILE{$\mtPgaError(\theta)$ decreases}
    \STATE $\widehat{\branchtreeSet} \leftarrow$ Forward$(\branchtreeSet,
\theta);$ \COMMENT{Forward propagation of the BDT ensemble $\branchtreeSet$
\label{line_forward}
(\autoref{sec_forward}).}
    \STATE $\theta \leftarrow$ Backward$(\branchtreeSet,
\widehat{\branchtreeSet});$ \COMMENT{Backward propagation
(\autoref{sec_backward}).}
\label{line_backward}
\ENDWHILE
\end{algorithmic}
\end{algorithm}

\subsection{Overview}
\autoref{alg_overview} provides an overview of our main algorithm.
The set of optimization variables, noted $\theta$ \minor{and declared line \ref{line_overallVariables}}, includes the
$(\sizeEncoding + \sizeDecoding)$ input \emph{and} output BDT bases, along with
their origins.
The optimization of
these variables follows the standard overall procedure for the optimization of
a neural network. 

\noindent
\minor{\textbf{\emph{(1)} Initialization:}}
First, $\theta$ is initialized\minor{, as detailed in \autoref{sec_initialization}.}

\noindent
\minor{\textbf{\emph{(2)} Forward propagation:}}
Then, the input ensemble of BDTs $\branchtreeSet$\minor{, illustrated in \autoref{fig_overview}a,}
traverses the network to generate a reconstructed ensemble of BDTs, noted
$\widehat{\branchtreeSet}$.
\minor{This occurs line \ref{line_forward} of \autoref{alg_overview} and it is illustrated from \autoref{fig_overview}c to \autoref{fig_overview}g with the dark blue arrows.}
This is traditionally
denoted as the \emph{Forward} propagation.
\minor{This part of our approach is documented in \autoref{sec_forward}.}

\noindent
\minor{\textbf{\emph{(3)} Backward propagation:}}
Given
$\widehat{\branchtreeSet}$, the energy
$\mtPgaError(\theta)$\minor{, detailed in \autoref{eq_mtWaeEnergy},}
can be evaluated and its
gradient $\nabla \mtPgaError(\theta)$ can be estimated by automatic
differentiation\minor{,} based on the application of the chain-rule on the composition
of BDT transformation layers. Given the gradient $\nabla
\mtPgaError(\theta)$, a step of gradient descent can be achieved to update the
optimization variables $\theta$.
\minor{This occurs line \ref{line_backward} of \autoref{alg_overview}.}
This is
traditionally denoted as the \emph{Backward} propagation\minor{. This part of our approach is detailed in \autoref{sec_backward}.}

\noindent
\minor{\textbf{\emph{(4)} Stopping condition:}}
\minor{The two steps of forward and backward propagations}
are then iterated until the energy stops decreasing\minor{,} in practice until it
decreases by less than $1\%$ between two iterations.

\subsection{Basis projection}
\label{sec_basisProjection}
We start by describing an efficient
\emph{Assignment/Update}
algorithm
for
the projection  of a BDT $\branchtree$ into a
BDT basis
$\pcaBasis(\bdtOrigin)$ (\autoref{eq_bdtProjection}), as it is a core
geometrical component used throughout our approach.


The purpose of the projection
(\autoref{eq_bdtProjection}) is to find a
set of coefficients $\alpha \in \mathbb{R}^{d'}$ to apply on the $d'$ BDT
vectors of $\pcaBasis(\bdtOrigin)$ to best estimate $\branchtree$.
The
geodesic analysis of merge trees \cite{pont_tvcg23} faces
a similar issue, but its
formulation (restricting
$\alpha$ to $[0, 1]^{d'}$) allows for an iterative,
brute-force
optimization.
Here, we introduce a more general and
efficient strategy.

\noindent
\textbf{\emph{(1)} Assignment step:} Let us  assume that we are given an
initial set of coefficients $\alpha$.
Then,
the \emph{estimation}
$\widehat{\branchtree}$ of $\branchtree$ is given by
$\widehat{\branchtree} \leftarrow \bdtOrigin + \pcaBasis(\bdtOrigin)\alpha$.
The purpose of the assignment step is to
refine the evaluation of
$\wassersteinTree(\branchtree, \widehat{\branchtree})$.
For this,
we first compute the optimal assignment\footnote{This
discussion describes the case where $\phi_*$ is a bijection
between off-diagonal points of the 2D birth/death plane. \julien{Appendix B}
details
the general case, where $\phi_*$
may send
points of $\branchtree$ to the diagonal of $\widehat{\branchtree}$, and
reciprocally.
} $\phi_*$
between $\branchtree$ and
$\widehat{\branchtree}$, w.r.t. \autoref{eq_wasserstein}.
Then,
$\wassersteinTree(\branchtree,
\widehat{\branchtree})$ can be re-written as:
\begin{eqnarray}
  \label{eq_wasserstein_constantAssignment}
  \wassersteinTree(\branchtree,
\widehat{\branchtree}) = \sum_{i = 1}^{|\branchtree|} || b_i -
\phi_*(b_i)||_2^2.
\end{eqnarray}

\noindent
\textbf{\emph{(2)} Update step:} Given the above estimation
$\widehat{\branchtree}$, the goal of the update step is to improve
the coefficients $\alpha$, in order to decrease
$\wassersteinTree(\branchtree,
\widehat{\branchtree})$.
Let $\widehat{\branchtree'}$ be a vector representation of
$\widehat{\branchtree}$. Specifically, $\widehat{\branchtree'}$ is a vector in
$\mathbb{R}^{2  |\widehat{\branchtree}|}$ which concatenates the
coordinates in the 2D birth/death plane of each branch $b_i$ of
$\widehat{\branchtree}$.
$\widehat{\branchtree'}$  can be decomposed into $\bdtOrigin' +
\big(\pcaBasis(\bdtOrigin')\big)'\alpha$, where
$\big(\pcaBasis(\bdtOrigin')\big)'$ is a $(2|\widehat{\branchtree}|)\times d'$
matrix.
Additionally, let $\branchtree'$ be a similar vector
representation of $\branchtree$, \emph{but} where the entries have been
specifically re-ordered such that, for each of its 2D entries, we have:
\begin{eqnarray}
  \label{eq_vectorReOrdering}
  (\branchtree')_i = \phi_*^{-1}\big((\widehat{\branchtree'})_i\big).
\end{eqnarray}
Intuitively, $\branchtree'$  
is a 
re-ordered vector representation of $\branchtree$,
such that its $i^{th}$ entry exactly matches though $\phi_*$ with the $i^{th}$ 
entry of $\widehat{\branchtree'}$. 
Given this vector representation, the Wasserstein distance
$\wassersteinTree(\branchtree,
\widehat{\branchtree})$ for a fixed optimal assignment  $\phi^*$
(\autoref{eq_wasserstein_constantAssignment}) can then be re-written
as an $L_2$ norm:
\begin{eqnarray}
\label{eq_wassersteinAsAnL2Norm}
\wassersteinTree(\branchtree,
\widehat{\branchtree}) = ||\branchtree'- \widehat{\branchtree'}||_2^2.
\end{eqnarray}
Then, given the optimal assignment $\phi_*$, the optimal
$\alpha_* \in \mathbb{R}^{d'}$ are:
\begin{eqnarray}
\nonumber
\begin{array}{lll}
\alpha_* &=& \argmin_\alpha ||\branchtree'- \widehat{\branchtree'}||_2^2\\
\alpha_* &=& \argmin_\alpha||\branchtree' - \Big(\bdtOrigin' +
\big(\pcaBasis(\bdtOrigin')\big)'\alpha\Big)||_2^2\\
\alpha_* &=& \argmin_\alpha||\branchtree' - \bdtOrigin' -
\big(\pcaBasis(\bdtOrigin')\big)'\alpha||_2^2.
\end{array}
\end{eqnarray}
Similarly to the Euclidean case (\autoref{eq_pseudoInverse}), it
follows then that $\alpha_*$ can be expressed as a function of the
pseudoinverse of $\big(\pcaBasis(\bdtOrigin')\big)'$:
\begin{eqnarray}
  \label{eq_optimalAlpha}
  \alpha_* = \Big(\big(\pcaBasis(\bdtOrigin')\big)'\Big)^{+} (\branchtree' -
\bdtOrigin').
\end{eqnarray}

At this stage, the estimation  $\widehat{\branchtree}$ can be \emph{updated}
with the above optimized coefficients $\alpha_*$: $\widehat{\branchtree}
\leftarrow \bdtOrigin +
\pcaBasis(\bdtOrigin)\alpha_*$.

The above \emph{Assignment/Update} sequence is then iterated.
Each iteration decreases the \emph{projection error} $e(\branchtree)$
constructively:
while 
the \emph{Update} phase (2) optimizes $\alpha$ (\autoref{eq_bdtProjectionError}) 
to minimize the projection error 
under the
current assignment $\phi_*$, the next \emph{Assignment} phase (1) further
improves (by construction) the assignments (term $\wassersteinTree$ in 
\autoref{eq_bdtProjectionError}), hence decreasing the projection
error overall. In our implementation, this iterative algorithm stops after a
predefined number of iterations $n_{it}$.

%

\subsection{Initialization}
\label{sec_initialization}
Now that we have introduced
the core low-level  procedure of our
approach (\autoref{sec_basisProjection}), we can detail the initialization
step of our framework, which consists in identifying a relevant initial value
for the overall optimization variable $\theta$ (line
\autoref{line_initialization}, \autoref{alg_overview}).
The BDT transformation
layers $\Pi_k$ are initialized one after the other, i.e. for increasing values
of $k$.

\noindent
\textbf{\emph{(1)} Input initialization:} For each BDT transformation layer
$\Pi_k$, its input origin $\bdtOrigin_k^{in}$ is initialized as the Wasserstein
barycenter $\branchtree_*$ \cite{pont_vis21} of the BDTs on its input.
Next, the first vector of $\pcaBasis_k^{in}$,
is given by the optimal
assignment
(w.r.t. \autoref{eq_wasserstein}) between $\bdtOrigin_k^{in}$ and the
layer's input BDT $\branchtree$ which maximizes
$\wassersteinTree(\bdtOrigin_k^{in}, \branchtree)$, i.e. which induces
the worst projection error $e(\branchtree)$ (\autoref{eq_bdtProjectionError}) 
given an empty basis. Next, the remaining $(d_k-1)$
vectors of $\pcaBasis_k^{in}$ are initialized one after the other, by
including at each step the vector formed by the optimal assignment between
$\bdtOrigin_k^{in}$ and the layer's input BDT $\branchtree$ which induces the
maximum projection error
$e(\branchtree)$ (\autoref{eq_bdtProjectionError}),
given the already
initialized vectors. 
Note that this step makes an extensive usage of the
projection procedure introduced in \autoref{sec_basisProjection}. 
Finally, if
the dimension $d_k$ of $\Pi_k$ is greater than
the number of input BDTs,
the
remaining vectors are initialized randomly, with a controlled norm (set
to the mean of the already initialized vectors).

\noindent
\textbf{\emph{(2)} Output initialization:} For each BDT transformation layer
$\Pi_k$, its output origin $\bdtOrigin_k^{out}$ and basis $\pcaBasis_k^{out}$
are initialized as random linear transformations of its input origin and basis.
Specifically, let $W$ be a random matrix of size
$(2|\bdtOrigin_k^{out}|\times 2|\bdtOrigin_k^{in}|)$. Given
the vector representation ${\bdtOrigin_k^{in}}'$ of $\bdtOrigin_k^{in}$ (see
\autoref{sec_basisProjection}), we initialize $\bdtOrigin_k^{out}$ such that:
${\bdtOrigin_k^{out}}' \leftarrow W {\bdtOrigin_k^{in}}'$. Similarly,
the output
basis of $\Pi_k$ is initialized such that: ${\pcaBasis_k^{out}}' \leftarrow W
{\pcaBasis_k^{in}}'$.

\setlength{\commentindent}{0.2\linewidth}

\begin{algorithm}[t]
 \fontsize{5.625}{9}\selectfont
 \algsetup{linenosize=\tiny}
    \caption{Forward propagation in our Wasserstein Auto-Encoder.}
    \label{alg_forward}

    \hspace*{\algorithmicindent} \textbf{Input1}: Set of input BDTs 
$\branchtreeSet =
\{\branchtree(f_1),
\dots, \branchtree(f_\ensembleSize)\}$.

\hspace*{\algorithmicindent} \textbf{Input2}: Current value of the overall
optimization variable $\theta$.

    \hspace*{\algorithmicindent} \textbf{Output}: Set of reconstructed BDTs
$\widehat{\branchtreeSet} =
\{\widehat{\branchtree(f_1)},
\dots, \widehat{\branchtree(f_\ensembleSize)}\}$.

\hrulefill

\begin{algorithmic}[1]


\FOR{$\branchtree \in \branchtreeSet$}
  \label{line_foreachBDT}
  \STATE $\branchtree_0 \leftarrow \branchtree$.
  \FOR{$k \in \{1, 2, \dots, \sizeEncoding + \sizeDecoding\}$}
    \label{line_foreachLayer}
    \STATE // For each BDT transformation layer $\Pi_k$.
    \STATE $\linearTransformation_k^{in}(\branchtree_{k-1}) \leftarrow
$basisProjection$(\branchtree_{k-1}, \bdtOrigin_k^{in}, \pcaBasis_k^{in}).$
\COMMENT{\autoref{sec_basisProjection}.}
    \STATE $\unitTransformation_k^{in}(\branchtree_{k-1}) \leftarrow
\activation\big(\linearTransformation_k^{in}(\branchtree_{k-1})\big)$.
\COMMENT{\autoref{sec_lowLevelGeometricalTools}.}
    \label{line_inputLayer}
    \STATE $\branchtree_k \leftarrow
\unitTransformation_k^{out}\big(\unitTransformation_k^{in}(\branchtree_{k-1})
\big) =
      \validBDT\big(\bdtOrigin_k^{out} +
\pcaBasis_k^{out}(\bdtOrigin_k^{out})
\unitTransformation_k^{in}(\branchtree_{k-1}) \big).$
\COMMENT{\autoref{sec_lowLevelGeometricalTools}.}
\label{line_reconstructBDT}
  \ENDFOR
  \STATE $\widehat{\branchtreeSet} \leftarrow \widehat{\branchtreeSet} \cup
\branchtree_{\sizeEncoding + \sizeDecoding}$
\ENDFOR
\end{algorithmic}
\end{algorithm}

\subsection{Forward propagation}
\label{sec_forward}
\autoref{alg_forward} presents the main steps of our forward propagation.
This procedure follows directly from our
formulation (\autoref{sec_lowLevelGeometricalTools}).
Each input BDT $\branchtree \in
\branchtreeSet$ is processed independently (line \autoref{line_foreachBDT}).
Specifically, $\branchtree$ will traverse the network one layer $\Pi_k$ at a
time
(line \autoref{line_foreachLayer}). Within each layer
$\Pi_k$, the projection through the input sub-layer $\unitTransformation_k^{in}$
is computed (line \autoref{line_inputLayer})
by
composing a non-linearity
$\activation$ with the basis projection
(\autoref{sec_lowLevelGeometricalTools}). This
yields a set of coefficients
representing
the input
BDT.
Next, following
\autoref{sec_lowLevelGeometricalTools},
these
coefficients are
transformed back into a valid BDT with the output sub-layer
(line \autoref{line_reconstructBDT}). At the end of this process, a set of
reconstructed BDTs $\widehat{\branchtreeSet}$ is available,
for a fixed value of $\theta$.


\subsection{Backward propagation}
\label{sec_backward}
Given the set of reconstructed BDTs $\widehat{\branchtreeSet}$ for the current
value of $\theta$ (\autoref{sec_forward}), the data fitting energy
(\autoref{eq_mtWaeEnergy}) is evaluated. Specifically, for each input BDT
$\branchtree \in \branchtreeSet$, the optimal assignment $\phi_*$ w.r.t.
\autoref{eq_wasserstein} is computed between $\branchtree$ and its
reconstruction, $\branchtree_{\sizeEncoding + \sizeDecoding}$, provided on the
output of the network. Next, similarly to \autoref{sec_basisProjection}
for
basis projections, the vector representation $\branchtree'$
of $\branchtree$ is constructed and re-ordered such that the $i^{th}$ entry of
this vector corresponds to the pre-image by $\phi_*$ of the $i^{th}$ entry of
$\branchtree_{\sizeEncoding + \sizeDecoding}'$ (c.f.
\autoref{eq_vectorReOrdering}). Then,
given the optimal assignment $\phi_*$,
similarly to
\autoref{sec_basisProjection},
$\wassersteinTree(\branchtree, \branchtree_{\sizeEncoding + \sizeDecoding})$
can be
expressed as an $L_2$ norm (\autoref{eq_wassersteinAsAnL2Norm}). Given the set
$\Phi_*$ of all the optimal assignments between the input BDTs and their output
reconstructions,
$\mtPgaError(\theta)$ is then evaluated:
\begin{eqnarray}
\label{eq_energy_evaluation}
\mtPgaError(\theta) = \sum_{i = 1}^N ||
\branchtree(f_i)' - \branchtree_{\sizeEncoding + \sizeDecoding}(f_i)'||_2^2.
\end{eqnarray}


At this stage,
for a given set $\Phi_*$ of optimal assignments,
the evaluation of
\autoref{eq_energy_evaluation} only involves
basic
operations
(as described in the previous
sections: vector re-orderings, pseudoinverse computations, linear 
transformations, and compositions).
All these
operations  are supported
by the
automatic differentiation capabilities of modern neural
frameworks
(in our case \emph{PyTorch} \cite{pytorch}), enabling the automatic estimation
of
$\nabla \mtPgaError(\theta)$.
Then, $\theta$ is updated
by gradient descent \cite{KingmaB14}.

Our overall optimization algorithm (\autoref{alg_overview}) can be
interpreted as a global instance of an \emph{Assignment/Update} strategy.
Each backward propagation updates the overall variable $\theta$ to improve the
data fitting energy (\autoref{eq_mtWaeEnergy}), while the next forward
propagation improves the network outputs and hence their assignments to the
inputs. 
In the remainder, the terms PD-WAE and MT-WAE refer to the usage of
our framework with persistence diagrams 
or merge trees respectively.
We refer the reader to the \julien{Appendix C} for a
detailed
discussion of the meta-parameters of our approach (e.g. layer number, layer 
dimensionality, etc).

%
%
%
%

\begin{figure}
  \centering

\includegraphics[width=\linewidth]{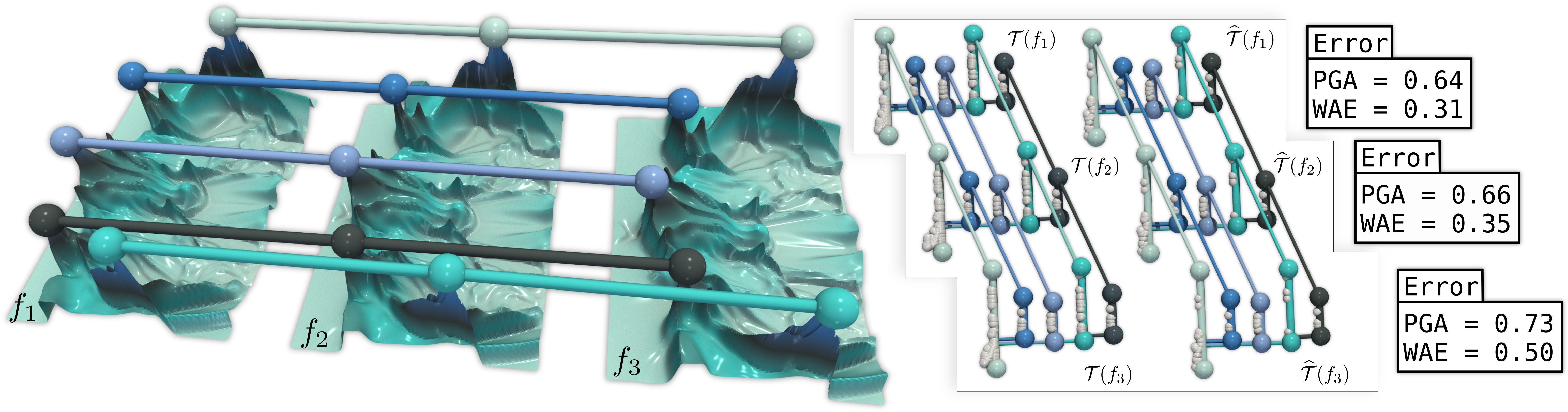}
  \caption{
  \journal{Application of our
  \minor{merge tree compression}
  to feature tracking
(experiment adapted from \cite{pont_vis21, pont_tvcg23} for
comparison purposes). The $5$ most persistent maxima (spheres) of three time 
steps (ion density during universe formation \cite{scivisOverall}) are tracked 
through time (left, $f_1$, $f_2$ 
and $f_3$) by considering the optimal assignment 
(\autoref{eq_wasserstein}) between the corresponding merge trees (inset, left). 
The same tracking procedure is applied to the merge trees compressed by WAE 
(inset, right).
Similarly to PGA \cite{pont_tvcg23}, the resulting tracking is identical with 
the compressed trees. However, in comparison to PGA \cite{pont_tvcg23}, for a 
target compression factor of $13.44$, the relative reconstruction error (right) 
is clearly improved with WAE.}
 }
  \label{fig_tracking}
\end{figure}

\section{Applications}
\label{sec_applications}

This section
illustrates
the utility of our framework in concrete visualization
tasks:
\minor{merge tree compression}
and dimensionality reduction.
\journal{These applications and use-cases are adapted from \cite{pont_tvcg23},
to facilitate comparisons between previous work on the linear encoding of
\minor{merge trees}
\cite{pont_tvcg23} and our novel non-linear framework.}

\subsection{\minor{Merge tree compression}}
\label{sec_dataReduction}

\journal{As discussed by Pont et al. \cite{pont_tvcg23},}
\journal{like}
any
data representation,
merge trees can benefit from lossy
compression.
\journal{For example, for the in-situ analysis of high-performance simulations \cite{AyachitBGOMFM15}, each individual time-step of the simulation can be represented and stored to disk in the form of a topological descriptor  \cite{BrownNPGMZFVGBT21}.}
%
\journal{In this context, this lossy compression eases the manipulation of the generated ensemble of topological descriptors (i.e. it facilitates its storage and transfer).}
Previous work has investigated the
\minor{compression}
of an
ensemble of merge trees via
linear encoding \cite{pont_tvcg23}.
In this
section, we improve this application 
by extending it to non-linear encoding,
thereby enabling more accurate
\minor{compressions.}
Specifically,
the input
ensemble $\branchtreeSet$ of BDTs is compressed, by only storing
 to disk:

\noindent
\emph{\textbf{(1)}} the output sub-layer of the last decoding layer of the
network,
noted $\unitTransformation^{out}_{\sizeEncoding+\sizeDecoding}$ (i.e.
its origin, $\bdtOrigin_{\sizeEncoding+\sizeDecoding}^{out}$, as well as its
basis, $\pcaBasis_{\sizeEncoding+\sizeDecoding}^{out}(
\bdtOrigin_{\sizeEncoding+\sizeDecoding}^{out})$)

\noindent
\emph{\textbf{(2)}} the \journal{corresponding} $N$ BDT
coefficients
$\alpha^{i}_{\sizeEncoding+\sizeDecoding} \in
\mathbb{R}^{d_{\sizeEncoding+\sizeDecoding}}$.

\journal{Note that an alternative
\minor{compression}
strategy would consist in storing
the $N$ BDT coefficients in latent space directly (i.e.
$\alpha^{i}_{\sizeEncoding} \in
\mathbb{R}^{d_{\sizeEncoding}}$), which would be typically more compact than
the $N$ BDT coefficients in the last output sub-layer
($\alpha^{i}_{\sizeEncoding+\sizeDecoding} \in
\mathbb{R}^{d_{\sizeEncoding+\sizeDecoding}}$). However, in order to decompress
this representation, one would need to store to disk the entire set of
$\sizeDecoding$ decoding layers. This significant overhead would only be
compensated for ensembles counting an extremely large number $N$ of members.
In our experiments (\autoref{sec_results}), $N=48$ for the largest ensemble.
Thus, we focus on the first strategy described above (storing potentially
larger sets of coefficients, but a smaller number of decoding layers).}

The compression factor can be controlled with two input parameters:
\emph{(i)}
$d_{\sizeEncoding+\sizeDecoding}$ controls the dimensionality of the last
decoding layer (hence its ability to capture small variabilities) and
\emph{(ii)} $|\bdtOrigin_{\sizeEncoding+\sizeDecoding}^{out}|$ controls the
size of the origin of the last decoding sub-layer (hence its ability to
capture small features). The resulting reconstruction error
(\autoref{eq_mtWaeEnergy}) will be minimized for large values of both
parameters, while the compression factor will be minimized for low values.
%
In
the following
experiments,
we
set both parameters to their default values
\julien{(Appendix C).}
To decompress a BDT $\branchtree(f_i)$,
its stored coefficients $\alpha^{i}_{\sizeEncoding+\sizeDecoding}$ are simply
propagated through the stored output sub-layer of the network,
$\unitTransformation^{out}_{\sizeEncoding+\sizeDecoding}$.

\begin{figure}
  \centering
  \includegraphics[width=\linewidth]{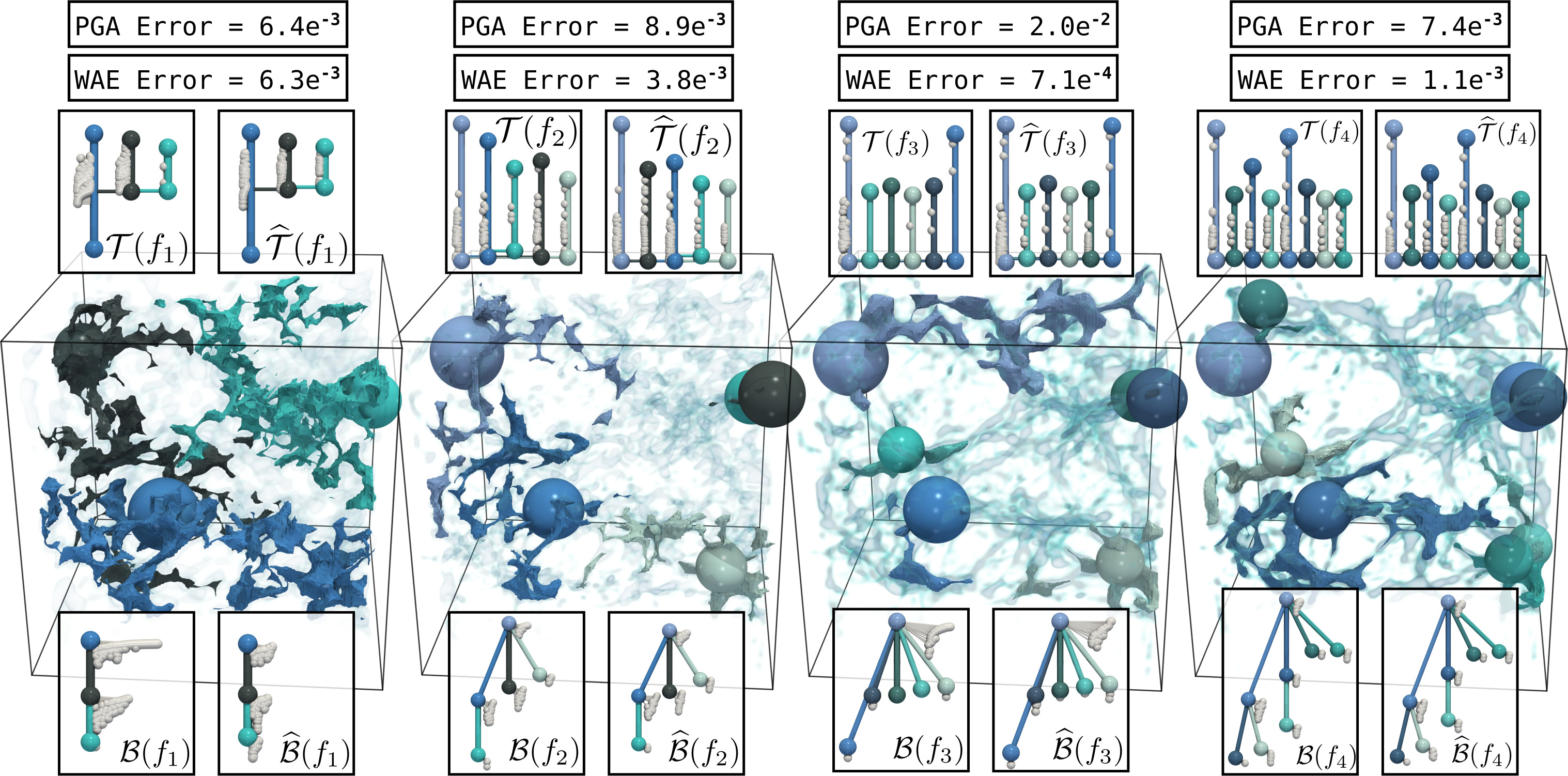}
\caption{\journal{Application of our
\minor{merge tree compression}
to topological
clustering (experiment adapted from \cite{pont_vis21, pont_tvcg23} for 
comparison purposes). The ensemble is clustered \cite{pont_vis21} 
based on the merge trees (top left insets) of the ensemble members (each
column represents a member from one of the $4$ clusters). 
The same clustering procedure  \cite{pont_vis21}  is applied to the merge trees 
compressed by WAE (top right insets). Similarly to PGA
\cite{pont_tvcg23}, the resulting clustering is identical with the compressed 
trees (it exactly matches the ground-truth classification \cite{pont_vis21}).
However, in comparison to 
PGA \cite{pont_tvcg23}, for a target compression factor of $15.09$, the 
relative reconstruction error (top) is clearly improved with WAE. Visually,
the compressed trees look very similar to the original ones: the prominent 
features (in colors, non-prominent features are shown with small white nodes) 
are well preserved in terms of number and persistence. The same holds for the 
compressed BDTs (bottom right insets) which are nearly isomorphic to the
original
BDTs (bottom left insets, the color shows the
assignment
between the original and compressed BDTs).}}
  \label{fig_clustering}
\end{figure}

\begin{figure*}
  \centering
  \includegraphics[width=\linewidth]{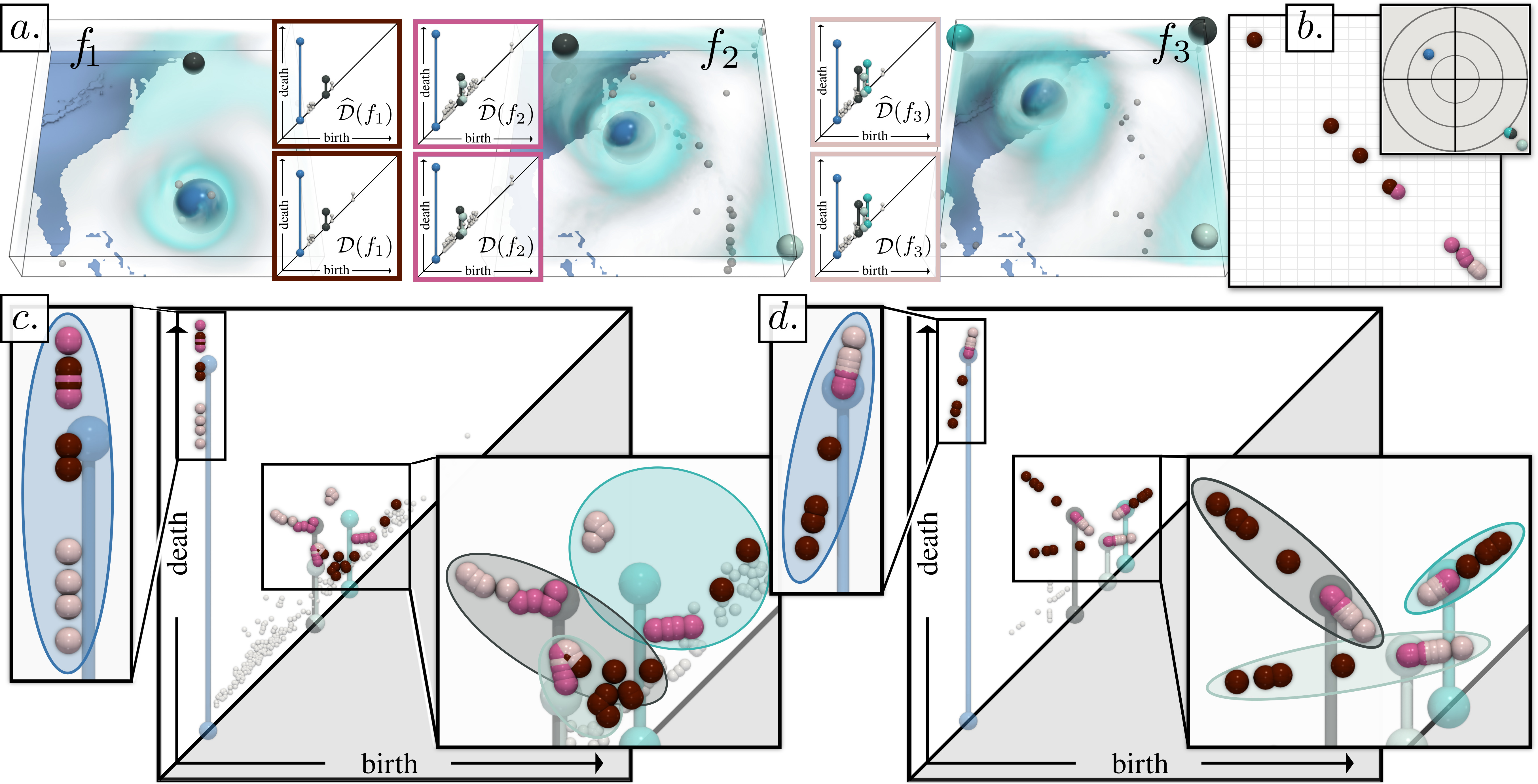}
  \caption{Visual analysis of
  the \emph{Isabel}
ensemble (one member per ground-truth class, top), with PD-WAE
(in this example, $\sizeEncoding = \sizeDecoding = 2$).
Our work enables
\minor{diagram compression,}
while providing reconstructed diagrams $\big((a)$, top insets$\big)$
which are
highly similar to the input (below). The 2D layout generated
by PD-WAE
$(b)$
recovers the temporal structure of the ensemble (the clusters are aligned in
chronological order, from dark red to light pink). The PCV (grey
inset) indicates that the peripheral gusts of wind in the data (white, black
and cyan spheres) grow in importance when moving towards the bottom right corner
of the latent space (their persistence increases over time).
In contrast,
the hurricane eye (light blue) exhibits less variability, but
with
stronger values towards the top left corner of the latent space (start of the
sequence).
The aggregated views
(bottom, overlapping all diagrams,
transparent: barycenter) for
the input $(c)$ and the latent space $(d)$
show that PD-WAE
nicely
recovers a per-feature
parameterization in the
latent space (one ellipse per barycenter feature),
which is
locally
consistent with the
data
temporal
evolution (cluster color).
}
  \label{fig_useCasePD}
\end{figure*}

\begin{figure}
  \centering
  \includegraphics[width=\linewidth]{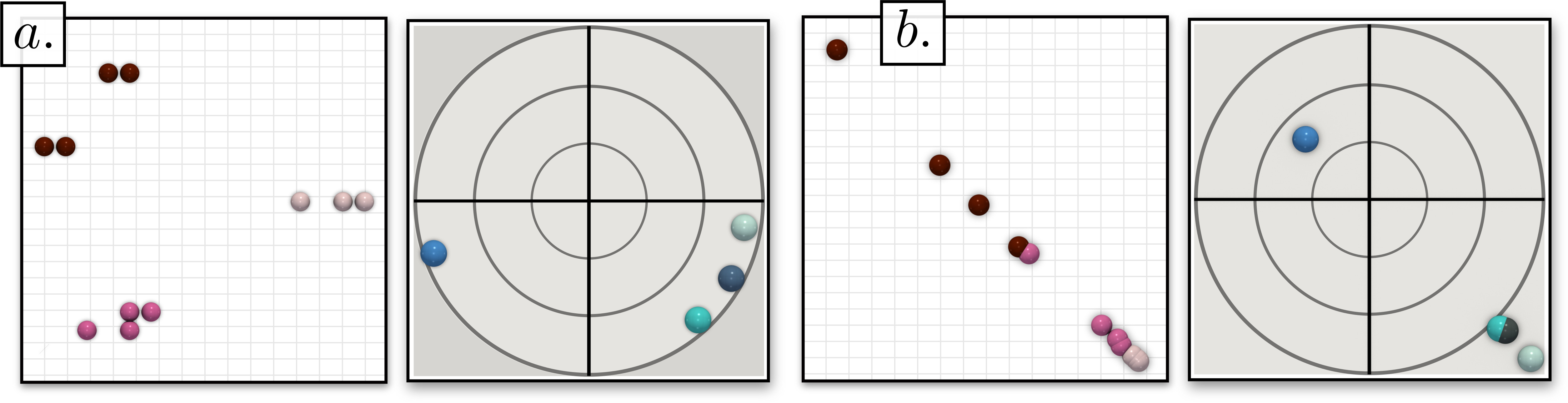}
  \caption{\journal{Qualitative comparison of the 2D layouts (left, white inset)
and PCVs (right, grey inset) between PD-PGA
\cite{pont_tvcg23} \emph{(a)} and PD-WAE \emph{(b)} for the \emph{Isabel}
ensemble. PD-WAE manages to recover the intrinsic temporal
structure of the ensemble and produces a linear alignment of the clusters in
order of temporal appearance (dark red, pink, light pink).}}
  \label{fig_pd_pga_vs_wae}
\end{figure}

Figs. \ref{fig_tracking} and
\ref{fig_clustering}
show two examples of visualization
\journal{tasks}
(feature tracking and ensemble
clustering, use cases replicated from \journal{\cite{pont_vis21, pont_tvcg23}
for comparison purposes}).
\journal{In these experiments, the BDTs have been compressed with the strategy
described above. Next, the de-compressed BDTs have been used as an input to
these two analysis pipelines.}
In both cases, the output \journal{obtained with the de-compressed BDTs} is
identical to the
\journal{output}
obtained with the
\emph{original} BDTs.
This shows the
viability of
\journal{the de-compressed}
BDTs
and \journal{it}
demonstrates the utility of
\journal{this}
\minor{compression}
scheme.




\subsection{Dimensionality reduction}
\label{sec_dimensionalityReduction}

\journal{This section describes how to use MT-WAE to generate 2D layouts of the
ensemble, for the global visual inspection of the ensemble.}
This is achieved
by setting
$d_{\sizeEncoding} = 2$
and
by embedding each
BDT
$\branchtree(f_i)$ as a point in the
plane,
\journal{at}
its \journal{latent} coordinates
$(\alpha_{\sizeEncoding}^i)_1$ and
$(\alpha_{\sizeEncoding}^i)_2$. This results in a summarization view of the
ensemble, grouping
similar BDTs \journal{together}
(\autoref{fig_teaser}c).
The flexibility of our framework allows to further improve the quality
of this 2D layout. Specifically, \julien{Appendix D} introduces two penalty
terms aiming at
\emph{(1)} improving the preservation of the Wasserstein metric
$\wassersteinTree$ and \emph{(2)} improving the preservation of the clusters of
BDTs.

\journal{We augment our 2D layouts with \emph{Persistence Correlation
Views} (PCV) which were introduced in \cite{pont_tvcg23}. In short, the PCV
embeds a branch $b$ of the barycenter $\branchtree_*$ \cite{pont_vis21} as
a point in 2D, in order to represent the variability of the corresponding
feature in the ensemble, as a function of the coordinates in latent space.
Specifically, the optimal assignments $\phi_{*_i}$ between $\branchtree_*$ and
each input BDT $\branchtree(f_i)$ is first computed (\autoref{eq_wasserstein}).
Next, for a given branch $b \in \branchtree_*$, the Pearson correlation
$\rho\big(p_{b_i}, (\alpha_{\sizeEncoding}^i)_1\big)$
between the
persistence $p_{b_i}$ of $\phi_{*_i}(b) \in \branchtree(f_i)$ and the first
coordinate in latent space $(\alpha_{\sizeEncoding}^i)_1$ is computed for the
ensemble (i.e. for $i \in \{1, 2, \dots, N\}$). Next, the Pearson correlation
$\rho\big(p_{b_i}, (\alpha_{\sizeEncoding}^i)_2\big)$ is computed similarly
with regard to the second coordinate in latent space
$(\alpha_{\sizeEncoding}^i)_2$. Finally, $b$ is embedded in the PCV at the
coordinates
$\Big(\rho\big(p_{b_i}, (\alpha_{\sizeEncoding}^i)_1\big), \rho\big(p_{b_i},
(\alpha_{\sizeEncoding}^i)_2\big)\Big)$.
To avoid clutter in the visualization, we only report the most persistent
branches of $\branchtree_*$  in the PCV. Intuitively, points in the PCV which
are located far away from the center, along a given direction, indicate
a strong correlation between that direction in latent space, and
the persistence of the corresponding feature in the ensemble.}

\journal{PCVs enable the identification
of
patterns of feature variability within the ensemble, as discussed
in \autoref{fig_useCasePD}. This case study considers the \emph{Isabel}
ensemble, which consists of $12$ scalar fields representing the wind velocity
magnitude in a hurricane simulation. The ensemble comes with a ground-truth
classification \cite{pont_vis21}: $4$ members correspond to the
formation of the hurricane (e.g. $f_1$, \autoref{fig_useCasePD}), $4$ other
members to its drift (e.g. $f_2$, \autoref{fig_useCasePD}) and $4$ other members
to its landfall (e.g. $f_3$, \autoref{fig_useCasePD}). For this ensemble, our
PD-WAE approach produces a 2D layout $\big($\autoref{fig_useCasePD}b)$\big)$
which manages to recover the temporal coherency of the ensemble: the formation
(dark red), drift (pink) and landfall (light pink) clusters are arranged
in order along a line $\big($direction $(1, -1)\big)$. This shows the
ability of PD-WAE to recover the intrinsic structure of the ensemble (here its
temporal nature). The PCV (grey inset) further helps appreciate the variability
of the features in the ensemble. There, each colored point indicates a 
persistent
feature of the barycenter:
the eye of the hurricane is represented by the blue sphere, while the cyan,
black and white sphere represent peripheral gusts of wind $\big($see the
matching features in the data, \autoref{fig_useCasePD}a)$\big)$. The PCV
clearly identifies two patterns of feature variability, along the direction $(1,
-1)$, which coincides with the temporal alignment of the clusters in latent
space. Specifically, it indicates that the persistence of the hurricane eye
will be larger
in the top left corner of the latent space, i.e.
towards the beginning of the temporal sequence (dark red
cluster). This is confirmed visually when inspecting the persistence diagrams of
the individual members (the blue feature is less persistent in $\diagram(f_3)$
than in $\diagram(f_1)$ and $\diagram(f_2)$). In short, this visually encodes
the fact
that the strength of the hurricane eye decreases with time.
In contrast, the features corresponding to peripheral wind gusts (cyan, black
and white spheres) exhibit a common variability pattern, distinct form that of
the  hurricane eye: the persistence of the corresponding features increases
as one moves along the direction $(1,
-1)$ in latent space, i.e.
as time increases (pink and light pink clusters).
This is confirmed visually in the individual members, where the
persistence of these features is larger in $\diagram(f_3)$ than in
$\diagram(f_1)$ and $\diagram(f_2)$. In short, this visually encodes the fact
that the strength of the peripheral wind gusts increases with time.
Overall, while the 2D layout generated by PD-WAE enables the
visualization of the intrinsic structure of the ensemble (here, its temporal
nature), the PCV enables the visualization and interpretation of the variability
in the ensemble at a feature level. The caption of \autoref{fig_teaser}
includes a similar discussion for MT-WAE.
}


%

\journal{\autoref{fig_pd_pga_vs_wae} provides a qualitative comparison of the
2D layouts and
Persistence Correlation Views (PCVs)
between PD-PGA
\cite{pont_tvcg23} and our novel non-linear framework, PD-WAE (see
\autoref{sec_quality} for an extensive quantitative comparison). Specifically,
it shows that, while PD-PGA $\big($\autoref{fig_pd_pga_vs_wae}a)$\big)$ manages
to isolate the
clusters well, its 2D layout does not recover the intrinsic, one-dimensional,
temporal structure of the ensemble. In contrast, as discussed above, PD-WAE
$\big($\autoref{fig_pd_pga_vs_wae}b)$\big)$ manages to recover this intrinsic
structure
and produces a linear
alignment of the clusters along the direction $(1, -1)$, in order of
their temporal appearance. This alignment greatly facilitates the
interpretation of the PCV, since time is now visually encoded there by the
linear direction $(1, -1)\big)$ (whereas it would be encoded by a
curve
in the case of
PD-PGA).}

In contrast to
\minor{standard}
auto-encoders,
our approach
\minor{explicitly manipulates}
topological descriptors throughout the network.
This
results in
improved interpretability and enables new
capabilities:

\noindent
\textbf{\emph{(1)} Latent feature transformation:} As
discussed in \autoref{fig_useCasePD},
it is now possible to visualize
how topological descriptors are
(non-linearly)
transformed by the auto-encoder.
Specifically, the aggregated views of the birth/death space (bottom)
illustrates how PD-WAE  \emph{unwraps} the
diagrams in  latent space,
nicely recovering
the data
temporal evolution \journal{at a feature level (see  the temporally consistent 
linear arrangements of points for each barycenter feature in \autoref{fig_useCasePD}d), from dark red to
pink and light pink)}.

\noindent
\textbf{\emph{(2)} Latent space navigation:}
Given a point in latent space,
it is
now
possible
to
efficiently
reconstruct its
BDT/MT
by
propagating its 
\journal{latent coordinates}
through the
decoding layers\journal{, enabling}
an interactive exploration of
\journal{the merge tree latent space}
(\autoref{fig_teaser}d).

\noindent
\textbf{\emph{(3)} Feature traversal analysis:}
For each consecutive layers $\Pi_k$ and $\Pi_{k+1}$, we compute the optimal
assignment (\autoref{eq_wasserstein}) between their input origins,
$\bdtOrigin_k^{in}$ and $\bdtOrigin_{k+1}^{in}$.
Next, we compute the optimal
assignment between the barycenter $\branchtree_*$ \cite{pont_vis21} of the
input ensemble and the first origin $\bdtOrigin_1^{in}$.
This yields an
explicit tracking of each  branch $b$ of  $\branchtree_*$ down to the latent
space.
We introduce the notion of \emph{Feature Latent
Importance} (FLI), given by the persistence of $b$ in latent space,
divided by its original persistence.
FLI indicates if a feature gains (or looses) importance in latent space.
This enables the identification of the most \emph{informative} features
in the ensemble.

This is illustrated in \autoref{fig_teaser}\journal{e)}, where
the cyan,
white, dark blue and black features
exhibit large
FLI values (red circles).
In the trees,
these
features are indeed present in most of the
\journal{ensemble}
$\big($\autoref{fig_teaser}\journal{d)}$\big)$, with
only moderate variations in persistence. Interestingly, the global
maximum of seismic wave (light blue feature) is not a very informative
feature
(light blue circle): while it is also present throughout the sequence,
its persistence decreases significantly (left to right). This
\journal{visually encodes}
that, as the seismic wave travels from the epicenter, the strength of its
global maximum is no longer significant in front of other local maxima (which
illustrates the energy diffusion process).

\begin{table}
\caption{Running times (in seconds) of our algorithm for PD-WAE and MT-WAE
computation (first sequential, then with 20 cores).}
\label{tab_timings}
\centering
\scalebox{0.55}{
  \begin{tabular}{|l|r|r||r|r|r||r|r|r|}
    \hline
    \rule{0pt}{2.25ex}  \textbf{Dataset} & $\ensembleSize$ & $|\branchtree|$ &
\multicolumn{3}{c||}{PD-WAE} & \multicolumn{3}{c|}{MT-WAE} \\
    & & & 1 c. & 20 c. & Speedup & 1 c. & 20 c. & Speedup \\
    \hline
      Asteroid Impact (3D) & 7 & 1,295 & 2,819.38 & 989.42 & 2.85 & 5,946.81 &
1,522.89 & 3.90 \\
      Cloud processes (2D) & 12 & 1,209 & 11,043.20 & 1,318.63 & 8.37 &
16,150.90 & 2,566.98 & 6.29 \\
      Viscous fingering (3D) & 15 & 118 & 1,345.31 & 268.78 & 5.01 & 3,727.64 &
417.31 & 8.93 \\
      Dark matter (3D) & 40 & 316 & 125,724.00 & 10,141.30 & 12.40 & 135,962.00
& 8,051.02 & 16.89 \\
      Volcanic eruptions (2D) & 12 & 811 & 4,925.15 & 638.58 & 7.71 & 4,151.04
&
449.14 & 9.24 \\
      Ionization front (2D) & 16 & 135 & 627.04 & 95.31 & 6.58 & 1,140.44 &
144.57 & 7.89 \\
      Ionization front (3D) & 16 & 763 & 17,285.10 & 1,757.71 & 9.83 &
101,788.00 & 5,350.46 & 19.02 \\
      Earthquake (3D) & 12 & 1,203 & 14,272.10 & 2,074.52 & 6.88 & 9,888.68 &
1,024.45 & 9.65 \\
      Isabel (3D) & 12 & 1,338 & 3,485.03 & 436.56 & 7.98 & 23,240.90 &
1,669.20
& 13.92 \\
      Starting Vortex (2D) & 12 & 124 & 215.46 & 95.45 & 2.26 & 281.80 & 147.51
& 1.91 \\
      Sea Surface Height (2D) & 48 & 1,787 & 41,854.70 & 2,901.39 & 14.43 &
222,594.00 & 13,540.20 & 16.44 \\
      Vortex Street (2D) & 45 & 23 & 460.46 & 152.35 & 3.02 & 117.88 & 38.88 &
3.03 \\
    \hline
  \end{tabular}
}
\end{table}

\section{Results}
\label{sec_results}

This section presents experimental results obtained on a computer with
two Xeon CPUs (3.2 GHz, 2x10 cores, 96GB of RAM). The input
merge trees were computed with FTM \cite{gueunet_tpds19} and pre-processed to
discard noisy features
(persistence simplification threshold:
$0.25\%$ of the data range).
%
We
implemented our approach in C++ (with OpenMP and
PyTorch's C++ API \cite{pytorch}),
as modules for TTK \cite{ttk17, ttk19}.
Experiments were performed on
a set of
$12$ public ensembles
described in
\cite{pont_vis21}, which includes a variety of simulated and acquired 2D and 3D
ensembles extracted from previous work and past SciVis contests
\cite{scivisOverall}.

\subsection{Time performance}
\label{sec_timePerformance}
The Wasserstein distance computation (\autoref{eq_wasserstein}) is
the
most
expensive
sub-procedure
of our approach. It
intervenes during
energy evaluation (\autoref{sec_backward}) but also at
each iteration of basis projection (\autoref{sec_basisProjection}), itself
occuring at each propagation iteration (\autoref{sec_forward}), for each input
BDT. To compute this distance, we use a fine-grain task-based
parallel algorithm \cite{pont_vis21}.
We  leverage
further
parallelism to
accelerate the process.
Specifically, for each
input BDT, its forward propagation (\autoref{sec_forward}) can be run in a
distinct parallel task. Similarly,
when evaluating the overall energy (\autoref{sec_backward}),
the
distance between an input BDT and its output reconstruction
is
computed
in a distinct parallel task for each BDT.
\autoref{tab_timings} evaluates the time performance of our framework for
persistence diagrams (PD-WAE) and merge trees (MT-WAE). In sequential mode, the
computation time is a function of the ensemble size ($\ensembleSize$), the
tree sizes ($|\branchtree|$)\minor{, and the network size ($\sizeEncoding+\sizeDecoding$)}.
\minor{Specifically, our approach computes, for each optimization iteration, $n_{it}\times(\sizeEncoding+\sizeDecoding)\times\ensembleSize$ Wasserstein distances, each of which requiring $\mathcal{O}(|\branchtree|^2)$ steps in practice.}
In parallel, the iterative nature of our
approach
(\autoref{alg_overview}) challenges
parallel efficiency.
However, timings are still   improved
after parallelization
(orders of minutes on average), with a very good parallel efficiency for the
largest ensembles \minor{(up to 95\%)}.

\subsection{Framework quality}
\label{sec_quality}

Figs. \ref{fig_tracking} and \ref{fig_clustering} report compression
factors for our application to
\minor{merge tree compression}
(\autoref{sec_dataReduction}).
These are ratios between the storage size of the input $\ensembleSize$ BDTs and
that of their compressed form.
For a fixed target compression factor, WAE clearly improves the reconstruction
error over linear encoding (PGA \cite{pont_tvcg23}).
\julien{Appendix E} extends this error comparison to all our test ensembles,
%
and
shows that, for
identical compression factors, our framework improves the reconstruction error
over PGA \cite{pont_tvcg23} by 37\% for persistence diagrams, and
52\% for merge trees, hence confirming the accuracy superiority of WAE
over PGA.

\begin{figure}
  \centering
\includegraphics[width=\linewidth]{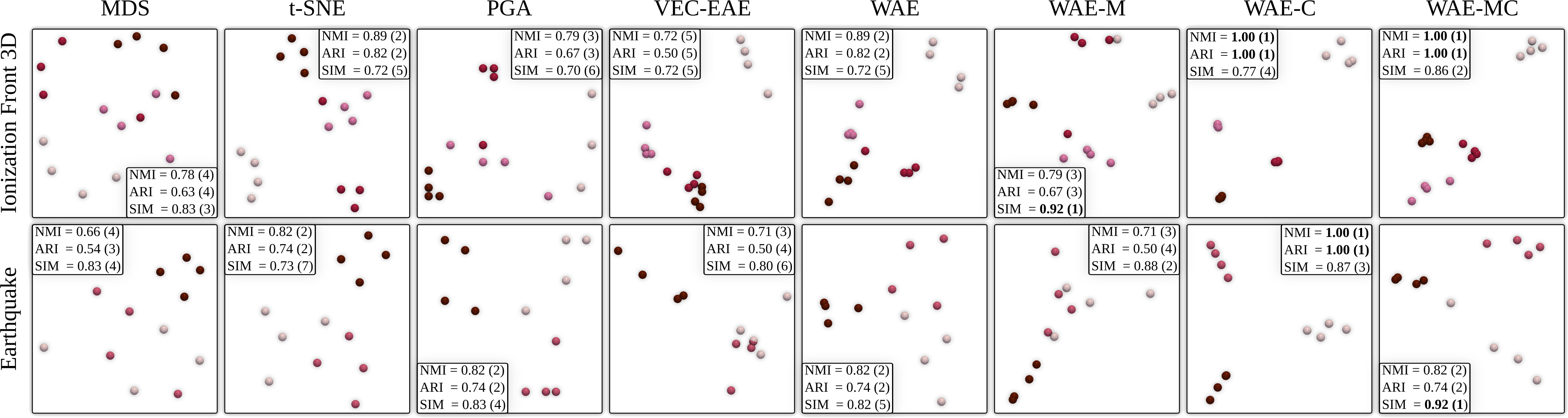}
  \caption{Comparison of planar layouts for typical dimensionality reduction
techniques, on two  merge tree ensembles.
The color encodes the classification ground-truth
\cite{pont_vis21}. For each quality score,
the best
value appears bold and the rank of the score
is shown in
parenthesis. \journal{This experimental protocol is adapted from
\cite{pont_tvcg23} for
comparison purposes.}}
  \label{fig_comparisonPlanarLayout}
\end{figure}

\autoref{fig_comparisonPlanarLayout} provides a visual comparison for the
planar layouts generated by a selection of typical dimensionality reduction
techniques, applied on the input merge tree ensemble (i.e.  each point is
a merge tree).
\journal{This experiment is adapted from \cite{pont_tvcg23} for comparison
purposes.}
This figure  reports quantitative
scores.
For a given technique,
to quantify its ability
to preserve
the \emph{structure} of the ensemble \minor{(i.e. its organization into ground-truth classes)}, we run $k$-means in the 2D layouts and
evaluate the quality of the resulting clustering (given the ground-truth
\cite{pont_vis21}) with the normalized mutual information (NMI) and
adjusted rand index (ARI).
To quantify its ability
to preserve the \emph{geometry} of the ensemble \minor{(i.e. to preserve its disposition within the Wasserstein metric space $\branchspace$)}, we report the metric
similarity indicator SIM \cite{pont_tvcg23}, which evaluates the preservation
of
the Wasserstein metric $\wassersteinTree$.
All these scores vary
between $0$ and $1$,
with $1$ being optimal.

MDS
\cite{kruskal78} and t-SNE \cite{tSNE} have been applied on the
distance matrix of
the input merge trees (Wasserstein distance,
\autoref{eq_wasserstein}, default parameters
\cite{pont_vis21}). By design, MDS preserves well the
metric $\wassersteinTree$ (good SIM),
at the expense of mixing
ground-truth classes together (low NMI/ARI).
t-SNE behaves symmetrically (higher NMI/ARI, lower SIM).
We applied PGA \cite{pont_tvcg23}
by setting
the origin size parameter to a value
compatible
to our latent space
($0.1
|\branchtreeSet|$, Appendix C).
As expected,
PGA
provides a trade-off between the
extreme behaviors of MDS and t-SNE,
with an improved cluster preservation over MDS (NMI/ARI),
and an improved metric preservation over t-SNE (SIM).
WAE also constitutes a trade-off between MDS and
t-SNE, but with improved quality scores over PGA.

\begin{table}
  \caption{
  Comparison of \emph{aggregated} layout quality scores (i.e. averaged over
\emph{all} merge tree ensembles, bold: best values).
WAE-MC provides both superior metric
(SIM)
and cluster
(NMI/ARI)
preservation to pre-existing techniques (MDS \cite{kruskal78}, t-SNE
\cite{tSNE}, PGA \cite{pont_tvcg23}, VEC-EAE).
  }
  \centering
  \scalebox{0.675}{
    \begin{tabular}{|l||r|r|r|r||r|r|r|r|}
      \hline
      \textbf{Indicator} & MDS
      & t-SNE
      & PGA & VEC-EAE & \textbf{WAE} & \textbf{WAE-M} & \textbf{WAE-C} &
\textbf{WAE-MC}\\
      \hline
      NMI & 0.78 & 0.83 & 0.82 & 0.71 & 0.84 & 0.76 & \textbf{0.96} & 0.87 \\
      ARI & 0.68 & 0.75 & 0.74 & 0.55 & 0.77 & 0.63 & \textbf{0.95} & 0.82 \\
      \hline
      SIM & 0.86 & 0.75 & 0.79 & 0.74 & 0.78 & \textbf{0.87} & 0.78 & 0.86 \\
      \hline
    \end{tabular}
  }
  \label{tab_qualityIndicators}
\end{table}

We compare our approach to a standard auto-encoder
(EAE, \autoref{sec_euclideanAutoEncoders}) applied on
\minor{the following}
vectorization of the
input merge trees. \minor{Note that several vectorizations of persistence diagrams have been studied \cite{Adams2015, Bubenik15, kim20}. However we focus in this work on merge trees and only few vectorizations have been documented for these \cite{LI2021}. Hence we focus on the following strategy, inspired by  \cite{LI2021}.}
Each input BDT $\branchtree(f_i)$ is embedded in
$\mathbb{R}^{2|\branchtree_*|}$, such that the $j^{th}$ entry of this vector
corresponds to the birth/death location of the branch of $\branchtree(f_i)$
which maps to the $j^{th}$ branch of the barycenter $\branchtree_*$
\cite{pont_vis21}.
Next, we feed these
vectorizations to an EAE, with the same meta-parameters as
our approach (i.e. number of
layers,
dimensionality per layer).
The corresponding results
appear
in
the VEC-EAE column.
Our approach (WAE)
outperforms this straightforward application of
EAE, with clearly
higher
clustering scores (NMI/ARI) and improved metric
scores (SIM).

\autoref{fig_comparisonPlanarLayout} also
reports the layouts obtained with
our approach after enabling the metric penalty term (WAE-M), the clustering
penalty term (WAE-C) and both (WAE-MC),
c.f. \julien{Appendix D}.
WAE-M  (respectively WAE-C) significantly improves the metric (respectively
cluster) preservation over MDS (repectively t-SNE). The
combination of
the two
terms (WAE-MC)
improves both
quality scores
\emph{simultaneously}:
it outperforms MDS
(SIM)
\emph{and} it improves t-SNE
(NMI/ARI). In
other
words, WAE-MC improves established
methods
by outperforming them on their dedicated  criterion
(SIM for MDS, NMI/ARI for t-SNE).

\julien{Appendix F} extends our visual analysis to all our test ensembles.
\autoref{tab_qualityIndicators} also extends  our quantitative analysis
to
all our test ensembles. It confirms
the clear superiority of WAE
over VEC-EAE. It also confirms that the combination of our
penalty terms
(WAE-MC) provides the best metric
(SIM) and cluster (NMI/ARI)
scores
over existing techniques.



\subsection{Limitations}
\journal{As discussed in \autoref{sec_metricSpace}, the parameter $\epsilon_1$
of the Wasserstein distance between merge trees ($\wassersteinTree$)
acts as a control knob, that balances the practical stability of the
metric
with its
discriminative power.
Specifically, for $\epsilon_1 = 1$, we have 
$\wassersteinTree =
\wasserstein{2}$ 
and $\wassersteinTree$ is stable, but
less discriminative.
Pont et al. \cite{pont_vis21} showed
experimentally that for relatively low values of $\epsilon_1$ ($0.05$),
$\wassersteinTree$ still behaved in a stable manner in practice for reasonable
noise levels.
Our overall MT-WAE framework behaves similarly.
Appendix G provides a detailed
empirical stability evaluation of our framework in the presence of additive
noise. In particular, this experiment shows that for
reasonable
levels of
additive
noise $\epsilon$ (normalized with regard to the function range),
typically $\epsilon < 0.1$, the recommended default value of $\epsilon_1$
($0.05$) results in a stable MT-WAE computation. For larger noise levels
($\epsilon > 0.1$), MT-WAE provides similar stability scores to PD-WAE, for
values of $\epsilon_1$ which are still reasonable in terms of discriminative
power ($\epsilon_1 = 0.1$).}

\journal{A possible direction to improve the practical stability of the
framework without having to deal with a control parameter such as $\epsilon_1$
would be to consider branch decompositions driven by other criteria than
persistence (such as hyper-volume \cite{carr04} for instance). However, the
persistence criterion plays a central role in the Wasserstein distance between
merge trees, as discussed by Pont et al. \cite{pont_vis21} (\journal{Sec. 4}),
in particular to guarantee that interpolated BDTs computed during geodesic
construction can indeed be inverted into a valid MT. Thus, other branch
decomposition criteria than persistence would require
to derive a
completely new procedure for several key components of our framework, such as
geodesic computation or barycenter estimation. This is an orthogonal research
direction to this work, which we leave for future work.}

Similarly to other optimization problems
based on
topological
descriptors
\cite{Turner2014, vidal_vis19, pont_vis21, pont_tvcg23}, our
energy is not convex.
However, our experiments indicate that
our initialization strategy (\autoref{sec_initialization}) leads to relevant
solutions,
which can be successfully applied for visualization
(\autoref{sec_applications}).


\begin{figure}
  \centering
  \includegraphics[width=\linewidth]{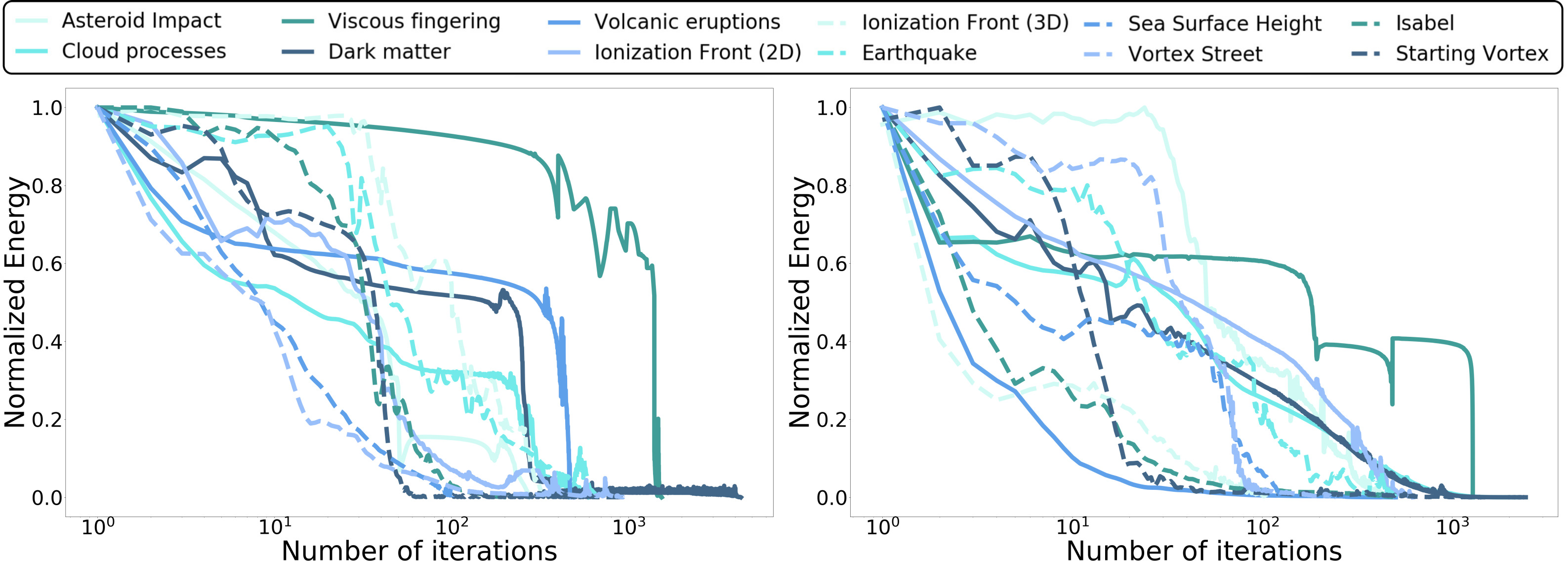}
  \caption{\minor{Evolution of the normalized reconstruction error along the
iterations, for PD-WAE (left) and MT-WAE (right).}}
  \label{fig_convergenceFigure}
\end{figure}


Since it is based on neural networks, our approach inherits from
their intrinsic limitations. Specifically, the energy is not guaranteed to
monotonically decrease
over the iterations.
However, this
theoretical limitation has never translated into a practical limitation in our
experiments.
\minor{\autoref{fig_convergenceFigure} reports the evolution of the normalized
reconstruction error for PD and MT-WAE computations, for all our test
ensembles.
In particular, it
shows that typical,
temporary
energy increases can indeed be observed (as often reported when optimizing neural networks), but without preventing the network from
converging overall (i.e. reaching a state where the energy decreases by
less than $1\%$ between consecutive iterations).}
Like other neural methods,
our approach is conditioned by the meta-parameters defining the network (i.e.
number of layers, dimensionality of each layer, etc.).
However, we ran
our experiments with
fairly
basic values for these
meta-parameters 
(as detailed in \julien{Appendix C)}
and still obtained
substantial improvements over linear encoding based on PGA \cite{pont_tvcg23}.
This indicates that
\journal{optimizing in the future}
these
meta-parameters is likely to
improve the quality of our framework,
however possibly at the expense of longer computations.
\minor{Finally, our application to merge tree compression does not guarantee any error bound in its current form, which we leave for future work.}

\section{Conclusion}

In this paper, we presented a computational framework for the
Wasserstein Auto-Encoding of merge trees (and persistence diagrams), with
applications to
\minor{merge tree compression}
and dimensionality reduction.
Our approach improves previous linear attempts at merge tree encoding, by
generalizing them to non-linear encoding, hence leading to lower reconstruction
errors. In contrast to traditional auto-encoders, our novel layer model enables
our neural networks to process topological descriptors natively, without
pre-vectorization. As
shown
in our experiments, this contribution
leads not only to superior accuracy (\autoref{sec_quality}) but also to
superior interpretability (\autoref{sec_dimensionalityReduction}): with
our work, it is now possible to
interactively explore the latent space and analyze how
topological
features are transformed by the network in its attempt to best encode the
ensemble.
Overall, the
visualizations derived from our
contribution (Figs.
\ref{fig_teaser}, \ref{fig_useCasePD})
enable the
interactive, visual inspection of the ensemble, both at a
global level (with our 2D layouts) and at a feature level.
Specifically, our novel notion of \emph{feature latent importance}
enables the identification of
the most informative features in the ensemble.

In the future, we will continue our work towards the development of further
statistical tools for the visual analysis of ensemble data, based on topological
descriptors.
\minor{In particular, there are several research avenues for improving our current approach.
For example, the Wasserstein distance between merge trees is subject to several meta-parameters (\autoref{sec_metricSpace}), for which we provide generic default values which have shown to be relevant in practice. A possible improvement could consist in letting the auto-encoding framework optimize these meta-parameters (on a per branch basis). However, this data-driven setup of the meta-parameters of our approach would come at the expense of extended running times.}
\minor{Another research avenue could consist in combining our framework with existing approaches on topological losses for image segmentation \cite{HuLSC19, HuWLSC21, StuckiPSMB23}, in order to also auto-encode the scalar data.}
\minor{Moreover, another direction could consist in training a single neural network for auto-encoding multiple ensembles at once. However, this would require to derive new normalization strategies, since, as the scalar fields can take arbitrarily distinct value ranges from an ensemble to the next, the Wasserstein distances between their members can also take arbitrarily distinct values, which would challenge an efficient sampling of the metric space.}
\minor{Finally,} we will investigate the usage of neural networks
exploiting topological descriptors for further visual analysis tasks, such as
trend
analysis or anomaly detection \minor{or shape classification \cite{shrec14}}.
In that context, we believe that our new
layer model (natively processing topological descriptors)
sets the foundations for an accurate and interpretable usage
 of topological representations with neural networks.

%


%

%

\ifCLASSOPTIONcompsoc
  \section*{Acknowledgments}{\small
This work is partially supported by the European Commission grant
ERC-2019-COG \emph{``TORI''} (ref. 863464, \url{https://erc-tori.github.io/}).}
\else
  \section*{Acknowledgment}
\fi


\ifCLASSOPTIONcaptionsoff
  \newpage
\fi



%
\bibliographystyle{abbrv-doi}
\bibliography{template}




%

\begin{IEEEbiography}[{\includegraphics[width=1in,height=1.25in,clip,
keepaspectratio]{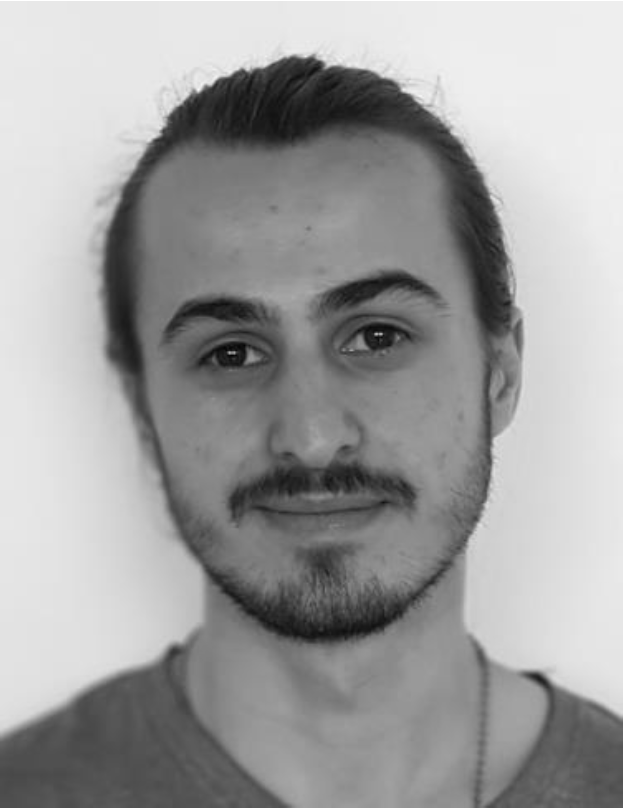}}]{Mathieu Pont}
is a Ph.D. student at Sorbonne Universite. He received a M.S. degree in Computer Science
from Paris Descartes University in 2020.
He is an active contributor to
the Topology ToolKit
(TTK), an open source library for
topological data analysis.
His notable contributions to TTK include distances, geodesics and barycenters
of merge trees, for feature tracking and ensemble clustering.
\end{IEEEbiography}


\begin{IEEEbiography}[{\includegraphics[width=1in,height=1.25in,clip,
keepaspectratio]{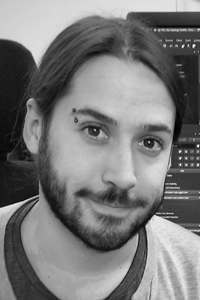}}]{Julien Tierny}
received the Ph.D. degree in Computer Science from the University
of Lille in 2008.
He is
a CNRS research director at
Sorbonne University. Prior to his CNRS tenure, he held a
Fulbright fellowship
and was a post-doctoral
researcher at the
University
of Utah.
His research expertise lies in topological methods for data analysis
and visualization.
He is the founder and lead developer of the Topology ToolKit
(TTK), an open source library for topological data analysis.
%
%
\end{IEEEbiography}




\clearpage
\setcounter{section}{0}
\appendices
\section*{Appendix}
\includegraphics{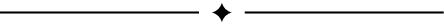}
\section{Topological Descriptors}

This appendix provides a brief description of the topological descriptors
considered in the main manuscript, namely the \emph{Persistence Diagram}
(PD, \autoref{sec_persistenceDiagrams}) and the \emph{Merge Tree} (MT),
specifically,
its
variant called \emph{Branch Decomposition Tree}, (BDT,
\autoref{sec_mergeTrees}).
We refer the
reader to textbooks \cite{edelsbrunner09} for
an
introduction to
computational topology.

\subsection{Persistence diagrams}
\label{sec_background_persistenceDiagrams}
\label{sec_persistenceDiagrams}
Given a piecewise linear (PL) scalar field $f_i : \domain \rightarrow 
\mathbb{R}$,
the \emph{sub-level set} of $f_i$, noted
$\sublevelset{{f_i}}(\isovalue)=\{p \in \domain~|
~f_i(p) < \isovalue\}$, is defined as the pre-image of  $(-\infty, \isovalue)$
by
$f_i$.
The \emph{super-level set} of $f_i$ is defined symmetrically:
$\superlevelset{{f_i}}(\isovalue)=\{p \in \domain~|
~f_i(p) > \isovalue\}$.
As $\isovalue$ continuously increases, the topology of
$\sublevelset{{f_i}}(\isovalue)$ changes at specific vertices of $\domain$,
called the \emph{critical points} of $f_i$ \cite{banchoff70}.
Critical points are classified by their \emph{index}
$\Index_i$: $0$ for minima, $1$ for $1$-saddles, $d-1$ for $(d-1)$-saddles and
$d$ for maxima.
In practice, $f_i$ is enforced to contain only isolated, non-degenerate
critical points \cite{edelsbrunner90, EdelsbrunnerHZ01}.
\journal{In} 3D,
connected components of $\sublevelset{{f_i}}(\isovalue)$ are created at local
minima and destroyed at $1$-saddles. One-dimensional cycles are created at
$1$-saddles and destroyed at $2$-saddles and voids are created at $2$-saddles
and destroyed at maxima.

The persistence diagram is a visual summary of the above topological features.
As shown in
\julien{Fig. 2 (main manuscript),}
it is closely related to the
\emph{merge tree}, which is the main topological
representation studied in
this paper. We first describe the persistence diagram
\journal{though}
as the metric
used in our work to measure distances between merge trees
\julien{(Sec. 2.2, main manuscript)}
\journal{generalizes}
an established metric between
persistence diagrams.

Specifically, in the domain, each topological feature of
$\sublevelset{{f_i}}(\isovalue)$ can be associated with a unique pair of
critical points $(c, c')$, corresponding to its \emph{birth} and \emph{death}.
The Elder rule \cite{edelsbrunner09} states that critical points can be
arranged
in pairs
according to this observation,
such that each
critical point appears in only one pair $(c, c')$, with $f_i(c) < f_i(c')$ and
$\Index_i(c) = \Index_i(c') - 1$.
For instance, if two connected components of $\sublevelset{{f_i}}(\isovalue)$
meet at a critical point $c'$, the \emph{younger} component
(created
last, in $c$) \emph{dies}, in favor of the \emph{older} one (created
first).

The
persistence diagram $\diagram(f_i)$ embeds each pair to a
single
point in 2D at coordinates $\big(f_i(c), f_i(c')\big)$.
The \emph{persistence} of a pair
is given by
its height $f_i(c') - f_i(c)$.
The
persistence diagram provides
a visual
overview of the features
of a dataset
\julien{(Fig. 2, main manuscript),}
where salient features stand out
from the diagonal while pairs corresponding to noise are located
near
the diagonal.



\subsection{Merge trees}
\label{sec_background_mergeTrees}
\label{sec_mergeTrees}
\revision{In the following, we introduce the main topological data
representation studied in this paper: the \emph{merge tree}. We also describe a
specific representation of the merge tree called the \emph{branch decomposition
tree},
which can be interpreted as a generalization of the extremum persistence
diagram, and
which plays a central role in the computation of distances between merge trees
\julien{(Sec. 2.2, main manuscript)}
.}

The \emph{join} tree, noted $\jointree(f_i)$, is a visual summary of the
connected components of $\sublevelset{{f_i}}(\isovalue)$ \cite{carr00}. It is
a 1-dimensional simplicial complex defined as the quotient space
$\jointree(f_i) = \domain / \sim$ by the equivalence relation $\sim$
which states that
$p_1$ and $p_2$ are equivalent
if
$f_i(p_1) = f_i(p_2)$ and if $p_1$ and $p_2$ belong to the same connected
component
of $\sublevelset{{f_i}}\big(f_i(p_1)\big)$.

The \emph{split} tree
\julien{(Fig. 2, main manuscript),}
noted $\splittree(f_i)$, is
defined symmetrically and describes the connected components of the super-level
set $\superlevelset{{f_i}}(\isovalue)$. Each of these two \emph{directed} trees
is called a
\emph{merge} tree (MT), noted generically $\mergetree(f_i)$ in the following.
Intuitively, these trees track the creation of connected components of
the sub
(or super) level sets at their leaves, and merge events at their
interior nodes.
To mitigate a phenomenon called \emph{saddle swap}, these trees are often
post-processed \cite{SridharamurthyM20, pont_vis21}, by merging adjacent
saddles in the tree if their relative difference in scalar value is smaller
than a threshold $\epsilon_1 \in [0, 1]$.

Merge trees
are often
visualized
via
a persistence-driven \emph{branch decomposition}
\cite{pascucci_mr04}, to make the persistence pairs captured by the tree stand
out. In this context, a \emph{persistent branch} is a monotone path on the tree
connecting the nodes corresponding to the creation and destruction (according
to the Elder rule, \autoref{sec_background_persistenceDiagrams})
of a connected component of sub (or super) level set. Then, the branch
decomposition provides a planar layout of the MT,
where each
persistent
branch is
represented as a vertical segment (center insets in
\julien{Fig. 2, main manuscript).}

The \emph{branch decomposition tree} (BDT),
noted
$\branchtree(f_i)$, is a directed tree
whose nodes are the persistent branches
captured
by the branch decomposition
and whose arcs denote adjacency relations between them in the
MT.
In
\julien{Fig. 2 (main manuscript)},
the BDTs (right
insets) can be interpreted as the dual of the branch decompositions
(center insets, with matching colors): each vertical segment in the
branch decomposition (center) corresponds to a node in the BDT (right) and each
horizontal segment (center, denoting an adjacency relation between branches)
corresponds to an arc in the BDT.
The BDT can be interpreted as a generalization of the extremum persistence
diagram:
like $\diagram(f_i)$, $\branchtree(f_i)$
describes the
population of (extremum) persistence pairs present in the data. However,
unlike the persistence diagram, it additionally captures adjacency relations
between them
\julien{(Fig. 2, main manuscript).}

Note that, the birth and death
of each persistent branch $b_i \in \branchtree(f_i)$, noted $(x_i, y_i)$, span
by construction an interval included in that of its parent $b_i' \in
\branchtree(f_i)$: $[x_i, y_i] \subseteq [x_i', y_i']$. This \emph{nesting
property} of BDTs \cite{pont_vis21} is a direct consequence of the Elder rule
(\autoref{sec_background_persistenceDiagrams}).

%

%

\cutout{Zhang \cite{zhang96} introduced a polynomial time algorithm for
computing a
constrained sequence of edit operations with minimal edit distance
(\autoref{eq_edit_distance_mapping}), and showed that the resulting distance is
indeed a
metric if each cost $\gamma$ for the above three edit operations is itself
a metric (non-negativity, identity, symmetry, triangle inequality).
Sridharamurthy et al. \cite{SridharamurthyM20} exploited this property to
introduce their metric,
by defining a distance-based cost model, inspired by the Wasserstein distance
between persistence diagrams (\autoref{sec_jackground_persistenceDiagrams}),
where $p_i$ and $p_j$ stand for the persistence pairs \emph{containing} the
nodes $n_i \in \mergetree(f_i)$ and $n_j \in \mergetree(f_j)$:
%
\begin{eqnarray}
  \gamma(n_i \rightarrow n_j) & = & \min\big(\pointMetric_\infty(p_i, p_j),
    \gamma(n_i \rightarrow \emptyset) + \gamma(\emptyset \rightarrow n_j)\big)\\
  \gamma(n_i \rightarrow \emptyset) & = & \pointMetric_\infty\big(p_i,
\projection(p_i)\big)\\
  \gamma(\emptyset \rightarrow n_j) & = & \pointMetric_\infty\big(p_j,
\projection(p_j)\big)
\end{eqnarray}
In our work, as described next, we introduce an alternative
edit distance
which adheres even further to the Wasserstein distance between persistence
diagrams,
to ease geodesic computation for merge
trees.}

\section{General Formulation of Basis Projection}
\label{sec_appendixGeneralProjection}

\begin{figure}
  \centering
  \includegraphics[width=0.95\linewidth]{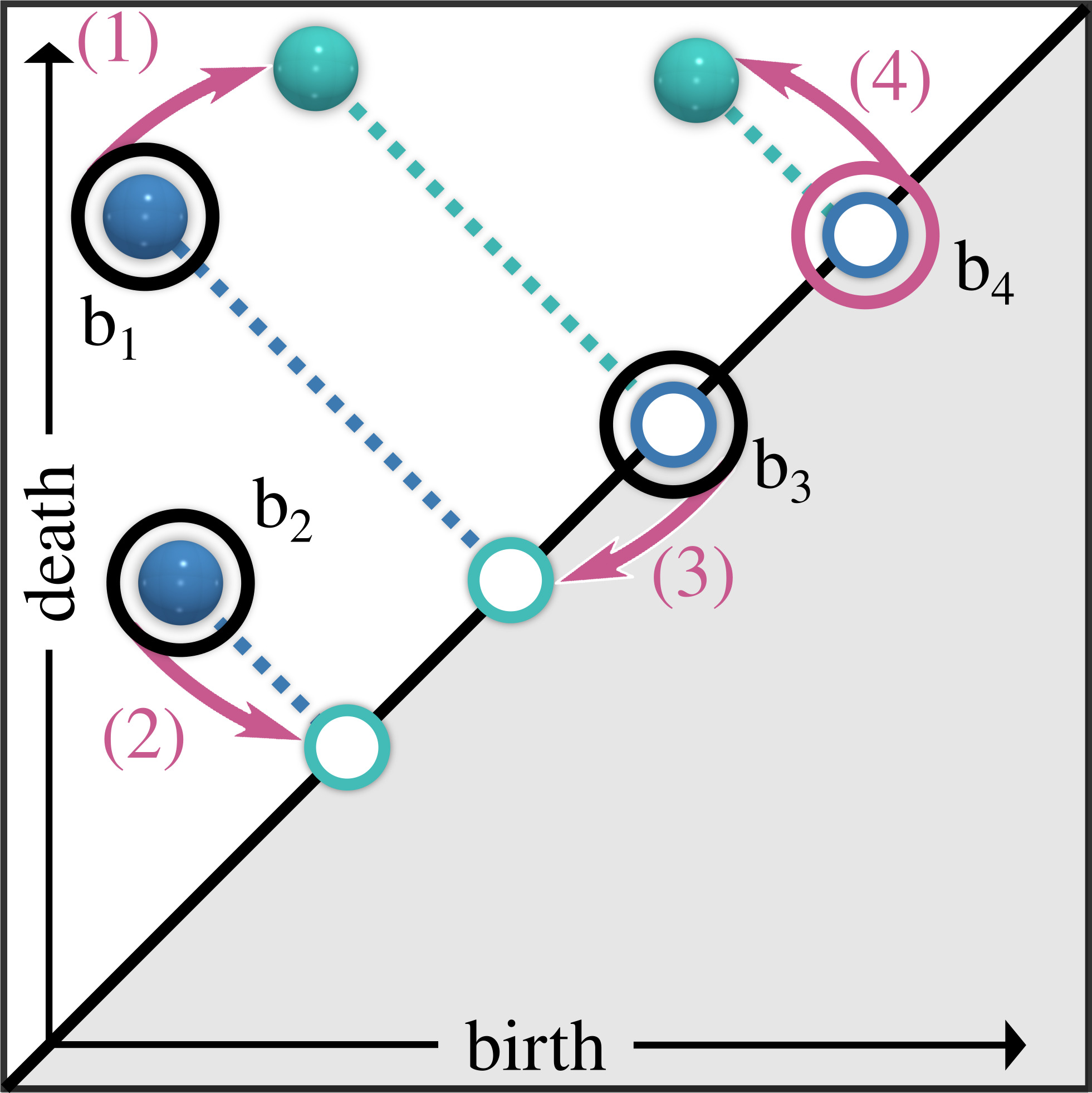}
  \caption{\minor{When considering an  assignment (purple arrows) in the
2D birth/death planes between \emph{augmented} BDTs, four cases can occur
(purple numbers). \textbf{Case \emph{(1)}:} an \emph{off-diagonal} branch
(blue dot) can be mapped to an \emph{off-diagonal} branch
(green dot). \textbf{Case \emph{(2)}:} an \emph{off-diagonal} branch
(blue dot) can
be mapped to an augmented \emph{diagonal} branch (green circle).
\textbf{Case
\emph{(3)}:} an augmented \emph{diagonal} branch (blue circle) can
be mapped to an augmented \emph{diagonal} branch (green circle).
\textbf{Case
\emph{(4)}:} an augmented \emph{diagonal} branch (blue circle) can
be mapped to an \emph{off-diagonal} branch (green dot). Sec. 4.2
(main manuscript) covers the cases \emph{(1)} and \emph{(2)}.
\autoref{sec_appendixGeneralProjection} generalizes this formulation to all
cases.
}}
  \label{fig_appendixGeneralProjection}
\end{figure}

The section 4.2 of the main manuscript presents an \emph{Assignment/Update}
algorithm to project an input BDT $\branchtree$ into a given BDT basis
$\pcaBasis(\bdtOrigin)$.

In the \emph{Assignment} phase, given an initial set of coefficients $\alpha
\in \mathbb{R}^{d'}$, the \emph{estimation} $\widehat{\branchtree}$ of
$\branchtree$ is given by :
\begin{eqnarray}
  \label{eq_estimation}
  \widehat{\branchtree} \leftarrow \bdtOrigin + \pcaBasis(\bdtOrigin)\alpha.
\end{eqnarray}

Given this estimation $\widehat{\branchtree}$, the assignment step first
evaluates the Wasserstein distance $\wassersteinTree(\branchtree,
\widehat{\branchtree})$.
For this, the optimal
assignment $\phi_*$ between $\branchtree$ and $\widehat{\branchtree}$ is
computed with regard to the Wasserstein distance (Eq. 1 of the main manuscript).

Then, given the optimal assignment $\phi_*$, the Wasserstein distance
$\wassersteinTree(\branchtree,
\widehat{\branchtree})$ can be re-written as:
\begin{eqnarray}
  \label{eq_projectionEnergy}
\wassersteinTree(\branchtree,
\widehat{\branchtree}) = \sum_{i = 1}^{|\branchtree|}
\begin{cases}
  0  \quad  \text{if both $b_i$ and $\phi_*(b_i)$ are on the diagonal,}\\
  ||b_i - \phi_*(b_i)||_2^2  \quad \text{otherwise.}
\end{cases}
\end{eqnarray}

The purpose of the subsequent \emph{Update} step is precisely to optimize
$\alpha$ in order to minimize the evaluation of
$\wassersteinTree(\branchtree,
\widehat{\branchtree})$ by the above equation.
For this, one needs to isolate from \autoref{eq_projectionEnergy} all the terms
involving $\alpha$ from those which do not.

In the simple case where $\phi_*$ describes a bijection between off-diagonal
points (case covered in the main manuscript), no branch of $\branchtree$
depends on $\alpha$. Then the isolation of the terms involving $\alpha$ is
simple: only the branches of $\widehat{\branchtree}$ depend on $\alpha$
(\autoref{eq_estimation}).

In the more general case, things are a bit more involved. As illustrated in
\autoref{fig_appendixGeneralProjection}, the computation of the optimal
assignment $\phi_* : \branchtree \rightarrow \widehat{\branchtree}$
(purple arrows) implies a pre-processing phase of \emph{augmentation} of
the 2D birth/death plane. As described in the section 2.2 of the main
manuscript, $\widehat{\branchtree}$ is augmented with the diagonal projections
of the branches of $\branchtree$ (\autoref{fig_appendixGeneralProjection},
green
circles) and $\branchtree$ is augmented with the diagonal projections of the
branches of $\widehat{\branchtree}$ (\autoref{fig_appendixGeneralProjection},
blue circles). This augmentation phase enables the modeling of the
destruction (or creation) of features during the assignment $\phi_*$ (between
the blue and green items,
\autoref{fig_appendixGeneralProjection}). Then, the following four cases can
occur (\autoref{fig_appendixGeneralProjection}):

\noindent
\textbf{Case \emph{(1)}:} An \emph{off-diagonal} branch $b_1 \in \branchtree$
is mapped to an \emph{off-diagonal} branch $\phi_*(b_1) \in
\widehat{\branchtree}$.
Then, the birth/death values of $b_1$ do not depend on
$\alpha$ and only the birth/death values of $\phi_*(b_1)$ do. This corresponds
to the case covered by the section 4.2
of the main manuscript.

\noindent
\textbf{Case \emph{(2)}:} An \emph{off-diagonal} branch $b_2 \in \branchtree$ is
mapped to a \emph{diagonal} branch $\phi_*(b_2) \in
\widehat{\branchtree}$. Then, the birth/death values of $b_2$
do not depend on
$\alpha$.
Then, this case is also
covered by
Sec. 4.2
(main manuscript).

\noindent
\textbf{Case \emph{(3)}:} A \emph{diagonal} branch $b_3 \in \branchtree$ is
mapped to a \emph{diagonal} branch $\phi_*(b_3) \in
\widehat{\branchtree}$. In that case, the ground distance $d_2\big(b_3,
\phi_*(b_3)\big)$ (section 2.2 of the main manuscript) is set to zero by
convention (first line of \autoref{eq_projectionEnergy}). This models the fact
that both $b_3$ and $\phi_*(b_3)$ are dummy features (with zero
persistence) and that their ground distance, which is arbitrary, should not be
taken into account in $\wassersteinTree(\branchtree,
\widehat{\branchtree})$. Therefore, we simply remove $b_3$ from
$\branchtree$ and $\phi_*(b_3)$ from $\widehat{\branchtree}$. This removal
discards this first line of \autoref{eq_projectionEnergy}, which can
then be re-written in the general form:
\begin{eqnarray}
\wassersteinTree(\branchtree,
\widehat{\branchtree}) = \sum_{i = 1}^{|\branchtree|} ||b_i -
\phi_*(b_i)||_2^2.
\end{eqnarray}
Then, with this removal,
the
diagonal-diagonal assignments do not constitute a special case anymore.

\noindent
\textbf{Case \emph{(4)}:} An \emph{off-diagonal} branch $b_4 \in \branchtree$ is
mapped to a \emph{diagonal} branch $\phi_*(b_4) \in
\widehat{\branchtree}$. In that case, $b_4$ turns out to be an \emph{augmented}
point of $\branchtree$. In \autoref{fig_appendixGeneralProjection}, these are
reported with circles while original (i.e. non-augmented) points are reported
with dots. Specifically, $b_4$ has been precisely inserted such that $b_4 =
\projection\big(\phi_*(b_4)\big)$:
\begin{eqnarray}
  \nonumber
  (b_4)_x = (b_4)_y =  \frac{1}{2} \Big( \big(\phi_*(b_4)\big)_x +
\big(\phi_*(b_4)\big)_y\Big).
\end{eqnarray}

Given \autoref{eq_estimation}, $\phi_*(b_4)$ can be re-written as:
\begin{eqnarray}
  \nonumber
  \phi_*(b_4) = o_4 + \pcaBasis(\bdtOrigin) \alpha,
\end{eqnarray}
where $o_4$ is a branch of $\bdtOrigin$. Then, the birth/death values of $b_4$
are:
\begin{eqnarray}
  \nonumber
  \begin{array}{l}
  (b_4)_x = (b_4)_y =  \frac{1}{2} \Big( \big( o_4 + \pcaBasis(\bdtOrigin)
\alpha \big)_x +
\big( o_4 + \pcaBasis(\bdtOrigin)
\alpha\big)_y\Big)\\
  (b_4)_x = (b_4)_y =
  \frac{1}{2} \big((o_4)_x + (o_4)_y\big)
  +
  \frac{1}{2} \Big( \big( \pcaBasis(\bdtOrigin)
\alpha \big)_x +
\big(\pcaBasis(\bdtOrigin)
\alpha\big)_y\Big).
  \end{array}
\end{eqnarray}
Then, it follows that $b_4$ can be re-written as:
\begin{eqnarray}
  \label{eq_diagonalPoint}
  b_4 = \projection(o_4) + \projection\big(\pcaBasis(\bdtOrigin)
\alpha\big).
\end{eqnarray}
From \autoref{eq_diagonalPoint}, it is clear that the coordinates of
$b_4$ are dependent on $\alpha$. In short, this is due to the fact that
$b_4$ has been purposely inserted in $\branchtree$ as the diagonal projection
of a branch of $\widehat{\branchtree}$ which, itself, depends on $\alpha$.
Thus, to account for this special case, we need to further isolate the terms of
\autoref{eq_diagonalPoint} depending on $\alpha$ (i.e.
$\projection\big(\pcaBasis(\bdtOrigin)$), as described next.

Similarly to \julien{Sec. 4.2}
(main manuscript), let
$\widehat{\branchtree}'$ be a vector representation of $\widehat{\branchtree}$.
Specifically $\widehat{\branchtree}'$ is a vector in
$\mathbb{R}^{2  |\widehat{\branchtree}|}$ which concatenates the
coordinates in the birth/death plane of each branch $b_i$ of
$\widehat{\branchtree}$.
$\widehat{\branchtree'}$  can be decomposed into $\bdtOrigin' +
\big(\pcaBasis(\bdtOrigin')\big)'\alpha$, where
$\big(\pcaBasis(\bdtOrigin')\big)'$ is a $(2|\widehat{\branchtree}|)\times d'$
matrix.
\journal{Also,}
let $\branchtree'$ be a similar vector
representation of $\branchtree$, \emph{but} where the entries have been
specifically re-ordered such that, for each of its 2D entries, we have:
\begin{eqnarray}
  \nonumber
  \label{eq_vectorReOrdering}
  (\branchtree')_i = \phi_*^{-1}\big((\widehat{\branchtree'})_i\big).
\end{eqnarray}

Then $\branchtree'$ can be decomposed as a sum of two vectors of $\mathbb{R}^{2
|\widehat{\branchtree}|}$:
\begin{eqnarray}
\nonumber
\branchtree' = \branchtree_1' + \branchtree_2',
\end{eqnarray}
such that $\branchtree_1'$ has all its entries
\journal{set to $0$}
except
those
covered by the above cases  \emph{(1)} and \emph{(2)}
(\autoref{fig_appendixGeneralProjection}), and that $\branchtree_2'$ has all its
entries
\journal{set to $0$}
except
those
covered by the above case \emph{(4)} (purple,
\autoref{fig_appendixGeneralProjection}).

Given \autoref{eq_diagonalPoint}, each non-zero entry $i$ of $\branchtree_2'$
can be re-written as:
\begin{eqnarray}
(\branchtree_2')_i = \projection(o_i) +
\projection \bigg( \Big(\big(\pcaBasis(\bdtOrigin)\big)'\alpha \Big)_i \bigg),
\end{eqnarray}
where $o_i$ is the $i^{th}$ entry of $\bdtOrigin$. Then, the vector
$\branchtree_2'$ can be further decomposed as follows:
\begin{eqnarray}
\nonumber
\branchtree_2' = \branchtree_3' + \branchtree_4',
\end{eqnarray}
such that each non-zero entry $i$ of $\branchtree_3'$ is equal to
$\projection(o_i)$ and each non-zero entry $i$ of $\branchtree_4'$ is equal to
$\projection \bigg( \Big(\big(\pcaBasis(\bdtOrigin)\big)'\alpha \Big)_i \bigg)$.

Let $\pcaBasis'_2$ be a $(2|\widehat{\branchtree}|)\times d'$
matrix such that:
\begin{eqnarray}
  \nonumber
  \branchtree_4' = \pcaBasis'_2 \alpha.
\end{eqnarray}

At this stage, we have:
\begin{eqnarray}
\label{eq_expressionAlpha}
\branchtree' = \branchtree_1' + \branchtree_3' +
\branchtree_4' = \branchtree_1' + \branchtree_3' +  \pcaBasis'_2 \alpha.
\end{eqnarray}
At this point, we managed to isolate the terms in $\branchtree'$ which are
dependent on $\alpha$ (i.e. $\pcaBasis'_2 \alpha$).
Then, similarly to the section 4.2 of the main manuscript, for a fixed optimal
assignment $\phi_*$, the Wasserstein distance $\wassersteinTree(\branchtree,
\widehat{\branchtree})$ can be re-written as an $L_2$ norm:
\begin{eqnarray}
\nonumber
\wassersteinTree(\branchtree,
\widehat{\branchtree}) = ||\branchtree' - \widehat{\branchtree'}||_2^2.
\end{eqnarray}

Then, given $\phi_*$, by using \autoref{eq_expressionAlpha}, the optimal
$\alpha_* \in \mathbb{R}^{d'}$ are:
\begin{eqnarray}
\nonumber
\begin{array}{lll}
\alpha_* &=& \argmin_\alpha ||\branchtree'- \widehat{\branchtree'}||_2^2\\
\alpha_* &=& \argmin_\alpha||\branchtree_1' + \branchtree_3' +
\pcaBasis'_2 \alpha - \Big(\bdtOrigin' +
\big(\pcaBasis(\bdtOrigin')\big)'\alpha\Big)||_2^2\\
\alpha_* &=& \argmin_\alpha||\branchtree_1' + \branchtree_3'
- \bdtOrigin'
-  \bigg(\Big(\big(\pcaBasis(\bdtOrigin')\big)' -
\pcaBasis'_2 \Big) \alpha\bigg)||_2^2.
\end{array}
\end{eqnarray}
Then, similarly to the Euclidean case (Eq. 4 of the main manuscript), it
follows then that $\alpha_*$ can be expressed as a function of the
pseudoinverse of $\Big(\big(\pcaBasis(\bdtOrigin')\big)' -
\pcaBasis'_2 \Big)$:
\begin{eqnarray}
  \label{eq_generalExpression}
  \alpha_* = \Big(\big(\pcaBasis(\bdtOrigin')\big)' -
\pcaBasis'_2 \Big)^+ (\branchtree_1' + \branchtree_3'
- \bdtOrigin').
\end{eqnarray}

In short, the general expression of the optimal coefficients $\alpha_*$
(\autoref{eq_generalExpression}) is a generalization of the Eq. 13 of the main
manuscript, such that the branches of $\branchtree$ dependent on $\alpha$ (case
\emph{(4)}) have been integrated within the pseudoinverse operation.

\section{Computational Parameters}
\label{sec_computationalAspects}
The Wasserstein distance $\wassersteinTree$ is subject to three parameters
($\epsilon_1$, $\epsilon_2$ and $\epsilon_3$, 
\julien{Sec. 2.2, main manuscript}),
for
which
we use the recommended default values
($\epsilon_1 = 0.05$, $\epsilon_2 = 0.95$, $\epsilon_3 = 0.9$,
\cite{pont_vis21}) when considering merge trees (MT-WAE).
In contrast, when considering persistence diagrams, we switch
$\epsilon_1$ to $1$ ($\epsilon_2$ and $\epsilon_3$ do not have any effect then)
and
$\wassersteinTree$
becomes equivalent to
$\wasserstein{2}$ 
\julien{(Sec. 2.2, main manuscript)}. Then
our framework
%
%
computes a Wasserstein Auto-Encoder
of extremum persistence diagrams (PD-WAE for short).

Our main algorithm is subject to
meta-parameters. $n_{it}$ 
 stands for the number of iterations in our 
basis projection procedure 
\julien{(Sec. 4.2, main manuscript)}.
In practice, we set 
$\julien{n_{it} = 2}$. 

The number, size and dimensionality of the layers of our 
MT-WAE are also meta-parameters.
\julien{Unless specified otherwise,}
we use 
only 
one encoding layer and one decoding layer, i.e. 
$\sizeEncoding = \sizeDecoding = 1$, with $d_{\sizeEncoding} = 2$ (for 
dimensionality reduction purposes) and $d_{\sizeEncoding+\sizeDecoding} = 16$.
For data reduction purposes and computational cost control, we also restrict 
the size of the origins and bases of the sub-layers of our network. Let 
$|\branchtreeSet|$ be the total number of branches in the ensemble, i.e. 
$|\branchtreeSet| = \sum_{i = 1}^{N}|\branchtree(f_i)|$. 
We restrict the 
maximum 
size of the following origins as follows:
%
%
%
%
$|\bdtOrigin_1^{in}| \leq 0.2 |\branchtreeSet|$,
$|\bdtOrigin_1^{out}| \leq 0.1 |\branchtreeSet|$,
$|\bdtOrigin_2^{in}| \leq 0.1 |\branchtreeSet|$,
$|\bdtOrigin_2^{out}| \leq 0.2 |\branchtreeSet|$. 
This origin size control also implicitly restricts 
the size of the corresponding bases. Overall, 
when 
integrating all these 
constraints, 
the 
number of variables in 
our  networks is bounded by $\big((d_{\sizeEncoding} + 1) \times 
2 \times (0.2 + 0.1) + (d_{\sizeDecoding} + 1) \times 2 \times (0.1 + 0.2)\big) 
\times |\branchtreeSet| = 12 |\branchtreeSet|$. In practice,
our 
networks optimized \julien{$68,902$} variables on average (per ensemble).

\section{Penalty Terms}

The flexibility of our framework allows to improve the quality
\journal{of the dimensionality reduction computed}
by MT-WAE.
Here, we introduce two penalty terms aiming at
\emph{(1)} improving the preservation of the Wasserstein metric
$\wassersteinTree$ and \emph{(2)} improving the preservation of the clusters of
BDTs.

\subsection{Metric penalty term}


In order to improve the preservation
of the Wasserstein metric $\wassersteinTree$ in the latent space, and hence in
the 2D layout, we introduce the following penalty term $P_M(\theta)$:
\begin{eqnarray}
  \nonumber
  \label{eq_metricPenalty}
P_M(\theta) = \sum_{\forall i \in \{1, \dots, N\}} \sum_{\forall j \neq i
\in \{1, \dots, N\}}
\Big( \wassersteinTree\big(\branchtree(f_i), \branchtree(f_j)\big) -
||\alpha^i_{\sizeEncoding} - \alpha^j_{\sizeEncoding} ||_2 \Big)^2.
\end{eqnarray}
Concretely, given two BDTs $\branchtree(f_i)$ and $\branchtree(f_j)$,
$P_M(\theta)$ penalizes the variations between their Wasserstein distances and
the Euclidean distances between their coordinates $\alpha^i_{\sizeEncoding}$
and $\alpha^j_{\sizeEncoding}$ in the latent space.

The integration of the penalty term $P_M(\theta)$ in our optimization
algorithm
(\julien{Sec. 4, main manuscript})
is
straightforward.
The Wasserstein distance matrix,
which stores at its entry $(i, j)$ the distance
$\wassersteinTree\big(\branchtree(f_i), \branchtree(f_j)\big)$, is
computed in a pre-processing stage.
Since this matrix is a constant during the
optimization,
the expression of $P_M(\theta)$
only involves
basic operations
%
supported by automatic differentiation.
Then, given a blending weight $\lambda_M \in [0, 1]$ (in practice, we set
$\lambda_M = 1$),
the penalty term $\lambda_M P_M(\theta)$ is simply added
to the expression of the
reconstruction energy $\mtPgaError(\theta)$
(\julien{Eq. 14, main manuscript}).
Next, the corresponding gradient is evaluated by automatic
differentiation and the overall energy is optimized by gradient descent
\cite{KingmaB14} , as
originally described in
\julien{Sec. 4.5 (main manuscript)}.

\subsection{Clustering penalty term}


We introduce an additional penalty
term to improve the preservation of the \emph{natural}
clusters of BDTs in the latent space, and hence in the 2D layout.
Let $C \in \mathbb{R}^{kN}$ be the vector modeling the input $k$ clusters:
the entry $(ik) + j$ of this vector is equal to $1$ if the BDT
$\branchtree(f_i)$ belongs to the cluster $j$, $0$ otherwise.
This \emph{input clustering vector} can be provided either by a pre-defined
ground-truth, by interactive user inputs or by any automatic clustering
algorithm. In our experiments, to construct this vector $C$, we used the
extension of the $k$-means clustering to the Wasserstein metric space of merge
trees \cite{pont_vis21}.
Next, in the latent space, we consider the
classic
$k$-means algorithm \cite{elkan03, celebi13}, where each BDT
$\branchtree(f_i)$ is clustered according to its latent coordinates
$\alpha^i_{\sizeEncoding} \in \mathbb{R}^2$.
This yields a set of $k$ centroids in the 2D latent space $c_l \in \mathbb{R}^2$
with $l \in \{0, 1, \dots, k-1\}$.
To evaluate the similarity between
this clustering and the input clustering vector $C$, we use the celebrated
\emph{SoftMax} function \cite{goodfellow16}. Specifically, we consider the
\emph{latent clustering vector} $C' \in \mathbb{R}^{kN}$, such that the entry
$(C')_{(ik) + j}$ denotes the probability that the 2D point
$\alpha^i_{\sizeEncoding}$ belongs to the cluster $j$ ($\beta$ is set to
$5$):
\begin{eqnarray}
  \nonumber
  \label{eq_softmax}
  (C')_{(ik) + j} = {\frac{e^{-\beta ||\alpha^i_{\sizeEncoding} -
c_j||_2}}{\sum_{l = 0}^{k-1} e^{-\beta ||\alpha^i_{\sizeEncoding} -
c_l||_2}}}.
\end{eqnarray}
Then, the clustering penalty term $P_C(\theta)$ is given by
the Kullback-Leibler divergence
(a standard
indicator for probability similarity):
\begin{eqnarray}
\nonumber
\label{eq_clusteringPenalty}
P_C(\theta) = \julien{KL(C, C')} =  \sum_{i = 0}^{kN-1} C(i)
log\Big({\frac{C(i)}{C'(i)}}\Big).
\end{eqnarray}
Similarly  to the metric penalty term, given a blending weight $\lambda_C \in
[0, 1]$ (in practice, we set $\lambda_C = 1$), the penalty term $\lambda_C
P_C(\theta)$ is added to the expression of
the reconstruction energy $\mtPgaError(\theta)$
(\julien{Eq. 14, main manuscript}).
The corresponding gradient is estimated by
automatic differentiation and the overall energy is optimized by gradient
descent \cite{KingmaB14}, as originally described in
\julien{Sec. 4.5 (main manuscript)}.

\begin{figure*}
\makebox[\linewidth]{
\centering
\includegraphics[width=0.85\linewidth]{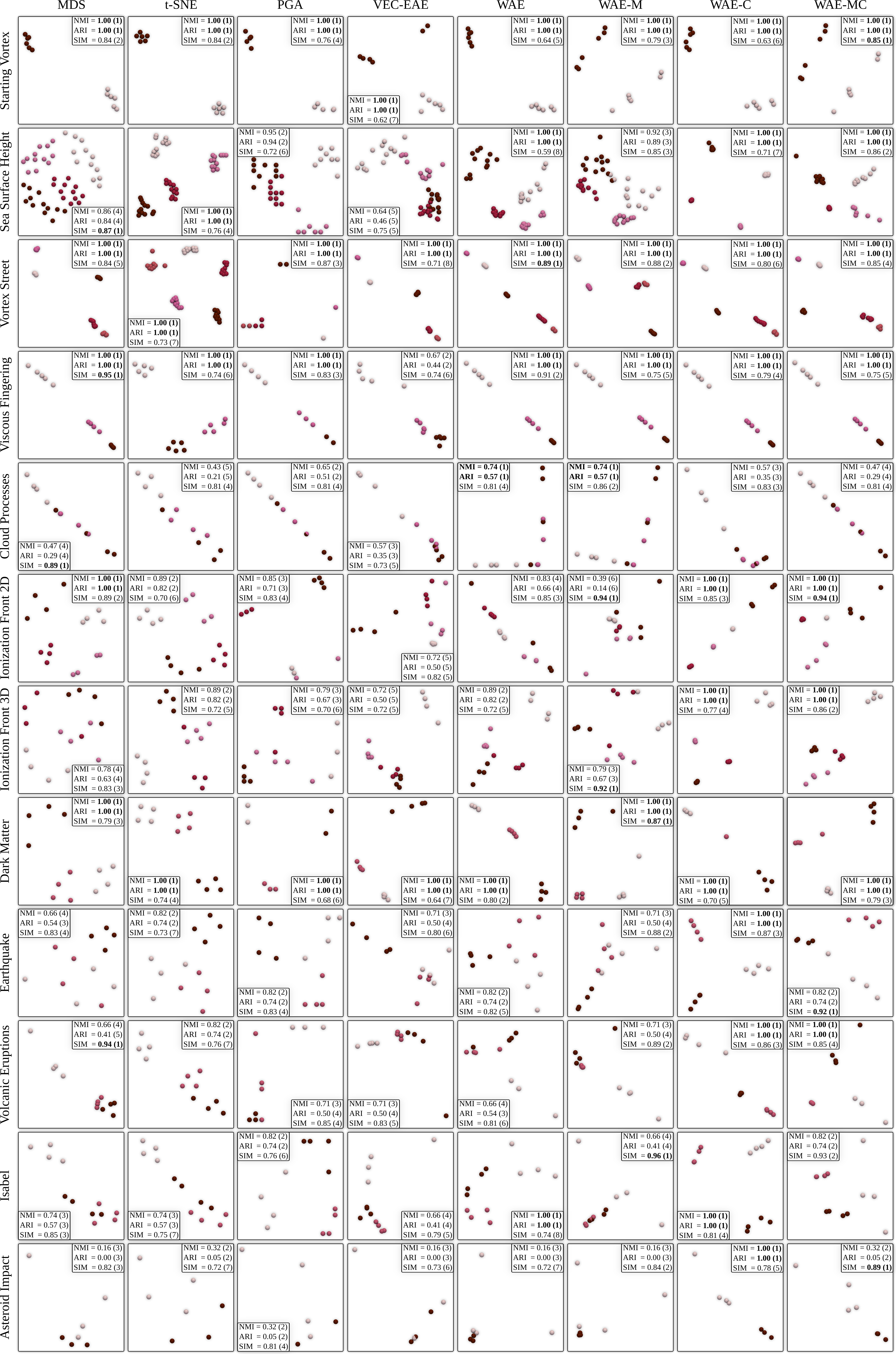}
}
\mycaption{Comparison of planar layouts for typical dimensionality
reduction techniques on all our merge tree ensembles. The color encodes the
classification ground-truth \cite{pont_vis21}. For each quality score,
the best
value appears bold and the rank of the score among all methods is in
parenthesis.}
\label{fig_embeddingAll}
\end{figure*}

\section{Data Reduction Experiments}

\begin{table}
  \caption{Comparison of the Average Relative Reconstruction (ARR) Error,
between
PD-PGA \cite{pont_tvcg23} ($d_{max} = 3$ and $N_1 \leq
0.1|\branchtreeSet|$) and our approach PD-WAE ($d_{\sizeEncoding} = 3$ and
$|\bdtOrigin_{\sizeEncoding}^{out}| \leq 0.1|\branchtreeSet|$ ), for identical
compression factors. Bold numbers in the \emph{Ratio} column indicate
instances where PD-WAE achieved a lower (hence better) reconstruction error.}
  \centering
  \scalebox{0.675}{
    \begin{tabular}{|l|r|r|r||r|r||r|}
      \hline
      \textbf{Dataset} & $N$ & $|\branchtree|$ & Compression &
\multicolumn{2}{c||}{ARR Error} & Ratio \\
       & & & Factor & PD-PGA \cite{pont_tvcg23} & PD-WAE & \\
      \hline
      Asteroid Impact (3D) & 7 & 1,295 & 7.36 & 0.01 & 0.01 & \textbf{0.73} \\
      Cloud processes (2D) & 12 & 1,209 & 7.99 & 0.12 & 0.10 & \textbf{0.81} \\
      Viscous fingering (3D) & 15 & 118 & 7.87 & 0.02 & 0.01 & \textbf{0.67} \\
      Dark matter (3D) & 40 & 316 & 8.68 & 0.01 & 4e-03 & \textbf{0.39} \\
      Volcanic eruptions (2D) & 12 & 811 & 7.56 & 0.02 & 0.01 & \textbf{0.36} \\
      Ionization front (2D) & 16 & 135 & 8.01 & 0.03 & 0.02 & \textbf{0.69} \\
      Ionization front (3D) & 16 & 763 & 7.68 & 0.05 & 0.03 & \textbf{0.65} \\
      Earthquake (3D) & 12 & 1,203 & 7.59 & 0.04 & 0.02 & \textbf{0.51} \\
      Isabel (3D) & 12 & 1,338 & 7.58 & 0.08 & 0.08 & \textbf{0.91} \\
      Starting Vortex (2D) & 12 & 124 & 7.39 & 0.01 & 3e-03 & \textbf{0.51} \\
      Sea Surface Height (2D) & 48 & 1,787 & 24.66 & 0.16 & 0.16 & 1.02 \\
      Vortex Street (2D) & 45 & 23 & 15.83 & 2e-03 & 9e-04 & \textbf{0.40} \\
      \hline
    \end{tabular}
  }
  \label{tab_compresson_pd}
\end{table}

\begin{table}
\caption{Comparison of the Average Relative Reconstruction (ARR) Error, between
MT-PGA \cite{pont_tvcg23} ($d_{max} = 3$ and $N_1 \leq
0.1|\branchtreeSet|$) and our approach MT-WAE ($d_{\sizeEncoding} = 3$ and
$|\bdtOrigin_{\sizeEncoding}^{out}| \leq 0.1|\branchtreeSet|$ ), for identical
compression factors. Bold numbers in the \emph{Ratio} column indicate
instances where MT-WAE achieved a lower (hence better) reconstruction error.}
  \centering
  \scalebox{0.675}{
    \begin{tabular}{|l|r|r|r||r|r||r|}
      \hline
      \textbf{Dataset} & $N$ & $|\branchtree|$ & Compression &
\multicolumn{2}{c||}{ARR Error} & Ratio \\
       & & & Factor & MT-PGA \cite{pont_tvcg23} & MT-WAE & \\
      \hline
      Asteroid Impact (3D) & 7 & 1,295 & 13.68 & 0.13 & 0.12 & \textbf{0.93} \\
      Cloud processes (2D) & 12 & 1,209 & 13.84 & 2e-04 & 7e-07 & \textbf{3e-03} \\
      Viscous fingering (3D) & 15 & 118 & 13.21 & 8e-04 & 7e-07 & \textbf{8e-04} \\
      Dark matter (3D) & 40 & 316 & 15.09 & 2e-04 & 2e-05 & \textbf{0.08} \\
      Volcanic eruptions (2D) & 12 & 811 & 13.83 & 0.01 & 2e-03 & \textbf{0.41} \\
      Ionization front (2D) & 16 & 135 & 13.44 & 0.19 & 0.14 & \textbf{0.78} \\
      Ionization front (3D) & 16 & 763 & 13.89 & 0.24 & 0.22 & \textbf{0.92} \\
      Earthquake (3D) & 12 & 1,203 & 14.07 & 0.14 & 0.10 & \textbf{0.75} \\
      Isabel (3D) & 12 & 1,338 & 14.03 & 3e-03 & 2e-03 & \textbf{0.72} \\
      Starting Vortex (2D) & 12 & 124 & 11.92 & 2e-04 & 2e-06 & \textbf{0.01} \\
      Sea Surface Height (2D) & 48 & 1,787 & 14.36 & 0.23 & 0.22 & \textbf{0.92} \\
      Vortex Street (2D) & 45 & 23 & 20.27 & 3e-04 & 9e-05 & \textbf{0.26} \\
      \hline
    \end{tabular}
  }
  \label{tab_compresson_mt}
\end{table}

Tables \ref{tab_compresson_pd} and \ref{tab_compresson_mt} report a comparison
between the reconstruction error generated by our Wasserstein
Auto-Encoder (WAE) approach and the Principal Geodesic Analysis (PGA) approach
by Pont et al. \cite{pont_tvcg23}, for the application to data reduction (Sec.
5.1 of the main manuscript), in the case of persistence diagrams
(\autoref{tab_compresson_pd}) and merge trees (\autoref{tab_compresson_mt}).

Specifically,
we compute the
reconstruction error
of each input BDT
$\branchtree(f_i)$
via the   distance
$\wassersteinTree$
to its reconstruction (computed by the method under consideration, PGA or WAE).
To be comparable across ensembles, this distance is then divided
by the maximum
$\wassersteinTree$ distance observed among two input BDTs in the
ensemble. Finally, this \emph{relative} reconstruction error is \emph{averaged}
over all the
BDTs of the
ensemble.

To enable a fair comparison, we set the number of axis of PGA,
noted $d_{max}$, to 3 (as reported in the original data reduction description
\cite{pont_tvcg23}) and we set the number of dimensions in the latent space of
WAE to the same value (i.e. $d_{\sizeEncoding} = 3$). We also set the
maximum size of the PGA origin, noted $N_1$, to $0.1|\branchtreeSet|$, where
$|\branchtreeSet|$ is the total number of branches in the ensemble, i.e.
$|\branchtreeSet| = \sum_{i = 1}^{N}|\branchtree(f_i)|$.
Similarly, for
WAE, we set the maximum size of the latent output origin
$|\bdtOrigin_{\sizeEncoding}^{out}|$ to $0.1 |\branchtreeSet|$.

For both methods (PGA and WAE), the compression factor is
fixed to a common value on a per ensemble basis. As discussed in the
section 5.1 of the main manuscript, the compression factor of WAE is controlled
by adjusting, for the last decoding layer, its dimensionality  noted
$d_{\sizeEncoding + \sizeDecoding}$, and the maximum size of its output
origin, noted $|\bdtOrigin_{\sizeEncoding+\sizeDecoding}^{out}|$.


Both tables show that
WAE
clearly
outperforms PGA \cite{pont_tvcg23} in terms of average relative
reconstruction error, with an average improvement
of
\julien{$37\%$} for persistence diagrams, and \julien{$52\%$} for merge trees.


Finally, note that for each ensemble, the merge tree based clustering
\cite{pont_vis21} computed from the input BDTs is strictly identical to the
clustering computed from the reconstructed BDTs. This confirms the viability of
our reconstructed BDTs, and their usability for typical visualization and
analysis tasks.
 
\section{Dimensionality Reduction Experiments}

\autoref{fig_embeddingAll} extends the Figure \journal{12} of the main
manuscript to all
our test ensembles.
It confirms visually the
conclusions of the table of
aggregated scores (Table \julien{2} of the main manuscript).

In particular, it confirms that WAE behaves as a \emph{trade-off} between
the respective advantages of standard techniques, such as MDS \cite{kruskal78}
and t-SNE \cite{tSNE}.
Specifically, MDS is known to preserve the
input metric
well, while t-SNE tends to better preserve the global structure of the data
(i.e. the ground-truth classification), at the expense of metric violation.
Our approach (WAE) provides a trade-off between these two extreme
behaviors: \emph{(i)} it improves over MDS in terms of structure preservation
(it provides equivalent or better NMI/ARI scores for 11 out of 12 ensembles) and
\emph{(ii)} it improves over t-SNE in terms of metric preservation (it provides
an equivalent or better SIM score for 9 out of 12 ensembles). WAE also
outperforms VEC-AE and improves PGA on most ensembles. Finally, the combination
of our two penalty terms, WAE-MC, simultaneously outperforms MDS on metric
preservation and t-SNE on cluster preservation (hence maximizing all criteria
at once), for 8 of the 12 ensembles.

\section{\journal{Empirical Stability Evaluation}}
\label{sec_stability}

\journal{As documented in the original paper  \cite{pont_vis21} introducing the
Wasserstein distance between merge trees ($\wassersteinTree$),
saddle swap instabilities in the merge trees are commonly addressed with a
\emph{saddle-merging} pre-processing \cite{SridharamurthyM20, pont_vis21}.
%
%
This procedure
consists in
moving each branch $b$ up the BDT $\branchtree(f)$, if
its saddle is too \emph{close} to that of its parent branch (i.e. closer in
normalized $f$ values than a threshold $\epsilon_1$, see \cite{pont_vis21}). As
documented by Pont et al. with practical stability evaluations
(see Fig. 14 of \cite{pont_vis21}), this simple saddle-merging pre-processing
drastically improves in practice the robustness of the metric
$\wassersteinTree$ to
additive noise.
%
Thus, this saddle-merging pre-processing is of paramount importance for the
practical usage of $\wassersteinTree$ on real-life datasets and Pont et al.
recommend to use $\epsilon_1 = 0.05$ as a default value. Note that this
parameter $\epsilon_1$ acts as a control knob, which balances the
practical stability of the metric with its discriminative power (for
$\epsilon_1 = 1$, $\wassersteinTree = \wasserstein{2}$).
}

\journal{In this appendix, we study the practical stability of our non-linear
framework for merge tree encoding (WAE) to additive noise, in order to document
the impact of the underlying metric's stability on the outcome of the analysis.}

\begin{figure*}
\makebox[\linewidth]{
\centering
\includegraphics[width=\linewidth]{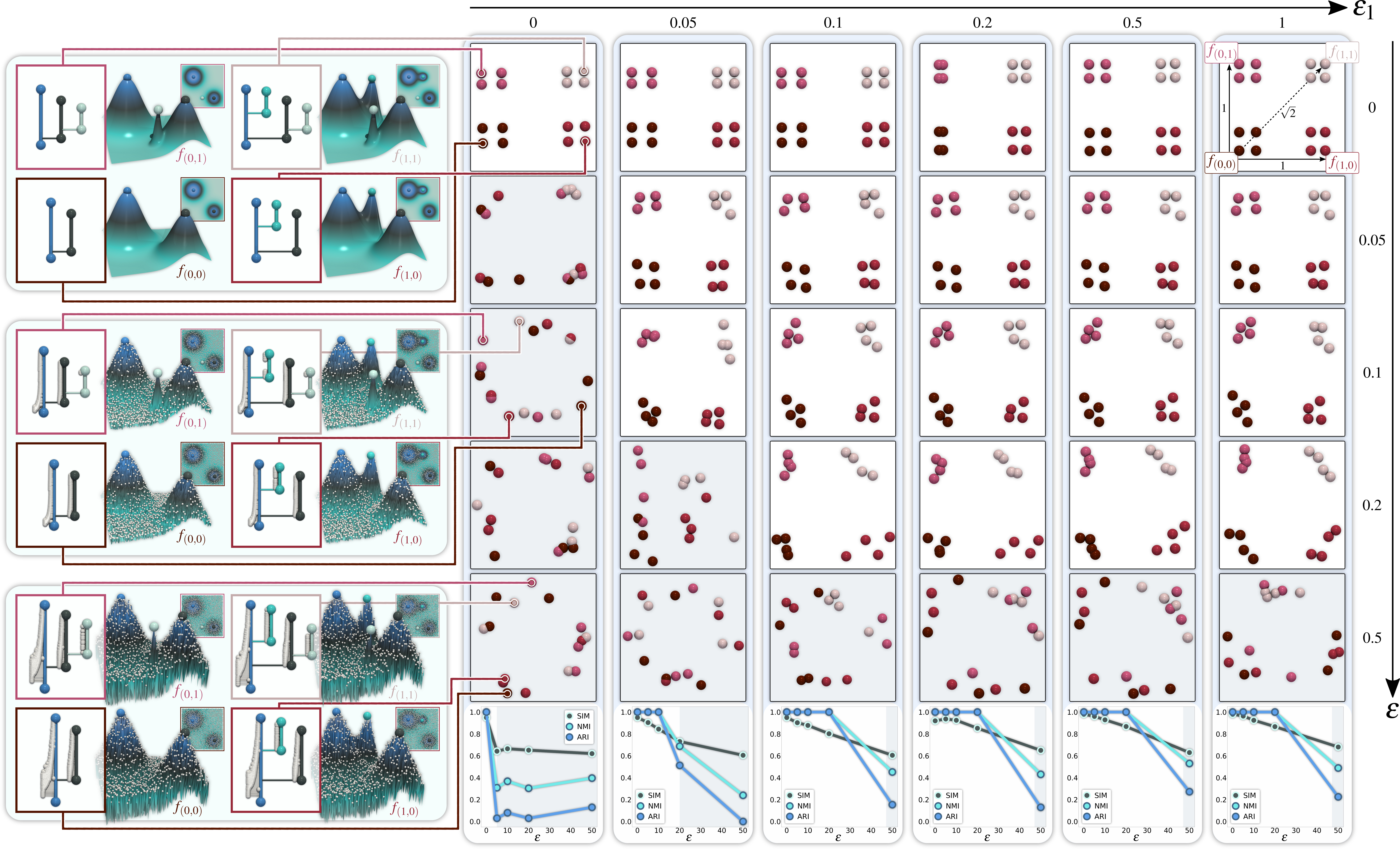}
}
\caption{\journal{\textbf{Empirical stability evaluation:} a synthetic ensemble
of
sixteen 2D scalar fields is specifically designed by sampling a 2D basis of
Gaussian mixtures,
with a
controlled
parameterization
(see \autoref{groundTruthData} for a detailed specification). This yields a
\emph{ground-truth} parameterization and classification of the ensemble (four
clusters: dark red, red, pink, light pink). Five versions of this
ensemble are created, for increasing levels of additive noise (from $\epsilon =
0$ to $\epsilon = 0.5$,
top to bottom). For each ensemble, a 2D
layout is generated by our non-linear framework WAE (right insets), for
increasing values of the parameter $\epsilon_1$
from left to right
(i.e. from the strict Wasserstein distance between merge trees,
$\wassersteinTree$ for $\epsilon_1 =
0$, to progressive
blends towards the Wasserstein distance between persistence diagrams,
$\wasserstein{2}$ for $\epsilon_1 = 1$).
%
In the 2D layout and the quality scores (bottom curves), a grey background
indicates an unstable computation (i.e. $NMI$ and $ARI$ are both below $1$).
For the default recommended value of the parameter $\epsilon_1$ ($0.05$
\cite{pont_vis21}), WAE with $\wassersteinTree$ recovers well the ground-truth
parameterization and classification (similarly to $\wasserstein{2}$), up to a
level of additive noise of $\epsilon = 0.1$.
For $\epsilon_1 \geq 0.1$, the 2D layouts generated by WAE provide a similar
level
of robustness for $\wassersteinTree$ and
$\wasserstein{2}$ (bottom curves).
}}
\label{fig_stability}
\end{figure*}

\subsection{\journal{Setup}}
\label{groundTruthData}

\journal{For this experiment, we specifically generated a synthetic ensemble,
in order to control both its intrinsic parameterization and its classification.
For this, we proceeded as follows.}

\journal{First, four 2D scalar fields
(\autoref{fig_stability}, top left inset) were generated by sampling a 2D basis
of Gaussian mixtures with controlled parameterization. Specifically, the scalar
field being the origin of the basis has two hills ($f_{(0, 0)}$, dark red
frame, top left
inset). The extremity of the first (horizontal) axis ($f_{(1, 0)}$, red frame,
top left inset) has exactly the same hills, but with a first \emph{additional}
maximum (cyan sphere). The extremity of the second (vertical) axis ($f_{(0,
1)}$, pink frame, top left inset) has a second additional maximum (white
sphere). Finally, the fourth dataset ($f_{(1, 1)}$, light pink frame, top left
inset) has both extra maxima (cyan and white spheres).
These
Gaussian mixtures were generated by adjusting the height of the additional
maxima (cyan and white spheres, \autoref{fig_stability}) such that their
diagrams describe a square on the Wasserstein metric space (see the top right
2D layout of \autoref{fig_stability}):}

\journal{\begin{eqnarray}
\nonumber
 \wasserstein{2}\big(\diagram(f_{(0, 0)}, \diagram(f_{(1, 0)})\big) &=&
\wasserstein{2}\big(\diagram(f_{(1, 0)}, \diagram(f_{(1, 1)})\big) \\
\nonumber
~ & = &\wasserstein{2}\big(\diagram(f_{(1, 1)}, \diagram(f_{(0, 1)})\big) \\
\nonumber
~ & = &\wasserstein{2}\big(\diagram(f_{(0, 1)}, \diagram(f_{(0, 0)})\big) \\
\nonumber
~ & =& 1,
\end{eqnarray}
and:
\begin{eqnarray}
\nonumber
 \wasserstein{2}\big(\diagram(f_{(0, 0)}, \diagram(f_{(1, 1)})\big) &=&
\wasserstein{2}\big(\diagram(f_{(1, 0)}, \diagram(f_{(0, 1)})\big) \\
\nonumber
~ & =& \sqrt{2}.
\end{eqnarray}
}

\journal{Next, we repeated this square generation process, around each corner
of the above square, but this time with a smaller side length (equal
to $0.15$ in the Wasserstein metric space, instead of $1$).}

\journal{Overall, this
results in
a total of $16$ scalar fields, specifically organized
along a \emph{ground-truth} $2$-dimensional parameterization of the Wasserstein
metric space, with a natural \emph{ground-truth} classification (corresponding
to the closest corner of the 2D grid, see the top right 2D layout of
\autoref{fig_stability}):
\begin{itemize}
 \item Class 1 (bottom left corner, dark red spheres in
\autoref{fig_stability}):
    \begin{itemize}
      \item $f_{(0, 0)}$, $f_{(0.15, 0)}$, $f_{(0.15, 0.15)}$, $f_{(0, 0.15)}$;
    \end{itemize}
 \item Class 2 (bottom right corner, red spheres in \autoref{fig_stability}):
    \begin{itemize}
    \item $f_{(0.85, 0)}$, $f_{(1, 0)}$, $f_{(1,
0.15)}$, $f_{(0.85, 0.15)}$;
    \end{itemize}
  \item Class 3 (top right corner, light pink spheres in
\autoref{fig_stability}):
    \begin{itemize}
    \item $f_{(0.85, 0.85)}$, $f_{(1, 0.85)}$, $f_{(1, 1)}$, $f_{(0.85, 1)}$;
    \end{itemize}
  \item Class 4 (top left corner, bright pink spheres in
\autoref{fig_stability}):
    \begin{itemize}
    \item $f_{(0, 0.85)}$, $f_{(0.15, 0.85)}$, $f_{(0.15, 1)}$, $f_{(0, 1)}$;
    \end{itemize}
\end{itemize}
Given the above ground-truth parameterization, we call the \emph{ground-truth
distance matrix}, noted $\mathbb{D}$, the matrix defined such that each of its
entries $(i, j)$ is equal to $\wasserstein{2}\big(\diagram(f_i),
\diagram(f_j)\big)$.
}

\journal{Next, we generated additional versions of the above ensemble, by
introducing a random additive noise
in the scalar fields, with a
 control on the maximum normalized amplitude $\epsilon \in [0, 1]$ (i.e. the
maximum amplitude of the noise is a fraction $\epsilon$ of the global function
range of the input scalar field). Specifically, we considered the noise levels
$\epsilon \in \{0, 0.05, 0.1, 0.2, 0.5\}$. Overall, this results in $5$
ensembles of $16$ scalar fields each.}

\subsection{\journal{Protocol}}
\journal{Given the above ensembles, we first consider our non-linear framework
for persistence diagrams, namely PD-WAE. Specifically, we generated, for each
noise level, a 2D layout of the ensemble with PD-WAE (see Sec. 5.2,
main manuscript). This is shown in the rightmost column of
\autoref{fig_stability} ($\epsilon_1 = 1$). We quantitatively evaluate the
quality of this 2D layout along two criteria: metric preservation and cluster
preservation.}

\journal{First, given the 2D layout of the ensemble, we compute a
distance matrix $D$ in 2D, which we compare to the \emph{ground-truth} distance
matrix $\mathbb{D}$ (see \autoref{groundTruthData}) with the \emph{SIM}
indicator \cite{pont_tvcg23} (which varies between $0$ and $1$, $1$ being
optimal). Second, given the 2D layout of the ensemble, we compute a $k$-means
clustering in 2D (with $k = 4$) and we compare the resulting classification to
the \emph{ground-truth} classification with the \emph{NMI} and \emph{ARI}
indicators (which vary between $0$ and $1$, $1$ being optimal).}

\journal{To study the stability to additive noise of our framework when
considering the Wasserstein distance between merge trees, we have replicated
the above experiment for $5$ more values of the control parameter $\epsilon_1$
(in \autoref{fig_stability}, from left to right:
$0$, $0.05$, $0.1$, $0.2$ and $0.5$).
Overall this results in the 2D array represented in
\autoref{fig_stability} where each column denotes a specific value of the
control parameter $\epsilon_1$ and where each line denotes a specific noise
level $\epsilon$.}

\subsection{\journal{Analysis}}
\journal{In the absence of noise ($\epsilon = 0$, top row) and for arbitrary
values of the parameter $\epsilon_1$, our non-linear WAE framework manages to
produce a 2D layout of the ensemble which is faithful to the ground-truth
parameterization (high \emph{SIM} values, bottom curves in
\autoref{fig_stability}) and which preserves the ground-truth clusters (colors
from dark red to light pink, \autoref{fig_stability}, high \emph{NMI/ARI}
values).}

\journal{As soon as noise is introduced ($\epsilon >= 0.05$), the strict
distance
$\wassersteinTree$ ($\epsilon_1 = 0$, leftmost column) becomes
unstable, as originally
documented by Pont et al. \cite{pont_vis21}. As a consequence, both the
ground-truth classification and parameterization are not recovered by MT-WAE in
the
2D layout: spheres of different colors are mixed together (as assessed
by the low \emph{NMI/ARI} values, leftmost curves,
\autoref{fig_stability})
and the spheres are no longer
organized along a 2D grid (as assessed by the lower \emph{SIM} values,
leftmost
curve, \autoref{fig_stability}). In contrast, with the original
Wasserstein distance
between persistence diagrams ($\epsilon_1 = 1$, rightmost column),
up to a significant level of noise ($\epsilon = 0.2$),
both
the ground-truth classification and parameterization are well preserved in the
2D layout generated by PD-WAE: the spheres with the same color remain clustered
(high \emph{NMI/ARI} values, rightmost curves) and the spheres are
properly
arranged along a 2D grid (high \emph{SIM} values, rightmost curve).}

\journal{For the recommended value of the control parameter $\epsilon_1$
($0.05$ \cite{pont_vis21}), MT-WAE still manages to recover well the
ground-truth parameterization and classification, up to a noise level of $
\epsilon = 0.1$ (perfect clustering, with high \emph{SIM} values). For a larger
value of $\epsilon_1$ ($\epsilon_1 \geq 0.1$), the 2D layouts generated by
MT-WAE are very
similar to these generated with PD-WAE (rightmost column), with
identical
stability indicators (\emph{SIM} and \emph{NMI/ARI} curves, bottom).}

\journal{In
conclusion,
this experiment shows that
for mild levels of noise ($\epsilon < 0.1$), the
recommended value of $\epsilon_1$ ($0.05$) results in a stable MT-WAE
computation.
For larger noise
levels,
MT-WAE provides similar
stability scores to PD-WAE for values of $\epsilon_1$ which are still
reasonable in terms of discriminative power ($\epsilon_1 = 0.1$).}




\end{document}